\documentclass[lettersize,journal]{IEEEtran}
\usepackage{amsmath,amsfonts}
\usepackage{amssymb}
\usepackage{algorithm}
\usepackage{algorithmicx}
\usepackage{array}
\usepackage[caption=false,font=footnotesize,labelfont=rm,textfont=rm]{subfig}
\usepackage{textcomp}
\usepackage{stfloats}
\usepackage{url}
\usepackage{verbatim}
\usepackage{graphicx}
\usepackage{cite}
\usepackage{ulem}
\usepackage{threeparttable}
\usepackage{newtxmath} % 设置数学公式为 Times 字体
\usepackage{enumitem}  % 加载 enumitem 包

\usepackage[
pdfauthor={derajan},
pdftitle={How to do this},
pdfstartview=XYZ,
bookmarks=true,
colorlinks=true,
linkcolor=blue,
urlcolor=blue,
citecolor=blue,
pdftex,
bookmarks=true,
linktocpage=true,   % makes the page number as hyperlink in table of content
hyperindex=true
]{hyperref}

\begin{document}

\title{Cognitive Fusion of ZC Sequences and Time-Frequency Images for Out-of-Distribution Detection of Drone Signals}

\author{Jie Li,
	Jing Li,~\IEEEmembership{Member,~IEEE,}
	Lu Lv,~\IEEEmembership{Member,~IEEE,}
	Zhanyu Ju,
	Fengkui Gong,~\IEEEmembership{Member,~IEEE,}
		\thanks{ \textit{(Corresponding author: Jing Li.)}}
		\thanks{Jie Li, Jing Li, Zhanyu Ju, and Fengkui Gong are with the State Key Laboratory of Integrated Services Network, Xidian University, Xi'an, Shaanxi 710071, China (e-mail: lijie\_372@stu.xidian.edu.cn; jli@xidian.edu.cn; juzhanyu@stu.xidian.edu.cn; fkgong@xidian.edu.cn).}
		\thanks{Lu Lv is with the School of Telecommunications Engineering, Xidian University, Xi'an 710071, China, (e-mail: lulv@xidian.edu.cn).}
}

\maketitle

\begin{abstract}
We propose a drone signal out-of-distribution detection (OODD) algorithm based on the cognitive fusion of Zadoff-Chu (ZC) sequences and time-frequency images (TFI). ZC sequences are identified by analyzing the communication protocols of DJI drones, while TFI capture the time-frequency characteristics of drone signals with unknown or non-standard communication protocols. Both modalities are used jointly to enable OODD in the drone remote identification (RID) task. Specifically, ZC sequence features and TFI features are generated from the received radio frequency signals, which are then processed through dedicated feature extraction module to enhance and align them. The resultant multi–modal features undergo multi–modal feature interaction, single–modal feature fusion, and multi–modal feature fusion to produce features that integrate and complement information across modalities. Discrimination scores are computed from the fused features along both spatial and channel dimensions to capture time-frequency characteristic differences dictated by the communication protocols, and these scores will be transformed into adaptive attention weights. The weighted features are then passed through a Softmax function to produce the signal classification results. Simulation results demonstrate that the proposed algorithm outperforms existing algorithms and achieves 1.7\% and 7.5\% improvements in RID and OODD metrics, respectively. The proposed algorithm also performs strong robustness under varying flight conditions and across different drone types.
\end{abstract}

\begin{IEEEkeywords}
Out-of-distribution detection, Zadoff-Chu sequences, time-frequency images, multi-modal feature interaction and fusion, drone remote identification, discrimination score.
\end{IEEEkeywords}

\section{Introduction}
\label{sec1}
\IEEEPARstart{T}{he} low-altitude network, with drones as its primary platforms, provides efficient and reliable technological support for the low-altitude economy, empowering intelligent innovations in logistics, aerial entertainment, urban management, and emergency communication\textcolor{blue}{\cite{ref147}},\textcolor{blue}{\cite{ref148}},\textcolor{blue}{\cite{ref154}}. With the widespread deployment of drones and their exceptional operational flexibility, drone detection and classification, also known as remote identification (RID), has attracted increasing attention in both cooperative and non-cooperative scenarios\textcolor{blue}{\cite{ref1}},\textcolor{blue}{\cite{ref31}},\textcolor{blue}{\cite{ref162}}. Drone RID enables the management of flight states for authorized drones to ensure orderly low-altitude operations, while enabling the identification of unlicensed or unauthorized drones to support subsequent regulatory enforcement. Existing drone RID works primarily rely on radar, vision, audio, and radio frequency (RF) methods, among which time-frequency electromagnetic analysis of RF signals has emerged as the most widely adopted approach\textcolor{blue}{\cite{ref126}}.

However, RID algorithms trained under closed-set assumptions often exhibit substantial performance degradation in open-set scenarios, primarily because classifiers tend to misidentify out-of-distribution (OOD) signals as in-distribution (ID) classes seen during training. Thus, effective OOD detection (OODD)\textcolor{blue}{\cite{ref182}} or open-set recognition\textcolor{blue}{\cite{ref139}} is crucial for reliable drone RID. By analyzing communication protocols, such as signal frame structures and modulation parameters, RID can be achieved for drones employing standardized communication protocols. Nevertheless, this method is less effective for drones utilizing non-standard or unknown modulation parameters and frame structures. Notably, for the widely deployed Da-Jiang Innovations (DJI) drones, proprietary Zadoff-Chu (ZC) sequences have been identified\textcolor{blue}{\cite{reftccn}}, and different types of drones exhibit distinct time-frequency characteristics, including bandwidth, duration, and transmission intervals, which lead to observable variations in time-frequency images (TFI). Therefore, deploying OOD signal detection through the cognitive fusion of ZC sequences and TFI constitutes a practical and effective strategy for achieving robust and reliable drone RID. 

\subsection{Related Works}
\begin{enumerate}[leftmargin=0pt, itemindent=2pc, listparindent=\parindent]
	\item{\textit{Drone RID}}:
	 For drones weighing less than 249 grams, their relatively small radar cross-section leads to weak detection features in radar echo, thereby increasing the probability of false alarms\textcolor{blue}{\cite{ref231}},\textcolor{blue}{\cite{ref163}}. In vision-based methods, the RID accuracy for images or videos is significantly degraded by factors such as motion blur, adverse weather, and intermittent drones visibility\textcolor{blue}{\cite{ref158}},\textcolor{blue}{\cite{ref170}},\textcolor{blue}{\cite{ref172}}. Acoustic-based RID algorithms require quiet environments and short flight distances, which are incapable of distinguishing individual drones within a swarm\textcolor{blue}{\cite{ref220}}. In contrast, RID based on extracting time and frequency domain features from RF signals is robust to variations in drone size and adverse weather. For instance, raw in-phase/quadrature (I/Q) sequences were sampled to distinguish individuals from 10 DJI Mini 2 drones using a convolutional neural network (CNN)\textcolor{blue}{\cite{ref193}}. The received signal strength (RSS), computed from the magnitude of I/Q sequences, has been employed for drone differentiation and localization over 865 MHz and 2.4 GHz bands\textcolor{blue}{\cite{ref156}}. Besides, joint communication and sensing techniques have been exploited to estimate the position and velocity of drones\textcolor{blue}{\cite{ref157}}.
	 
	 Drone signals contains uplink frequency hopping spread spectrum (FHSS) control signals and downlink orthogonal frequency division multiplexing (OFDM) video transmission signals, which exhibit distinct time-frequency characteristics governed by communication protocols. Thus, the received RF signals can be transformed into TFI via short-time Fourier transform (STFT) to simultaneously reveal both signal modulation parameters and frame structures. The transmission durations and bandwidths extracted from TFI were utilized to identify drones, and a spectrogram windowing algorithm was proposed to reduce computational complexity\textcolor{blue}{\cite{ref153}}. To improve robustness under diverse environmental conditions, a federated semantic regularization-based RID algorithm was proposed and evaluated across different flight distances\textcolor{blue}{\cite{ref221}}. By analyzing frame structures and modulation parameters, ZC sequences with favorable correlation properties have been utilized to enhance RID performance under low signal-to-noise ratio (SNR) and interference-prone conditions\textcolor{blue}{\cite{reftccn}}. However, algorithms designed exclusively for ID drone signals will misclassify OOD signals that unseen during training as ID, thereby degrading RID performance in practical deployment scenarios.
	 
	 \item{\textit{OOD Detection}}:
	 Existing OODD methods are broadly categorized into training modifications and post-hoc analysis\textcolor{blue}{\cite{ref182}}. The former integrates OOD data during the training or fine-tuning of the classification model, whereas the latter achieves OODD by exploiting the statistical properties of the input data\textcolor{blue}{\cite{ref140}}. Methods that leveraging auxiliary OOD data primarily employ generative adversarial networks or diffusion-based generative models to improve OODD accuracy, but they face challenges in simulating the full diversity of potential OOD samples\textcolor{blue}{\cite{ref243}},\textcolor{blue}{\cite{ref242}},\textcolor{blue}{\cite{ref244}},\textcolor{blue}{\cite{ref241}}.
	 
	 Post-hoc methods identify OOD samples solely from the statistical characteristics induced by the interaction between the input and the model, which have been widely adopted in practice. In neural networks, the feature vector from the final layer will passed through a fully connected layer to generate logits, which are then transformed into classification probabilities via the Softmax function. Since Softmax function typically yields higher confidence scores for ID samples and lower scores for OOD samples, a threshold was applied to facilitate OODD\textcolor{blue}{\cite{ref32}}. However, Softmax may produce overconfident predictions for OOD samples, leading to spuriously high confidence scores. To mitigate this, a Dirichlet distribution-based output was proposed to jointly model the predicted class and the associated decision uncertainty\textcolor{blue}{\cite{ref222}}. This uncertainty was further extended to capture three types of uncertainty in image feature matching\textcolor{blue}{\cite{ref76}}, or represented by a confidence score generated from an additional confidence branch integrated into the network\textcolor{blue}{\cite{ref245}}. OODD has also been achieved by encoding abstract features and employing Euclidean distance-based similarity prediction\textcolor{blue}{\cite{ref194}}. To address the scale inconsistency and inter-dimensional correlation inherent in Euclidean distance, Mahalanobis distance based on feature normalization and the covariance matrix was adopted, which has been shown to improve the classification performance\textcolor{blue}{\cite{ref246}}. Leveraging the fact that ID samples are more easily to be reconstructed by encoder–decoder architecture\textcolor{blue}{\cite{ref230}}, a reconstruction error-based algorithm was proposed for OODD, which revealed that low-frequency image components were reconstructed more easily than high-frequency ones\textcolor{blue}{\cite{ref136}}. Since convolutional kernels in neural networks exhibit stronger responses to inputs matching the ID distribution, OODD was performed by analyzing the average magnitude of kernel responses and the frequency of salient activations in several convolutional layers\textcolor{blue}{\cite{ref78}}. For a feature vector of size \(H\!\times \!W\!\times\!C\!\), different channels contributed either positively or negatively to OODD, thus channel-wise masking vectors were computed based on inter-class similarity and variance, and the impact of discarding specific numbers of channels were analyzed to maximize OODD performance\textcolor{blue}{\cite{ref91}}. Recently, large language models (LLMs) have been integrated with task-specific instruction tuning to enhance OODD generalization across diverse tasks\textcolor{blue}{\cite{ref232}}. Furthermore, when a model repository contains sufficiently diverse pre-trained OOD models, a zoo-based approach has been shown to effectively leverage model diversity and complementarity to improve OODD performance\textcolor{blue}{\cite{ref89}}.
\end{enumerate}

\subsection{Motivations and Contributions}
Although previous studies have proposed numerous algorithms for drone RID and OODD, several key challenges still remain, as outlined below:

\begin{itemize}
	\item{The issue of OODD in drone RID remains underexplored. Most existing works treat RID and OODD as separate problems, neglecting their inherent interdependence. Considering the diverse range of drone communication protocols, including both standardized formats such as those employed by DJI and numerous non-standard proprietary protocols, OODD inevitably emerges as a critical challenge within the RID task.
	}
	\item{Features for OODD in drone RID must be carefully designed to account for the unique signal characteristics. While numerous feature extraction and fusion algorithms have been proposed for classification tasks in computer vision, these approaches are predominantly tailored to object image classification\textcolor{blue}{\cite{ref231}},\textcolor{blue}{\cite{ref164}} or single-modality features\textcolor{blue}{\cite{ref163}}, making them less applicable for OODD in the context of drone RID.
	}
\end{itemize}

To address the above challenges, we propose a drone signal OODD algorithm based on the cognitive fusion of ZC sequences and TFI, which is evaluated on three practical datasets named DroneRFa\textcolor{blue}{\cite{refset1}}, DroneRFb-DIR\textcolor{blue}{\cite{refset2}}, and RFUAV\textcolor{blue}{\cite{refset3}}. Specifically, ZC sequence feature and TFI feature derived from RF signals are fed into a neural network, where they undergo feature interaction and fusion through specially designed modules to form a cognitive fusion feature. The fused feature is further enhanced by spatial and channel-wise attention weight based on the time-frequency characteristic of RF signals. Finally, signal classification is performed via the Softmax function. The contributions of this paper are summarized as follows.

\begin{itemize}
	\item{ZC sequences and TFI are cognitively fused to enable drone signal OODD. For drone types employing ZC sequences, the ZC sequence feature can accurately achieve drone RID even under low-SNR and interference-prone conditions. For drone types without ZC sequences, the TFI feature captures the time–frequency characteristic differences dictated by the communication protocols, enabling OODD even for drones operating with non-standard communication protocols.
	}
	\item{A feature extraction, interaction, and fusion network specifically designed for ZC sequences and TFI is developed. ZC sequence feature and TFI feature are generated based on the analysis of RF signals' frame structures and modulation parameters, with corresponding parallel feature extraction network branches designed to enhance and align these features. Then, multi-modal feature interaction, single-modal feature fusion, and multi-modal feature fusion modules are introduced to fully integrate and complement the information from different modalities.
	}
	\item{Exploiting the differences in signal time–frequency characteristics, adaptive attention weights are applied to the fused features along both spatial and channel dimensions. For the cognitively fused features, the inter-class similarity and variance across different signal categories are computed separately in the spatial and channel dimensions, and then combined to obtain discriminative scores that quantify the time–frequency characteristic differences. The scores are subsequently transformed by the neural network into adaptive spatial and channel attention weights to enhance the OODD accuracy.
	}
	\item{Numerical results demonstrate that the proposed algorithms effectively achieve OODD in drone RID task across different channel conditions and drone types. Specifically: 1) The proposed algorithm achieves at least 1.7\% and 7.5\% improvement in RID and OODD accuracy over the existing algorithms, proving the effectiveness of cognitive multi-modal feature interaction and fusion of ZC sequences and TFI. 2) The algorithm exhibit strong robustness across variable flight distances, line-of-sight (LoS) conditions, and drone types. 3) The visualization results of adaptive attention weights indicate that the analysis of time-frequency characteristic can achieve the discrimination of different communication protocols.
	}
\end{itemize}
\subsection{Outline and Notations}
The signal analysis is presented in Section \ref{sec2}. The details of drone OODD algorithm are discussed in Section \ref{sec3}. Simulation evaluation is presented in Section \ref{sec4}. And this paper is summarized in Section \ref{sec5}.

\section{Signal Analysis}
\label{sec2}
\subsection{Signal Dateset}

The DroneRFa dataset was collected in an urban environment by sampling with \(f_s=100\)  MHz over the 915 MHz, 2.4 GHz, and 5.8 GHz frequency bands. The dataset includes a wide range of drone-related signals, such as uplink control signals, downlink video transmission signals, and specialized broadcast frames, as well as signals from other wireless technologies, including Wi-Fi and Bluetooth. Each drone mission commenced from a stationary state, followed by flight over a certain distance before returning, thereby providing RF signal data across varying flight distances.

The DroneRFb-DIR dataset was collected in an urban environment by sampling the 2.4 GHz frequency band with \(f_s=80\) MHz and includes the same types of signals as those in the DroneRFa dataset. Notably, this dataset provides RF signals captured under both LoS and non-LoS (NLoS) flight conditions, as well as signals from multiple individual drones of the same type.

The RFUAV dataset was collected by sampling the 915 MHz, 2.4 GHz, and 5.8 GHz frequency bands with \(f_s=100\) MHz, and includes the same types of signals as those in the DroneRFa dataset. As one of the most recent drone RF signal datasets, it encompasses not only widely used DJI drones but also other drone types such as the DAUTEL EVO NANO and Herelink HX4.

The interpretation of the binary labels is provided in Table \ref{tab:table1}. Furthermore, to more realistically evaluate the robustness of the proposed algorithm in practical scenarios, additive white Gaussian noise (AWGN) is introduced into the RF signals, forming the dataset used in this paper.

\begin{table*}[!t]   
	\caption{Explanation of the Dataset's Labels}  
	\label{tab:table1} 
	\centering
	\begin{threeparttable}
		\begin{tabular}{|c|c|c|c|c|}   
			\hline   \textbf{Label1} & \textbf{Labe2} & \textbf{Labe3} & \textbf{Labe4} & \textbf{Drone Types} \\   
			\hline   T00000 & D00 & S00 S01 & L0\(\sim\)L3 & Background Noise \\ 
			\hline   T00001 & D00 D01 D10 & S00 S01 & L0\(\sim\)L3 & DJI Phantom 3 \\ 
			\hline   T00010 & D00 D01 D10 & S00 S01 & L0\(\sim\)L3 & DJI Phantom 4 Pro \\ 
			\hline   T00011 & D00 D01 D10 & S00 & L0\(\sim\)L3 & DJI MATRICE 200 \\ 
			\hline   T00100 & D00 D01 D10 & S00 & L0\(\sim\)L3 & DJI MATRICE 100 \\ 
			\hline   T00101 & D00 D01 D10 & S00 S01 & L0\(\sim\)L3 & DJI Air 2S \\
			\hline   T00110 & D00 D01 D10 & S00 & L0\(\sim\)L3 & DJI Mini 3 Pro \\
			\hline   T00111 & D00 D01 D10 & S00 & L0\(\sim\)L3 & DJI Inspire 2 \\
			\hline   T01000 & D00 D01 D10 & S00 & L0\(\sim\)L3 & DJI Mavic Pro \\
			\hline   T01001 & D00 D01 D10 & S00 & L0\(\sim\)L3 & DJI Mini 2 \\
			\hline   T01010 & D00 & S00 & L0\(\sim\)L3 & DJI Mavic 3 \\
			\hline   T01011 & D00 & S00 & L0\(\sim\)L3 & DJI MATRICE 300 \\ 
			\hline   T01100 & D00 & S00 S01 & L0\(\sim\)L3 & DJI Phantom 4 Pro RTK \\
			\hline   T01101 & D00 & S00 & L0\(\sim\)L3 & DJI MATRICE 30T \\
			\hline   T01110 & D00 & S00 & L0\(\sim\)L3 & DJI AVATA \\
			\hline   T01111 & D00 & S00 & L0\(\sim\)L3 & Manually assembled Drone with DJI communication modules \\
			\hline   T10000 & D00 & S00 S01 & L0\(\sim\)L3 & DJI Mavic 3 Pro \\
			\hline   T10001 & D00 & S00 S01 & L0\(\sim\)L3 & DJI Mini 2 SE \\
			\hline   T10010 & D00 & S00 S01 & L0\(\sim\)L3 & DJI Mini 3 \\
			\hline   T10011 & D00 & S00 & L0\(\sim\)L3 & DAUTEL EVO NANO \\
			\hline   T10100 & D00 & S00 & L0\(\sim\)L3 & DJI FPV COMBO \\
			\hline   T10101 & D00 & S00 & L0\(\sim\)L3 & FLYSKY FS I6X \\
			\hline   T10110 & D00 & S00 & L0\(\sim\)L3 & Herelink Hx4 \\
			\hline   T10111 & D00 & S00 & L0\(\sim\)L3 & SKYDROID H12 \\
			\hline   T11000 & D00 & S00 & L0\(\sim\)L3 & FUTABA T16IZ \\
			\hline   
		\end{tabular}
		\begin{tablenotes}
			\footnotesize
			\item[1] D00, D01, and D10 denote the flight distance range of 20\(\sim\)40, 40\(\sim\)80, and 80\(\sim\)150 m, respectively.
			\item[2] S00 and S01 denote LoS and NLoS flight condition, respectively.
			\item[3] L0, L1, L2, and L3 denote the sampling duration range of 10, 20, 50, and 100 ms, respectively.
		\end{tablenotes}            % 添加命令
	\end{threeparttable}  
\end{table*}

\subsection{Frame Structure and Modulation Parameters}
Let \(x(l)\) denote the received complex RF signal sequences of length \(L\) with \(f_s\), STFT is applied to obtain \(\mathbf{X}\) for revealing its time-frequency characteristics. The value of \(\mathbf{X}\) at time index \(t\) and frequency index \(f\) is given by
\begin{equation}
	\label{Eq1}
	{{\mathbf{X}}_{t,f}}=\left| \sum\limits_{u =-\infty }^{+\infty }{x(u){h}(u-t){{\text{e}}^{\frac{-\text{j}2\pi\!fu}{U} }}} \right|,
\end{equation}
where \(h(u-t)\) denotes the window function with length \(U\), and the fast Fourier transform size is also equal to \(U\). For signal \(x(l)\) of length \(L\), the computational complexity of generating one TFI is \(\mathcal{O}( L\log U )\). \(\mathbf{X}\) with size \(T\!\times\!F\) will be converted into TFI in RGB format using MATLAB built-in functions, which include logarithmic compression and normalization. TFI is typically resized to \(224\!\times\!224\!\times\!3\), and the computational complexity of this transformation is \(\mathcal{O}(4TF)\).
\begin{figure*}[!t]
	\centering
	\subfloat[]{\includegraphics[width=2.3in]{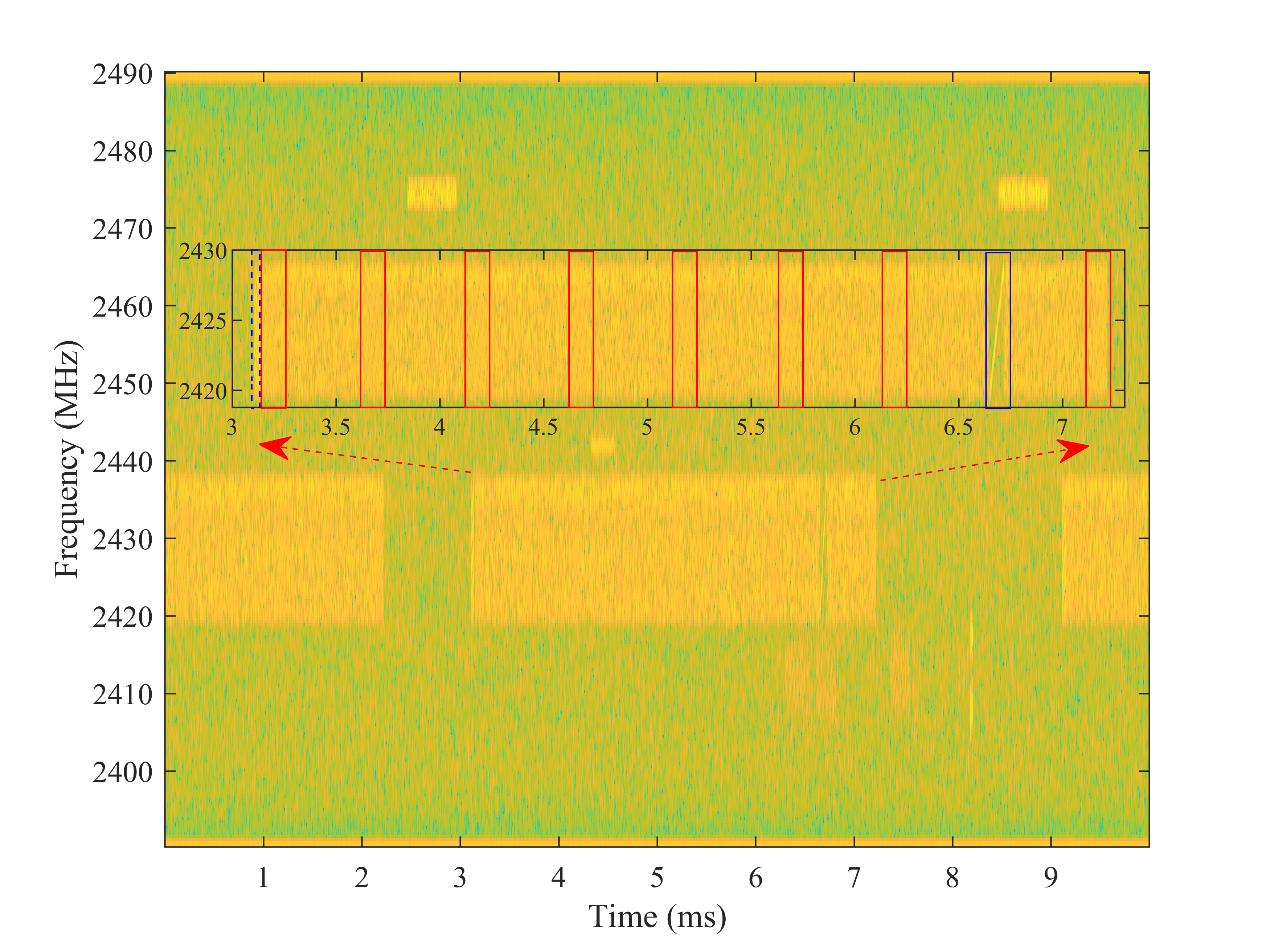}
		\label{Fig_tfi_detail_ori}}
	\hfil
	\subfloat[]{\includegraphics[width=2.3in]{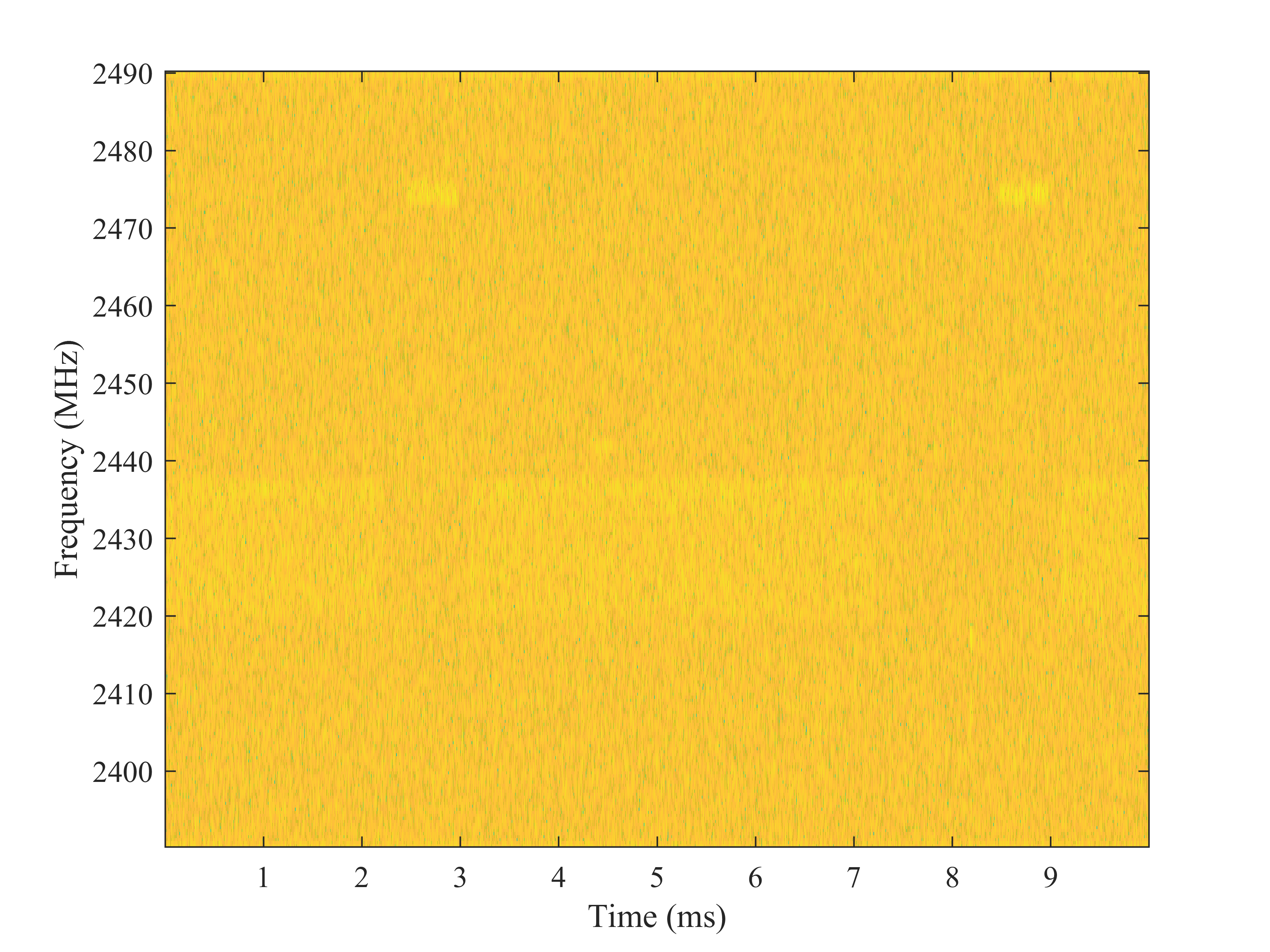}
		\label{Fig_tfi_detail_noi}}
	\hfil
	\subfloat[]{\includegraphics[width=2.3in]{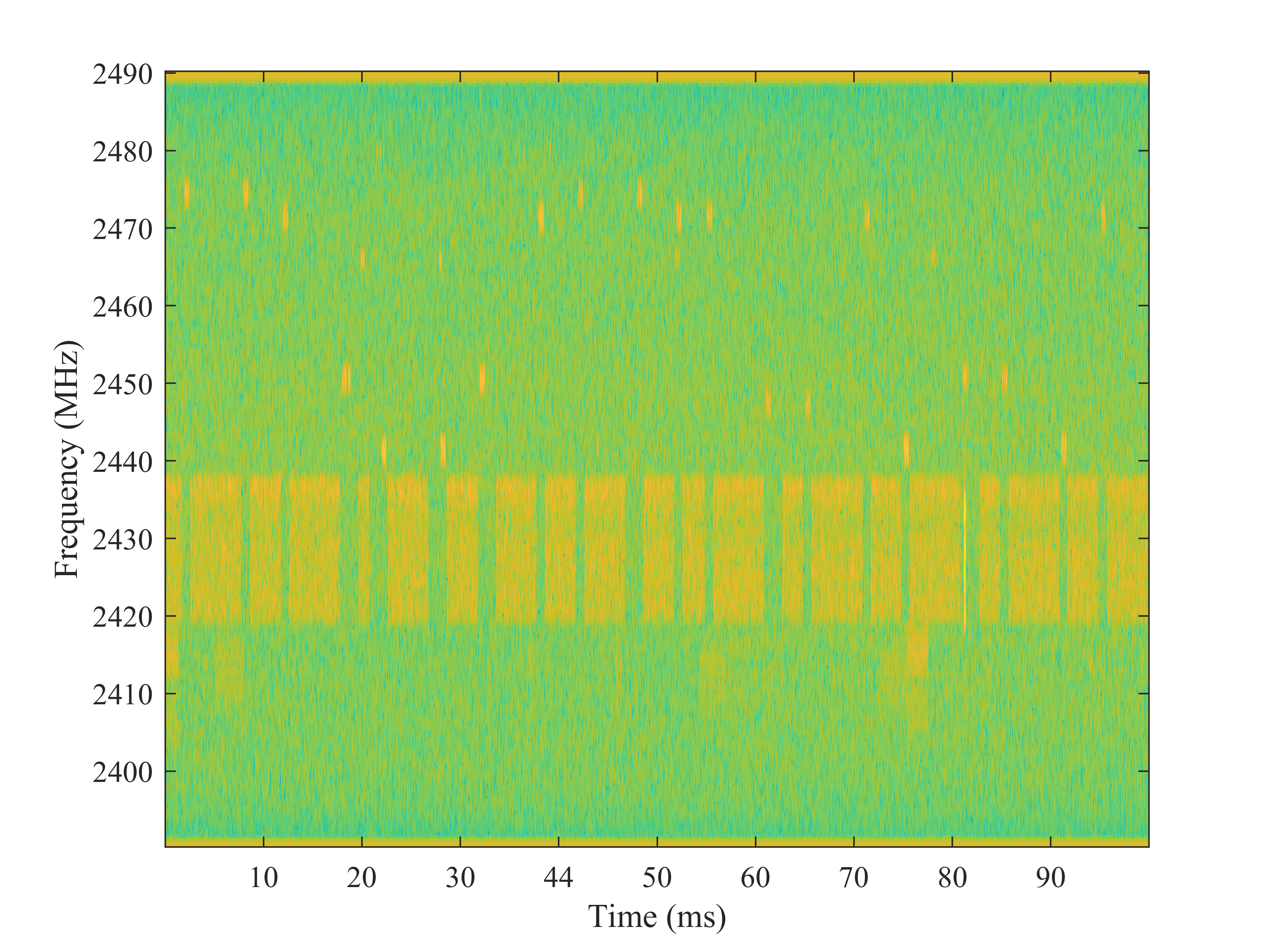}
		\label{Fig_tfi_detail_len}}
	\hfil
	\caption{TFI of T1011D00S00. (a) \(\tiny{L = 10^6}\) at SNR = 15 dB. (b) \(\tiny{L = 10^6}\) at SNR = -7 dB. (c) \(\tiny{L = 10^7}\) at SNR = 15 dB.}
	\label{Fig_tfi1}
\end{figure*}

Fig. \ref{Fig_tfi_detail_ori} illustrates the TFI of T1011D00S00, where a video transmission signal frame is highlighted. The blue dashed box, blue solid box, and red solid box indicate three distinct types of preambles for channel estimation and synchronization, among which the latter two have been identified ZC sequences with different roots\textcolor{blue}{\cite{reftccn}}. It is worth noting that:
\begin{itemize}
	\item{Preambles at the frame head are not always present.
	}
	\item{Different types of drone  may adopt multiple ZC sequences with varying roots—or none at all.
	}
	\item{For a given drone type, although multiple ZC sequences with different roots may be utilized, typically only one root is used most frequently.
	}
	\item{The most frequently used ZC root differs across drone types, whereas less frequently used roots may be shared among several types.
	}
\end{itemize}

Although time-frequency characteristics regulated by communication protocol such as bandwidth, duration, and transmission interval can be visually recognized from TFI, the TFI becomes increasingly ambiguous in the presence of complex interference and low SNR conditions, as illustrated in Fig. \ref{Fig_tfi_detail_noi}.

Therefore, an in-depth exploration of communication protocols to extract more robust and interpretable features is essential. Based on the estimation of bandwidth \(B\), number of subcarriers \(N\) and \(N_u\), the parameters of the most frequently used ZC sequence have been identified for drones\textcolor{blue}{\cite{reftccn}}. Specifically, ZC sequence can be generated by
\begin{equation}
	\label{Eq2}
	{z_r}(v)={\text{e}^{-\text{j}\frac{\pi rv(v+1)}{V}}},v=0,1,\ldots ,V-1,
\end{equation}
where \(r\) and \(V=N_u + 1\) denote the root and sequence length, respectively. For ZC sequences with the same \(r\) and \(V\), the cross-correlation results between them will have a amplitude peak, while weak cross-correlation characteristic for different \(r\) even with the same \(V\). Thus, cross-correlation results between drone RF signals and the locally generated ZC sequences can be utilized to improve the drone RID accuracy and robustness under situations with complex interference and low SNR.

\subsection{Feature Generation}
\begin{enumerate}[leftmargin=0pt, itemindent=2pc, listparindent=\parindent]
	\item{\textit{TFI Feature}}:
	The TFI feature refers to the RGB image matrix \(\mathbf{I}\) converted from \(\mathbf{X}\), and \(\mathbf{I}\) will undergo a nonlinear transformation to expand the dynamic range of low values while keeping the range of high values relatively unchanged\textcolor{blue}{\cite{ref133}}, which can be expressed as
	\begin{equation}
		\label{Eqlog}
		\mathbf{I}=\log(\mathbf{I}).
	\end{equation}
	
	Besides, the variations in the sampling duration or \(L\) will affect how the time-frequency characteristics of the signal are represented with fixed \(f_s\). Specifically, as illustrated in Fig. \ref{Fig_tfi_detail_ori}, the fine-grained intra-frame texture is clearly preserved with low computational complexity for relatively small \(L\). However, inter-frame characteristics such as frame duration and transmission intervals cannot be effectively captured. Conversely, as shown in Fig. \ref{Fig_tfi_detail_len}, increasing \(L\) changes the representation of both intra-frame and inter-frame time-frequency features, but at the cost of significantly higher computational complexity.
	
	In this paper, \(L=[10^5,2\times10^5,5\times10^5,10^6]\) corresponding to sampling durations \([\text{L}0,\text{L}1,\text{L}2,\text{L}3]\) are considered, among which \(L=10^6\) is selected to retain sufficient intra-frame details while incorporating enough frames to reflect inter-frame features, as illustrated in Fig. \ref{Fig_tfi}. It can be observed that when \(L=10^6\), TFI ensures sufficiently low computational complexity while capturing rich inter-frame time-frequency characteristics, which can be leveraged to distinguish RF signals to achieve drone RID.
	\begin{figure*}[!t]
		\centering
		\subfloat[]{\includegraphics[width=1.7in]{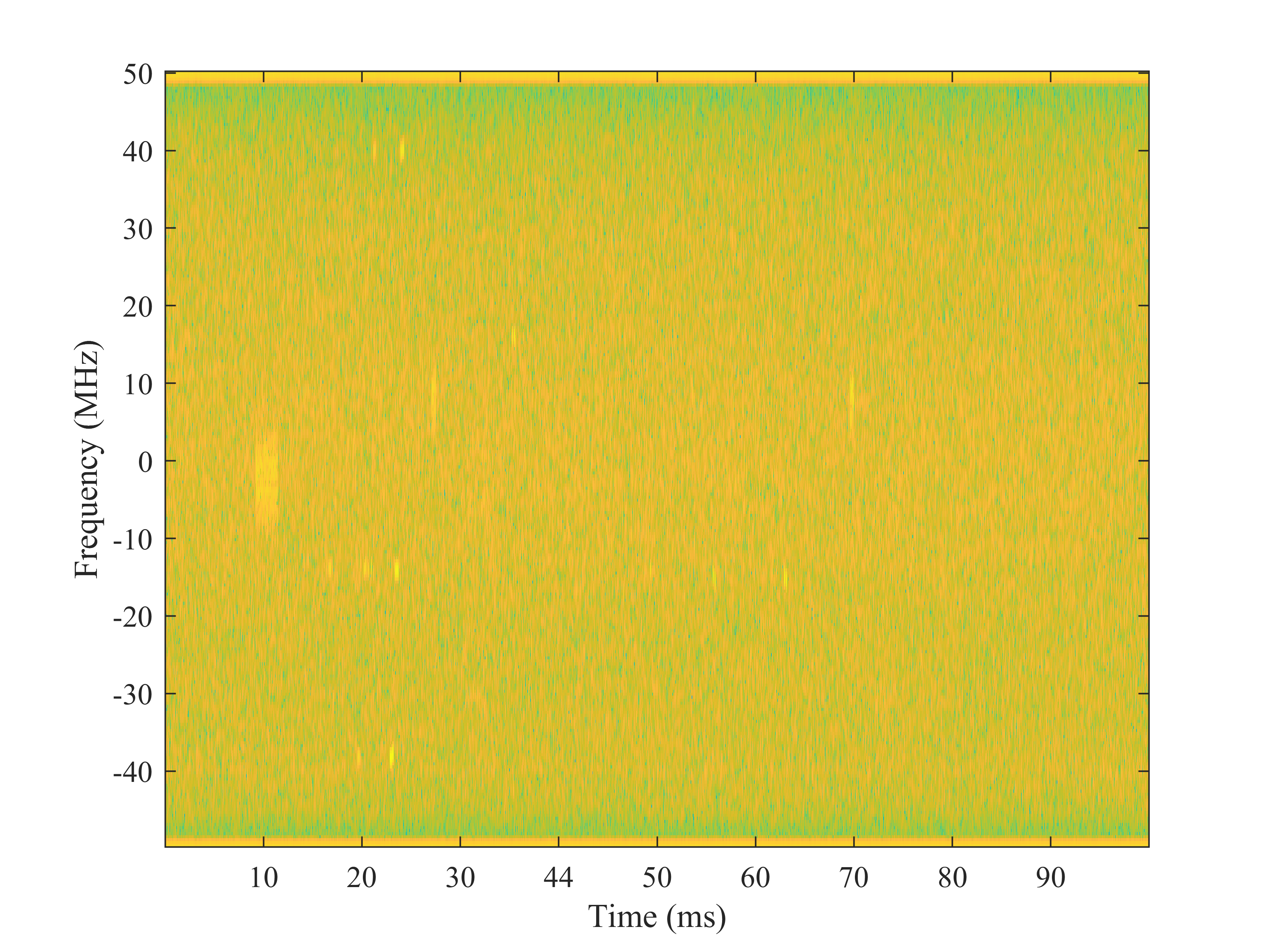}
			\label{tfi_1}}
		\hfil
		\subfloat[]{\includegraphics[width=1.7in]{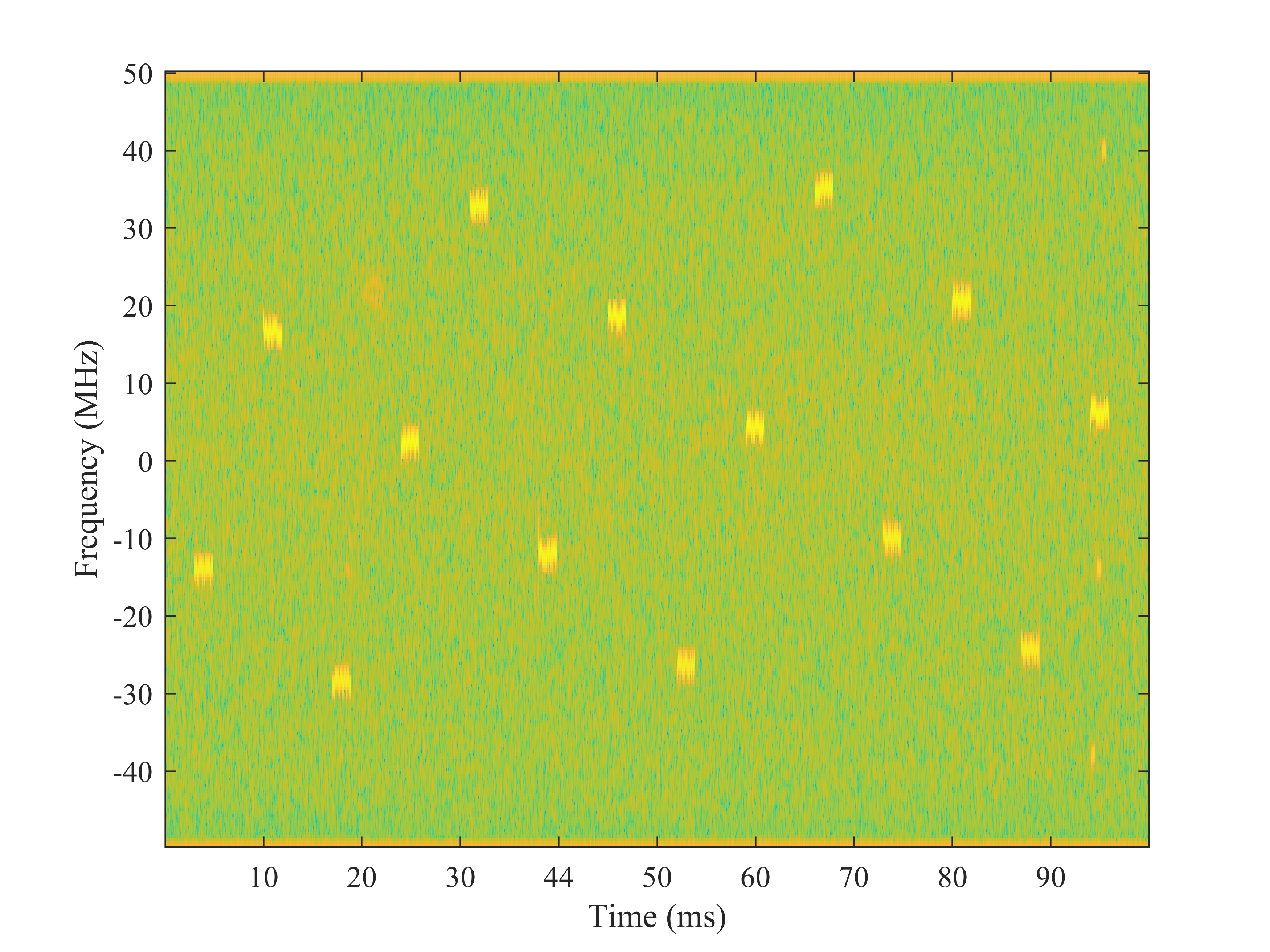}
			\label{tfi_2}}
		\hfil
		\subfloat[]{\includegraphics[width=1.7in]{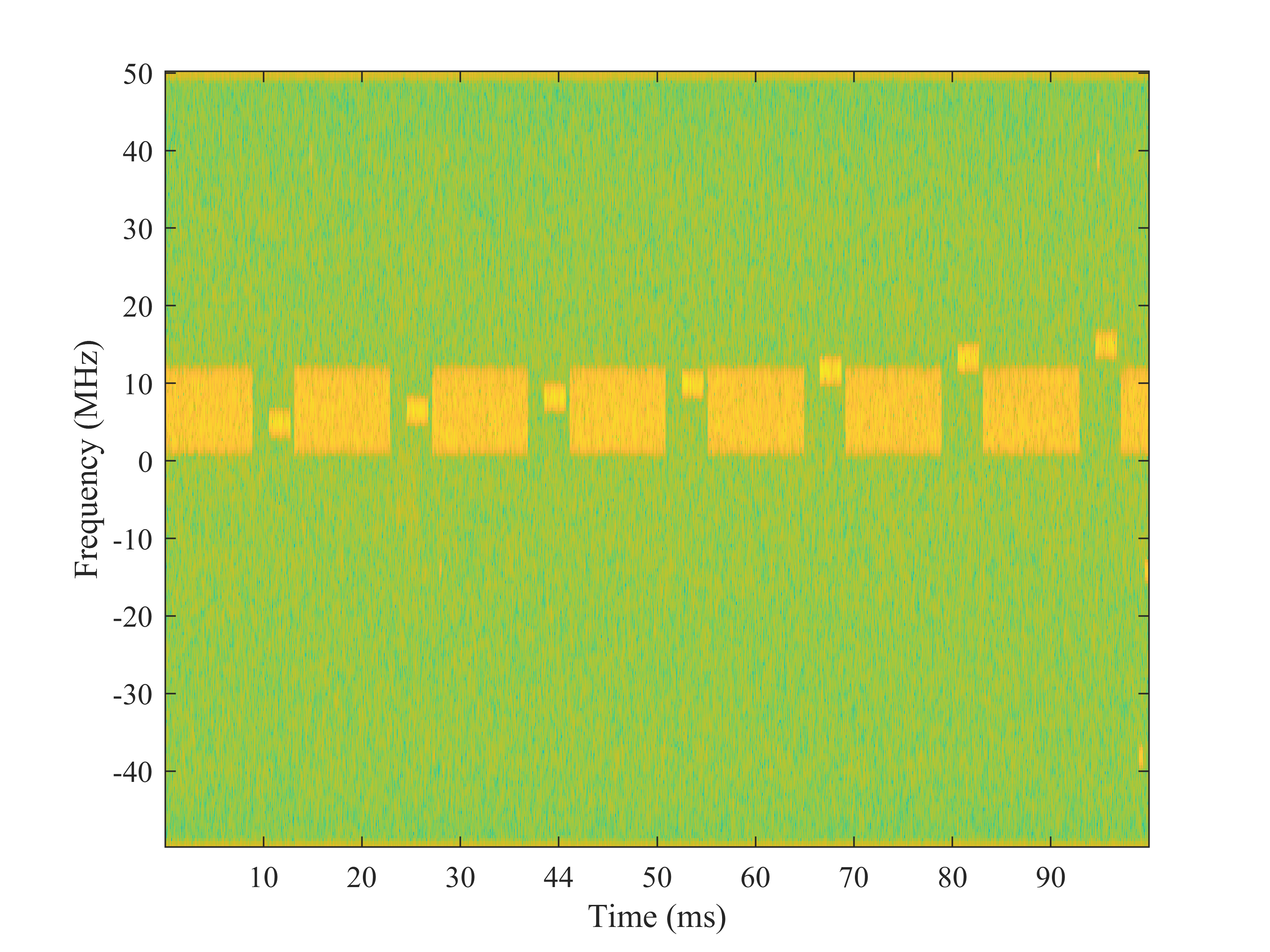}
			\label{tfi_3}}
		\hfil
		\subfloat[]{\includegraphics[width=1.7in]{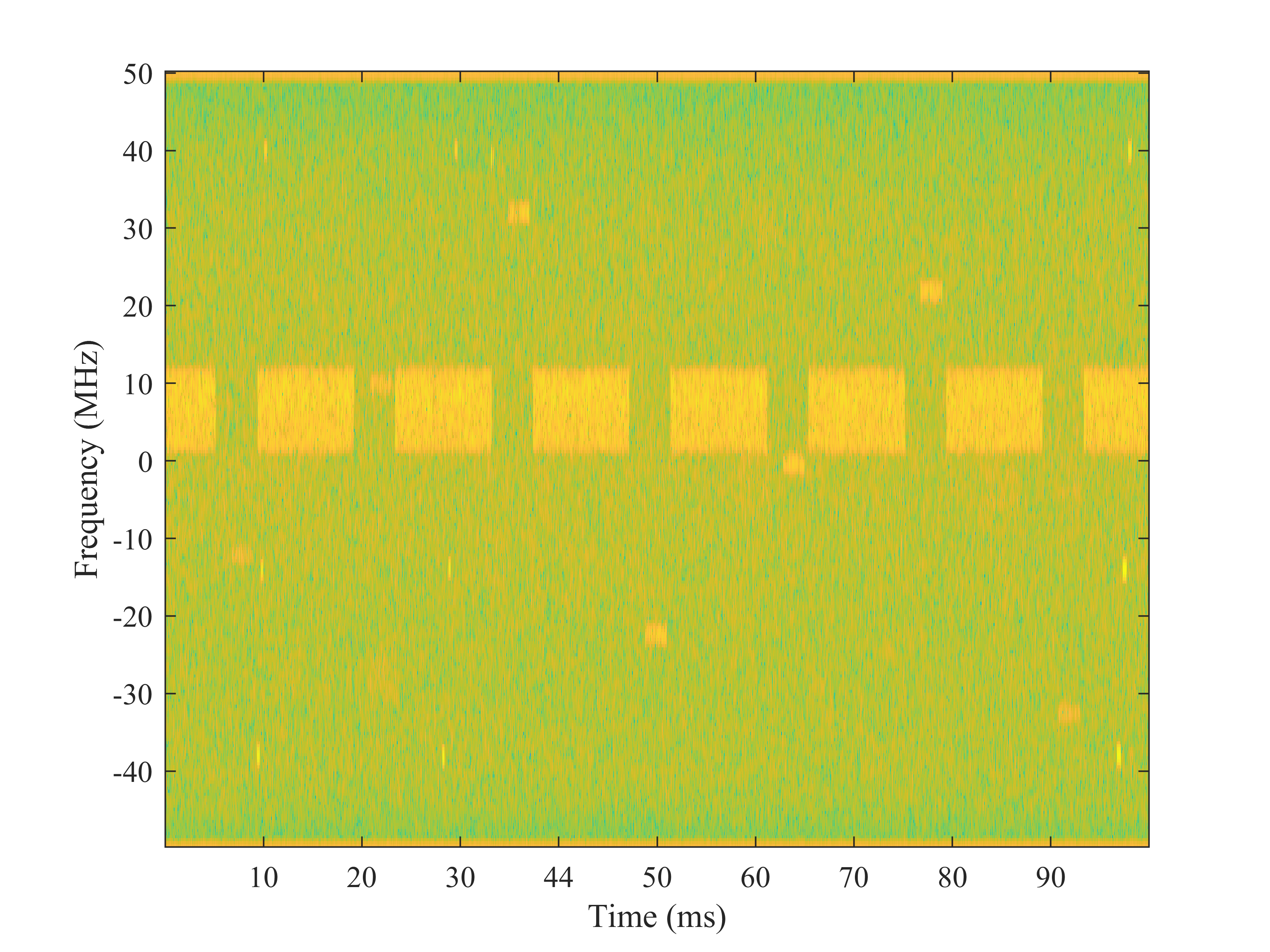}
			\label{tfi_4}}
		\hfil
		\subfloat[]{\includegraphics[width=1.7in]{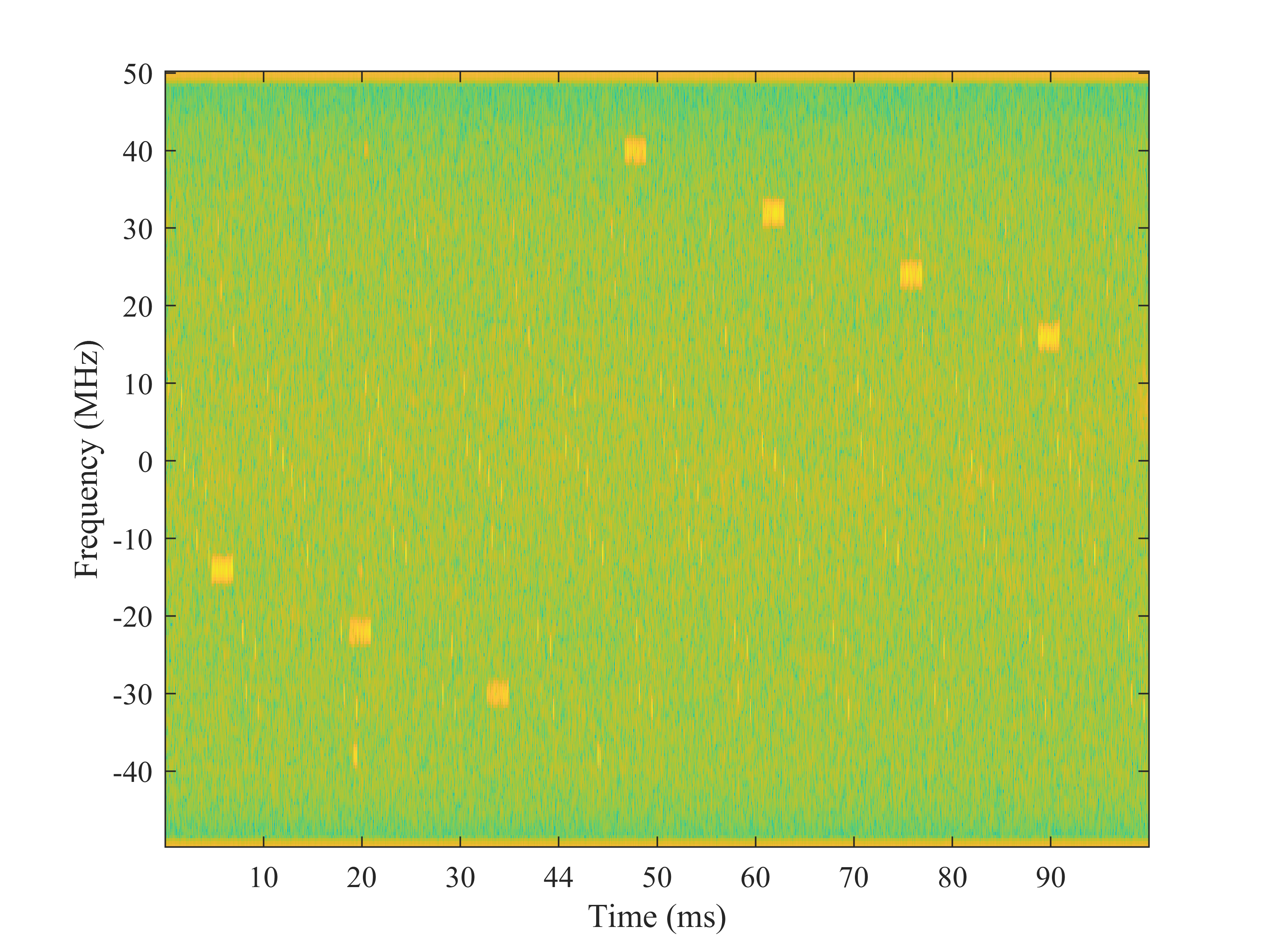}
			\label{tfi_5}}
		\hfil
		\subfloat[]{\includegraphics[width=1.7in]{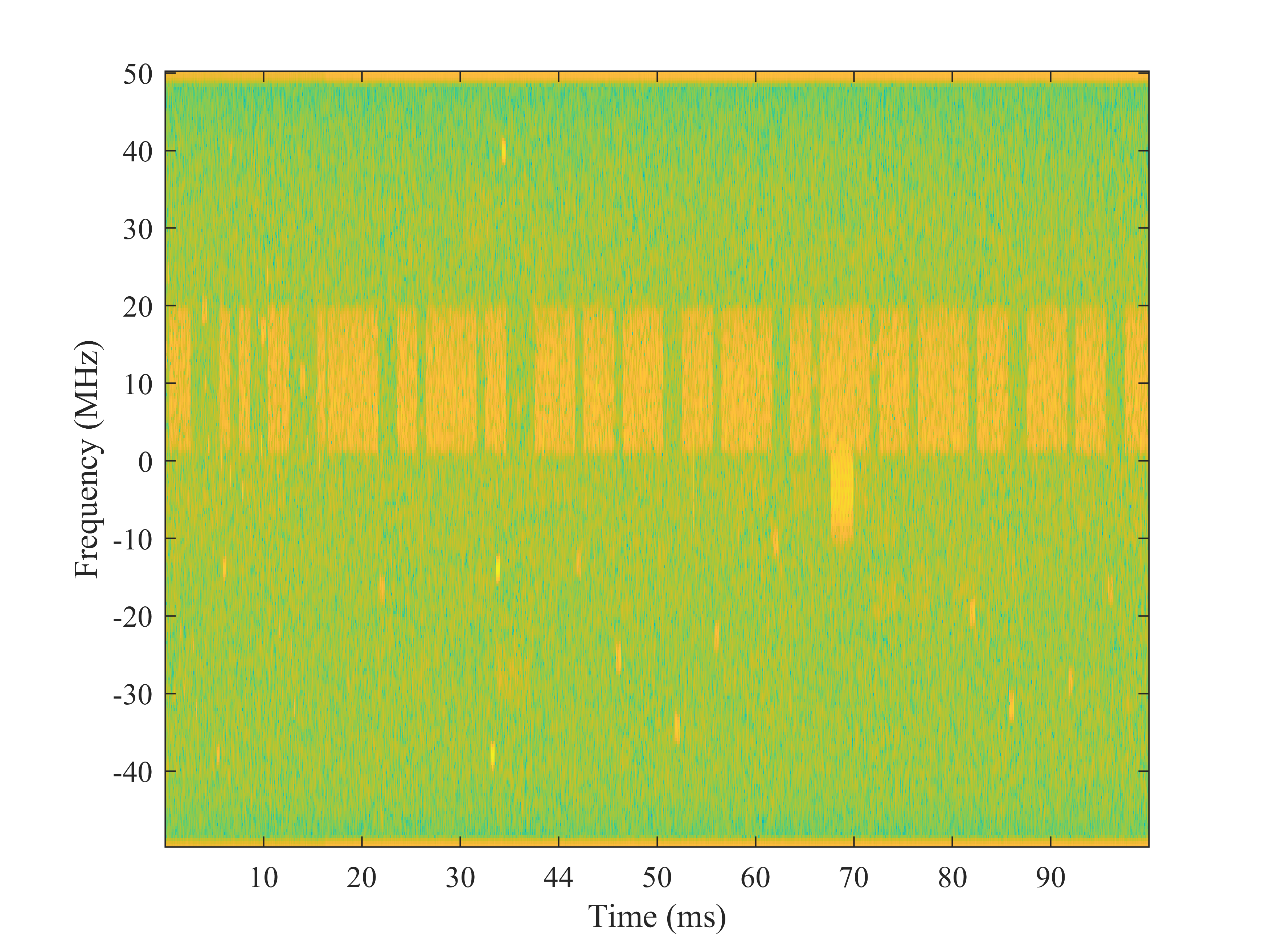}
			\label{tfi_6}}
		\hfil
		\subfloat[]{\includegraphics[width=1.7in]{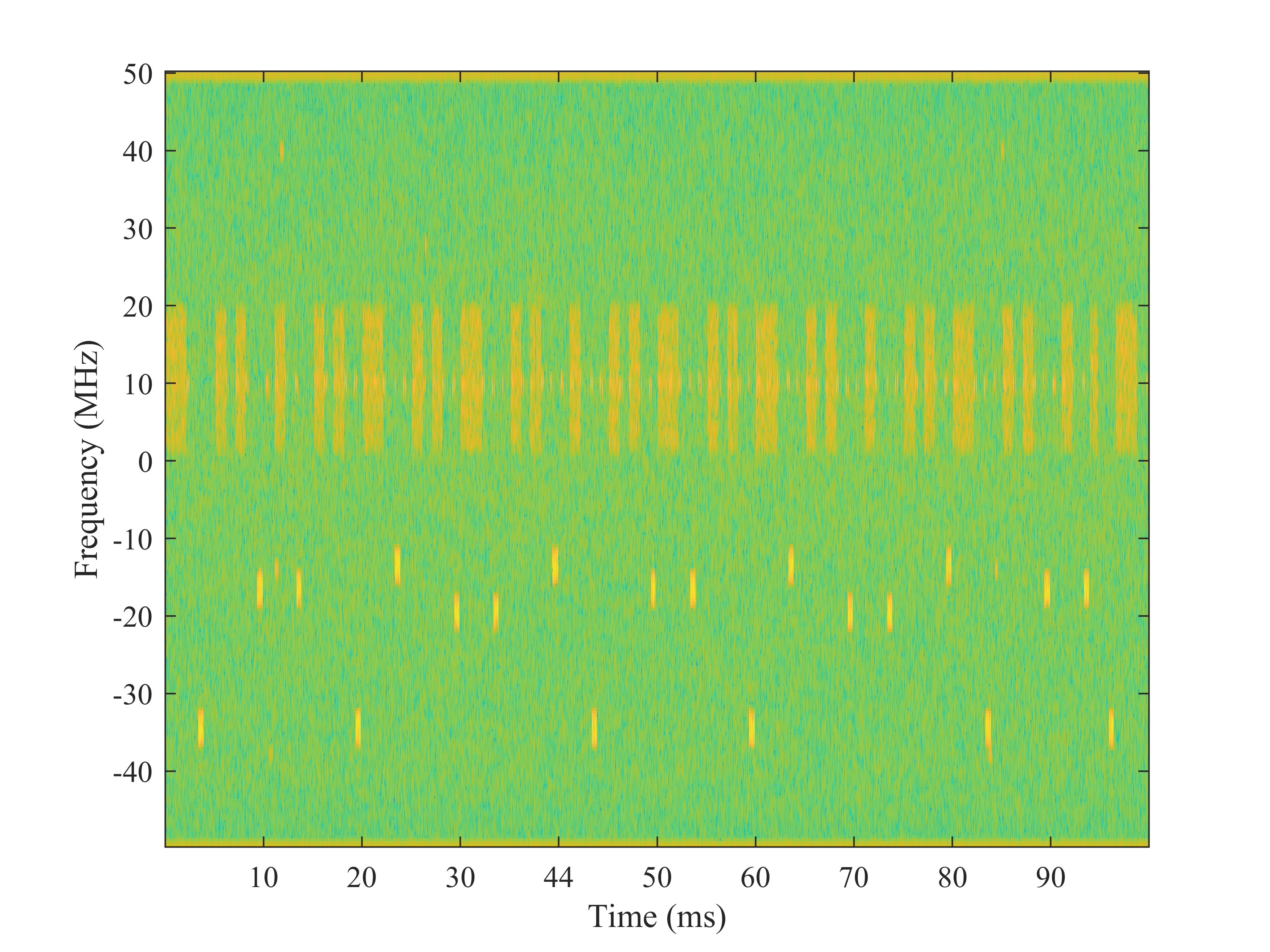}
			\label{tfi_7}}
		\hfil
		\subfloat[]{\includegraphics[width=1.7in]{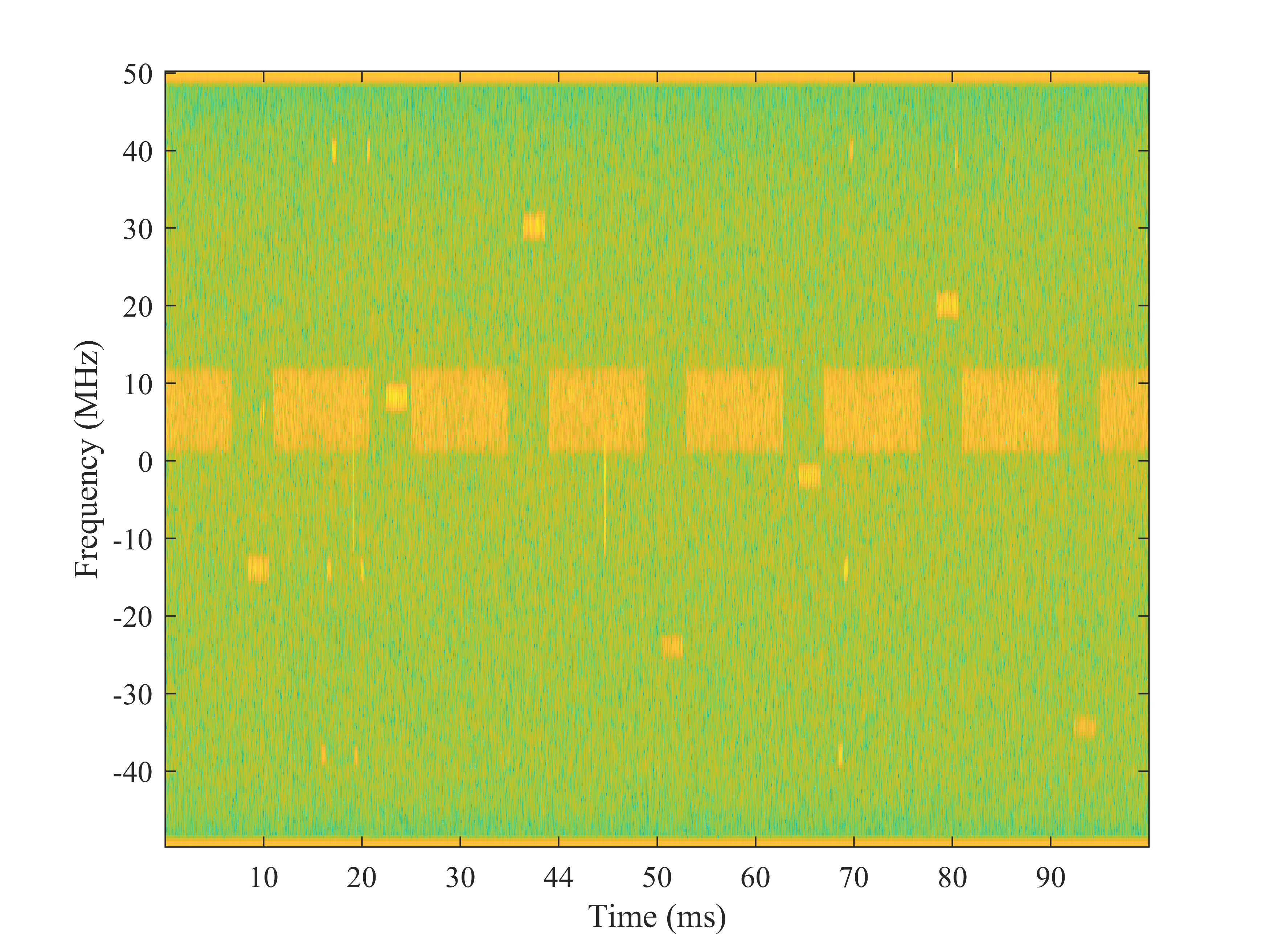}
			\label{tfi_8}}
		\hfil
		\subfloat[]{\includegraphics[width=1.7in]{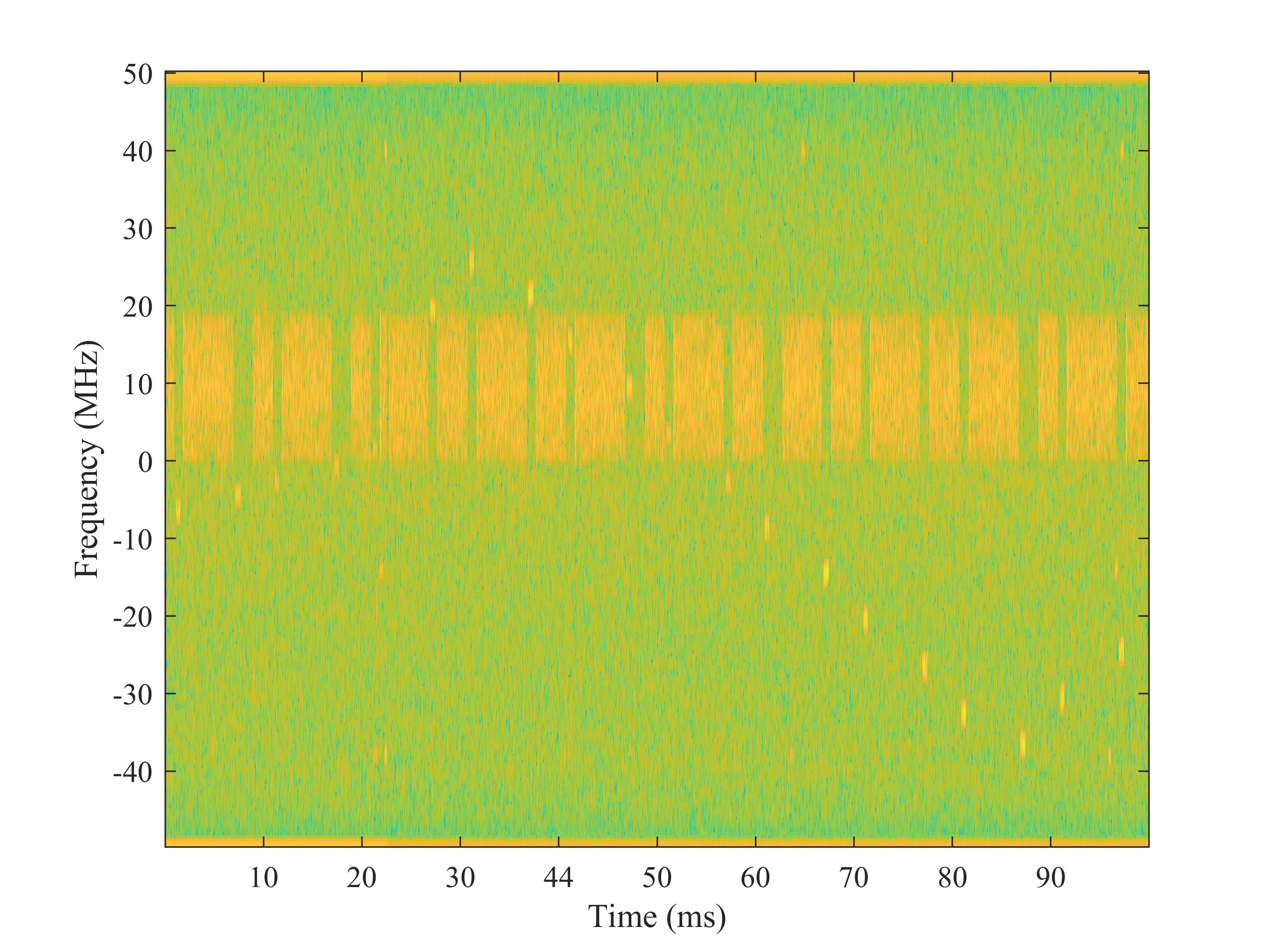}
			\label{tfi_9}}
		\hfil
		\subfloat[]{\includegraphics[width=1.7in]{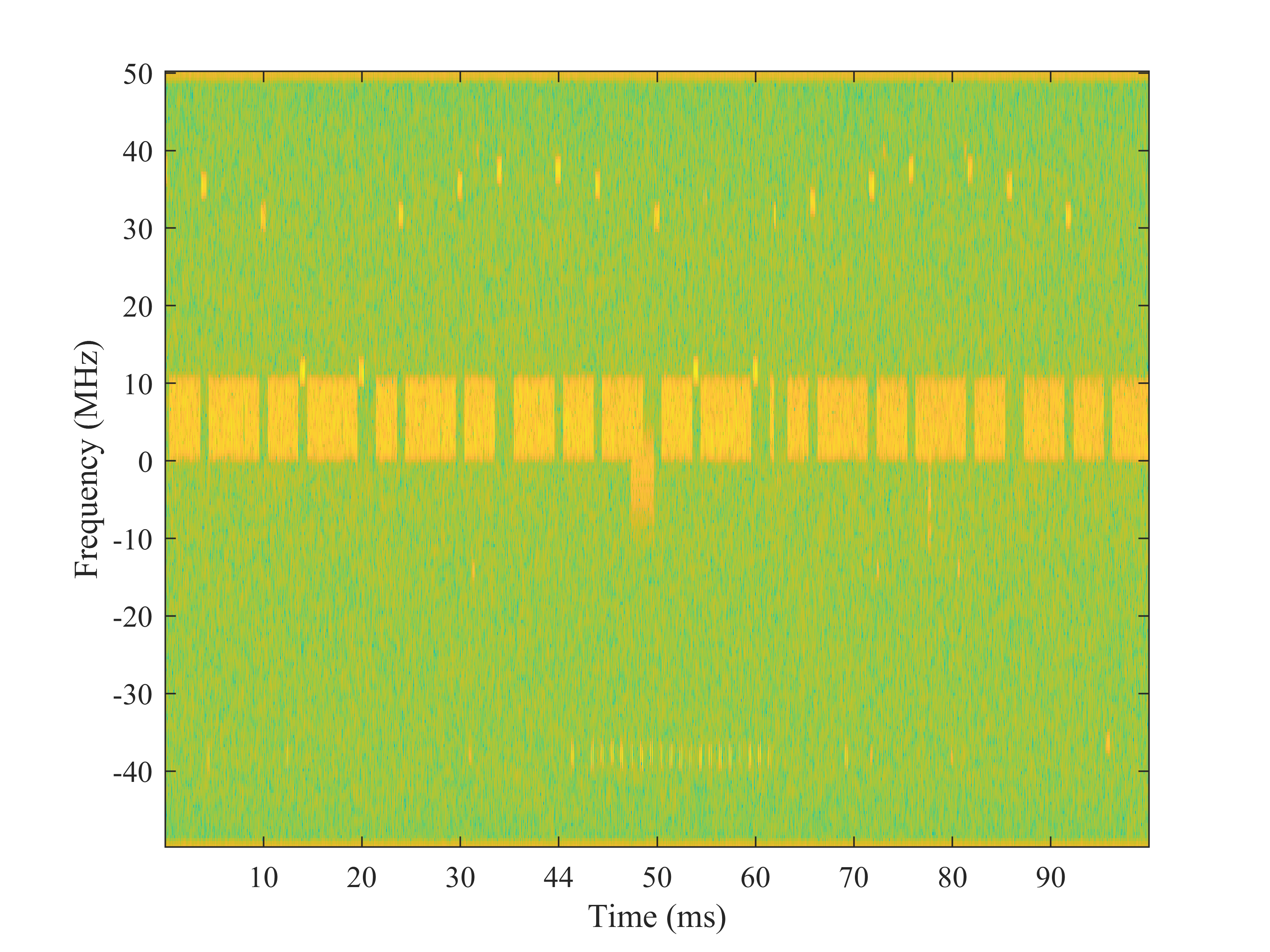}
			\label{tfi_10}}
		\hfil
		\subfloat[]{\includegraphics[width=1.7in]{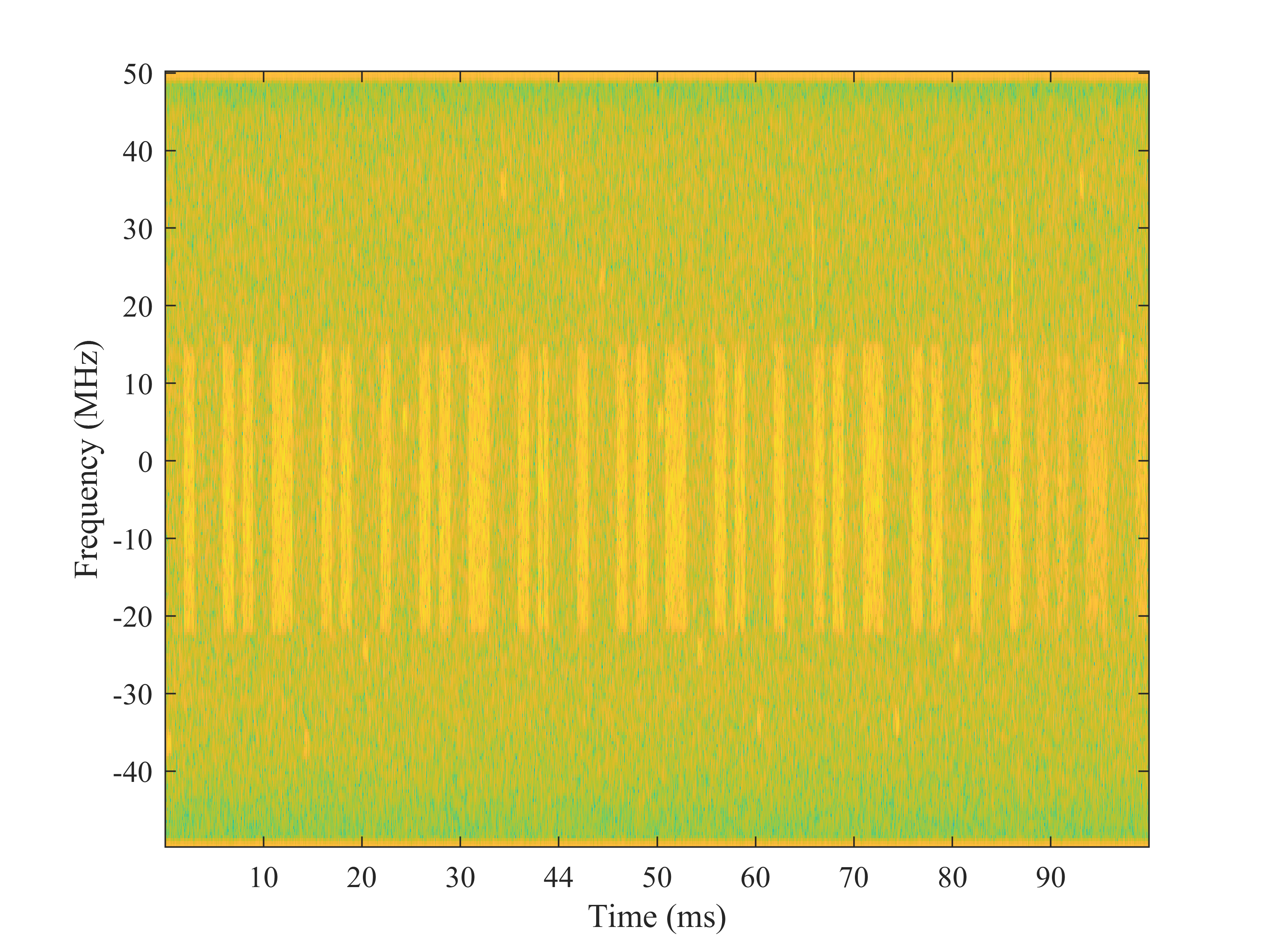}
			\label{tfi_11}}
		\hfil
		\subfloat[]{\includegraphics[width=1.7in]{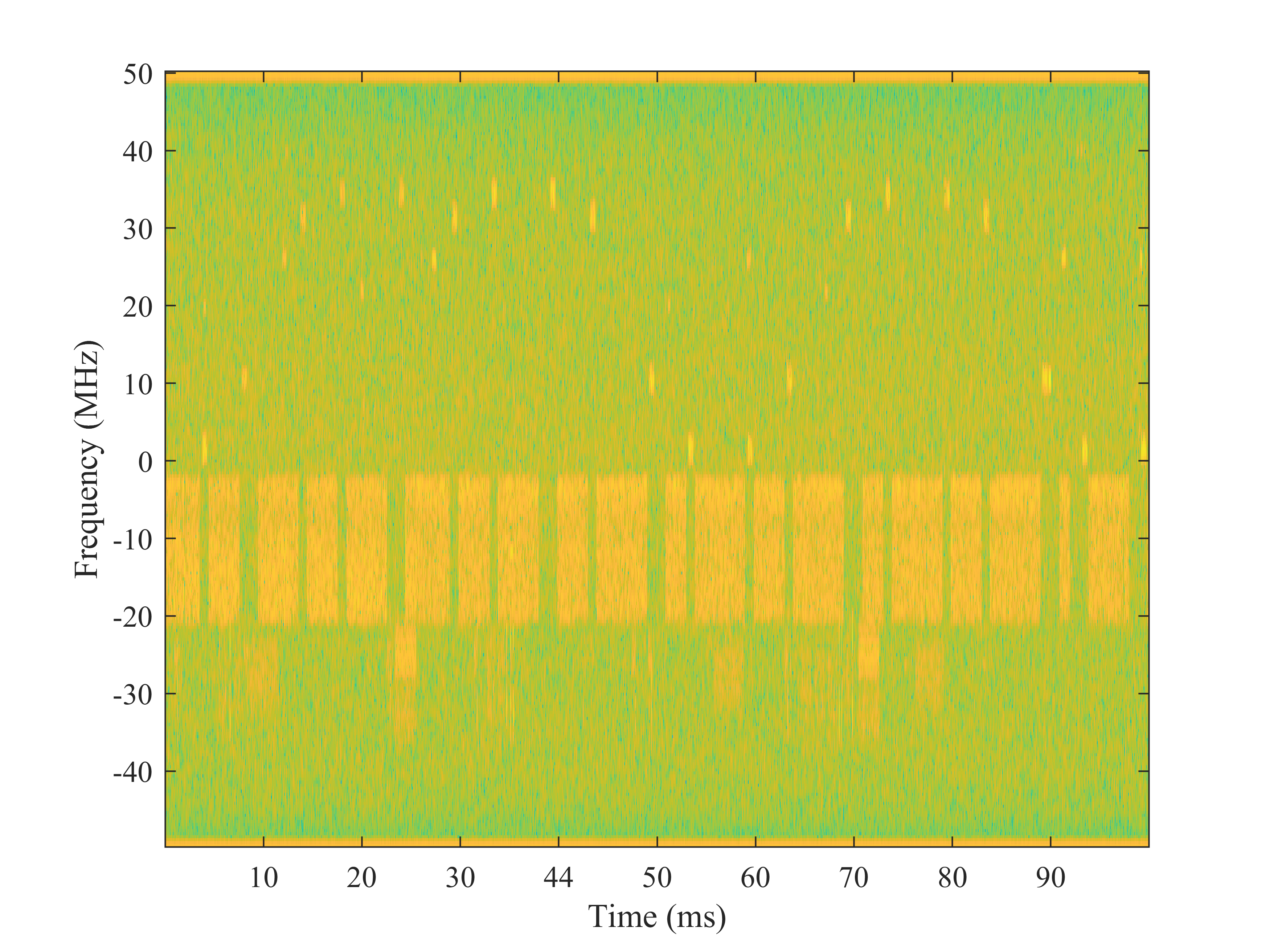}
			\label{tfi_12}}
		\hfil
		\subfloat[]{\includegraphics[width=1.7in]{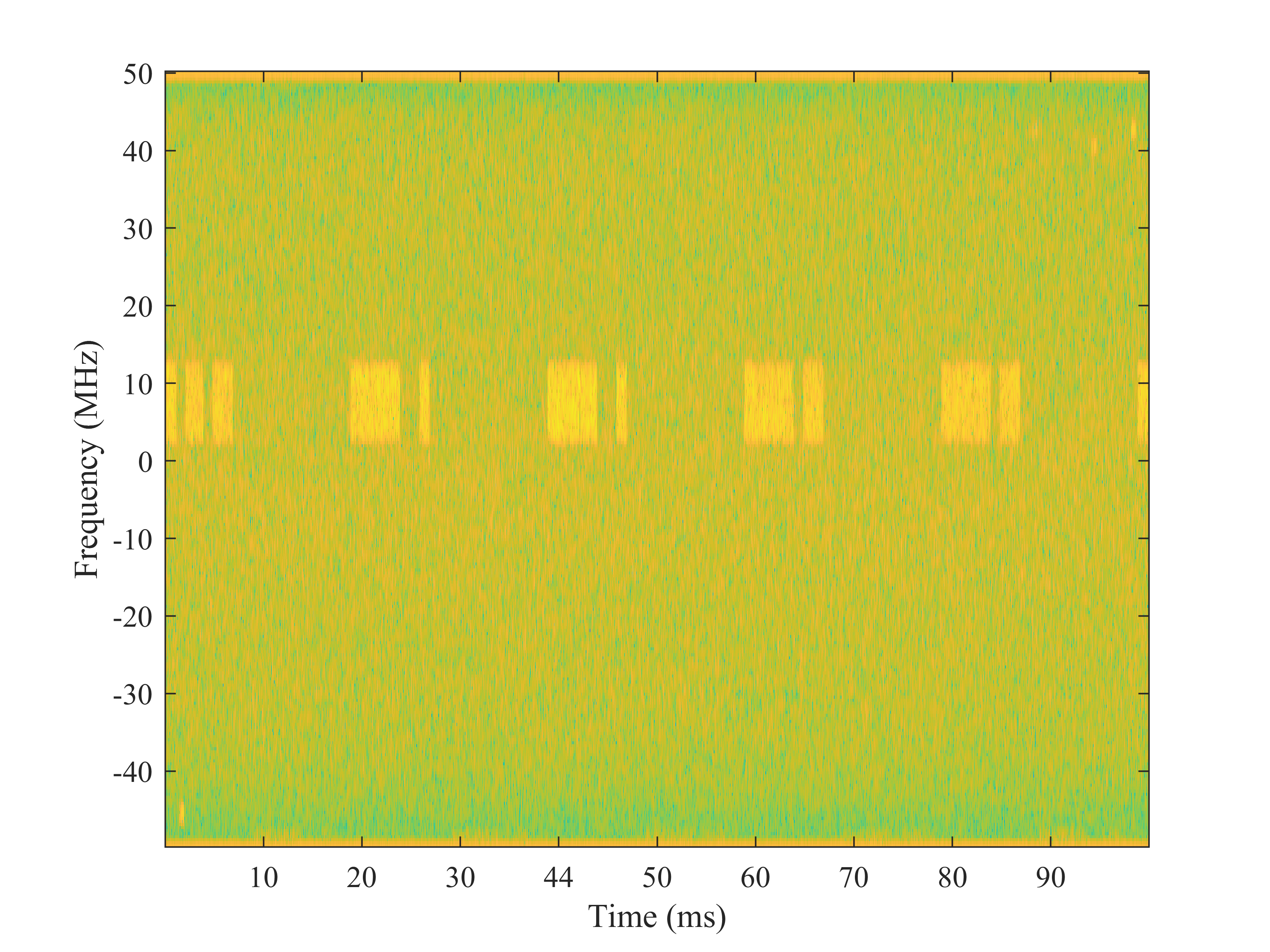}
			\label{tfi_13}}
		\hfil
		\subfloat[]{\includegraphics[width=1.7in]{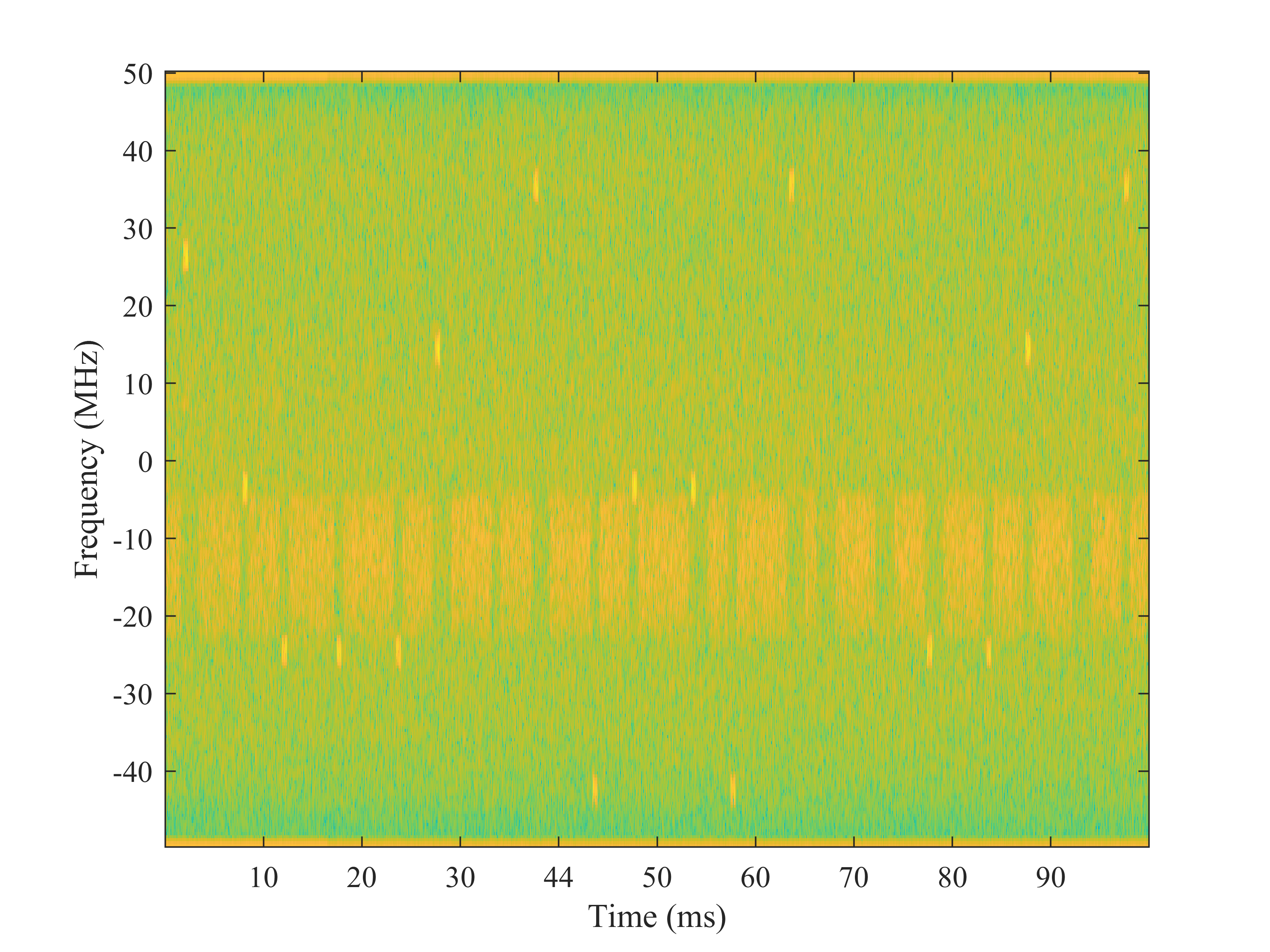}
			\label{tfi_14}}
		\hfil
		\subfloat[]{\includegraphics[width=1.7in]{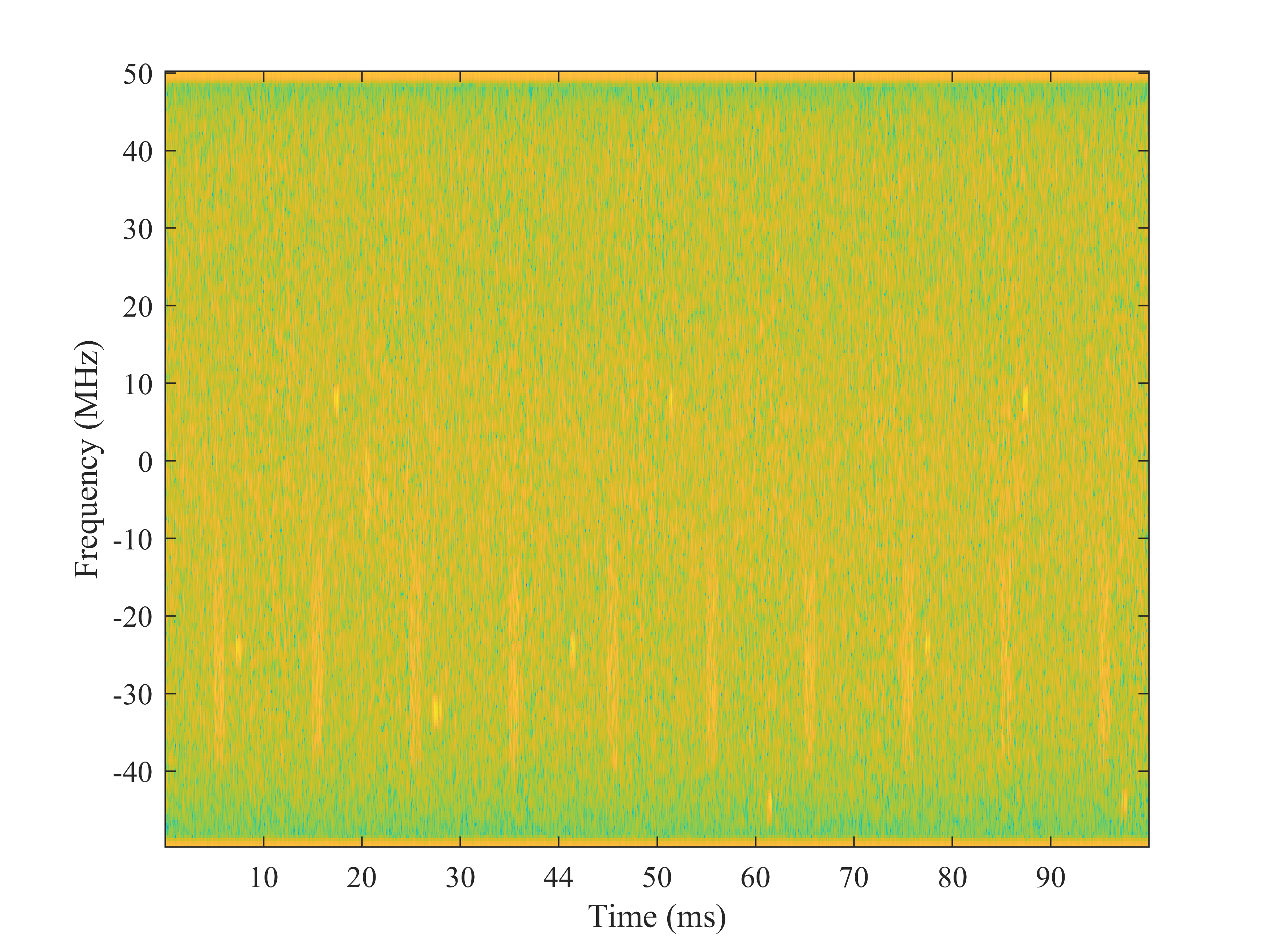}
			\label{tfi_15}}
		\hfil
		\subfloat[]{\includegraphics[width=1.7in]{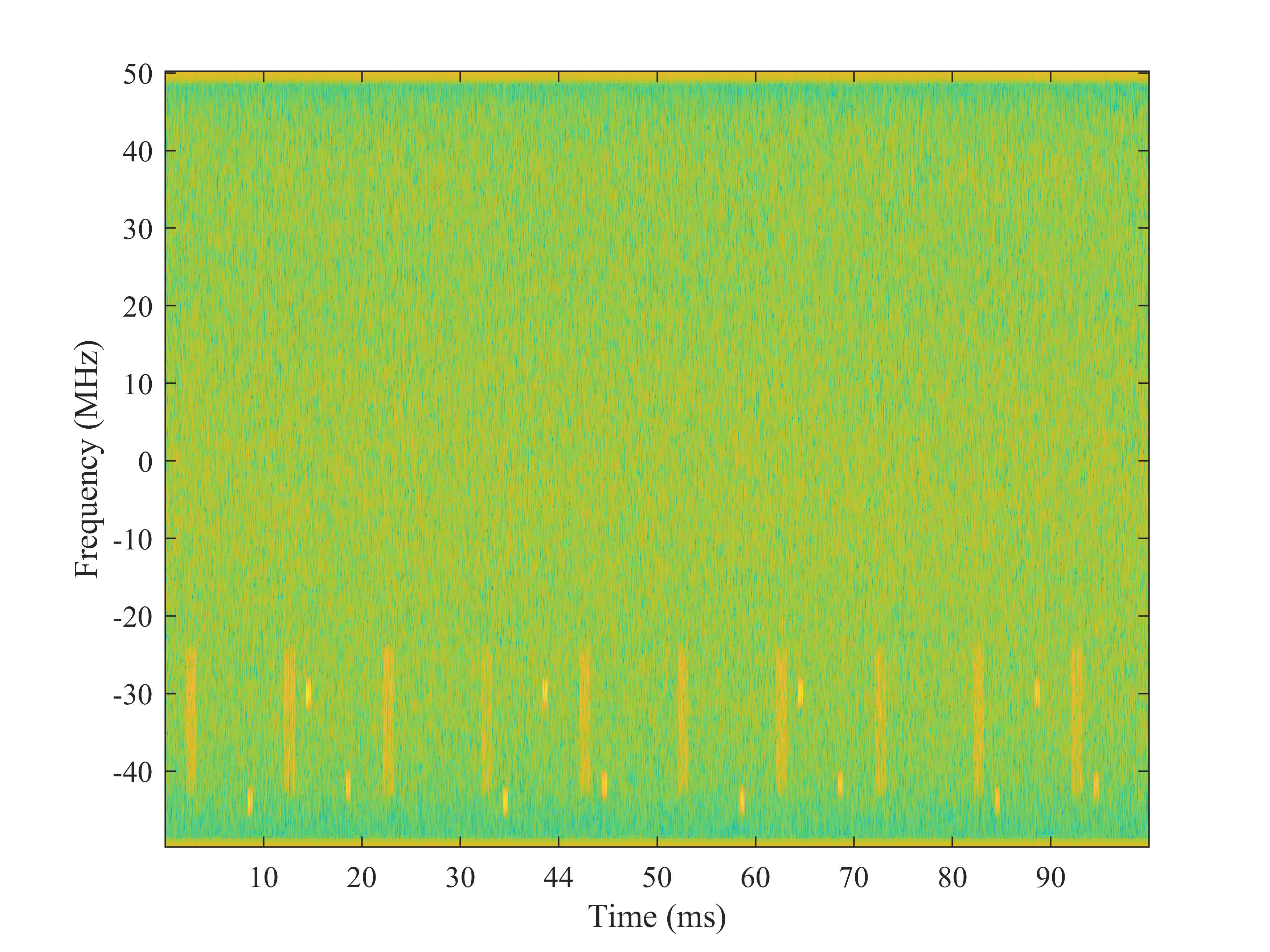}
			\label{tfi_16}}
		\caption{TFI of RF signals with \(\tiny{10^7}\) samples at SNR = 15 dB. (a) T0000D00S00. (b) T0001D00S00. (c) T0010D00S00. (d) T0011D00S00. (e) T0100D00S00. (f) T0101D00S00. (g) T0110D00S00. (h) T0111D00S00. (i) T1000D00S00. (j) T1001D00S00. (k) T1010D00S00. (l) T1011D00S00. (m) T1100D00S00. (n) T1101D00S00. (o) T1110D00S00. (p) T1111D00S00.}
		\label{Fig_tfi}
	\end{figure*}
	
	\item{\textit{ZC Sequence Feature}}:
	To extract ZC sequence feature via cross-correlation with \(x(l)\), \({z_r}(v)\) will be up-sampled with factor \(\frac{f_s}{B}\) to obtain \(y(v)\) with length \(L_a\). The cross-correlation results for \(x(l)\) can be obtained by
	\begin{equation}
		\label{Eq3}
			\gamma \left( m \right)=\frac{\left| \sum\limits_{k=1}^{{{L}_{a}}}{{{y}_{1}}\left( k \right)x_{1}^{*}\left( k+m \right)} \right|}{{{L}_{a}}\sqrt{\text{Var1}\times \text{Var2}}},m=0,1,\ldots ,L-{{L}_{a}}-1,  \\
	\end{equation}
	where the zero-mean value sequences \({{y}_{1}}(m)=y(m)-\frac{1}{{{L}_{a}}}\sum\nolimits_{m=1}^{{{L}_{a}}}{y(m)}\) and \({{x}_{1}}(k+m)=x(k+m)-\frac{1}{{{L}_{a}}}\sum\nolimits_{m=1}^{{{L}_{a}}}{x(k+m)}\). \(\text{Var1}\) and \(\text{Var2}\) aim to normalize the cross-correlation results, which denote the variance of \(\sum\nolimits_{m=1}^{{{L}_{a}}}{y(m)}\) and \(\sum\nolimits_{m=1}^{{{L}_{a}}}{x(k+m)}\), respectively.
	
	However, considering the complexity of \textcolor{blue}{(\ref{Eq3})} is \(\mathcal{O}(L_aL)\), and ZC sequences typically exist in every signal frame, the trade-off between RID latency and accuracy should be carefully considered. Based on the prior work\textcolor{blue}{\cite{reftccn}}, 20 non-overlapping segments of length \(5\times10^4\) are randomly selected from \(x(l)\), and concatenated to form a new complex sequence of length \(10^6\). \(\gamma(m)\) between the new sequence and  the locally generated ZC sequence can be computed by \textcolor{blue}{(\ref{Eq3})}, which will be reshaped to compress redundant information and reduce the number of weighted parameters for neural network. Each row of the ZC sequence feature \(\mathbf{R}\) consists of the cross-correlation results with different ZC sequences, and the \(i-\)th row of \(\mathbf{R}\) can be expressed as
	\begin{equation}
		\label{Eq4}
		\mathbf{R_i}\!=\!
		\left[ \begin{matrix}
			\!\max\! \left[ \begin{matrix}
				\gamma ( 1 )  \\
				\gamma ( 2 )  \\
				\vdots   \\
				\gamma ( U )  \\
			\end{matrix} \right] \!&\! \max\! \left[ \begin{matrix}
				\gamma( U\!+\!1 )  \\
				\gamma( U\!+\!2 )  \\
				\vdots   \\
				\gamma( 2U )  \\
			\end{matrix} \right] \!&\! \cdots  \!&\! \max\! \left[ \begin{matrix}
				\gamma ( U(V\!-\!1)\!+\!1 )  \\
				\gamma ( U(V\!-\!1)\!+\!2 )  \\
				\vdots   \\
				\gamma ( UV )  \\
			\end{matrix} \right]  \\
		\end{matrix} \right],
	\end{equation}
	where \(U\) and \(V\) are hyper-parameters used to control the ZC sequence feature size, which are set as \(10^2\) and \(10^4\) in this paper. As illustrated in Fig. \ref{Fig_zc}, the ZC sequence feature reveals that several types of drone employing ZC sequences exhibit prominent and regulated cross-correlation peaks in \(\mathbf{R}\), whereas drones not utilizing ZC sequences and background signals do not display such characteristics. The ZC sequence feature maintains robust correlation properties even under interference and low SNR conditions, while drones without ZC sequences can be effectively distinguished using TFI features. The
	overall computational complexity of feature generation is \(\mathcal{O}(L\log U + 4TF + 9L_aUV)\) for DroneRFa dataset.
	\begin{figure*}[!t]
		\centering
		\subfloat[]{\includegraphics[width=1.7in]{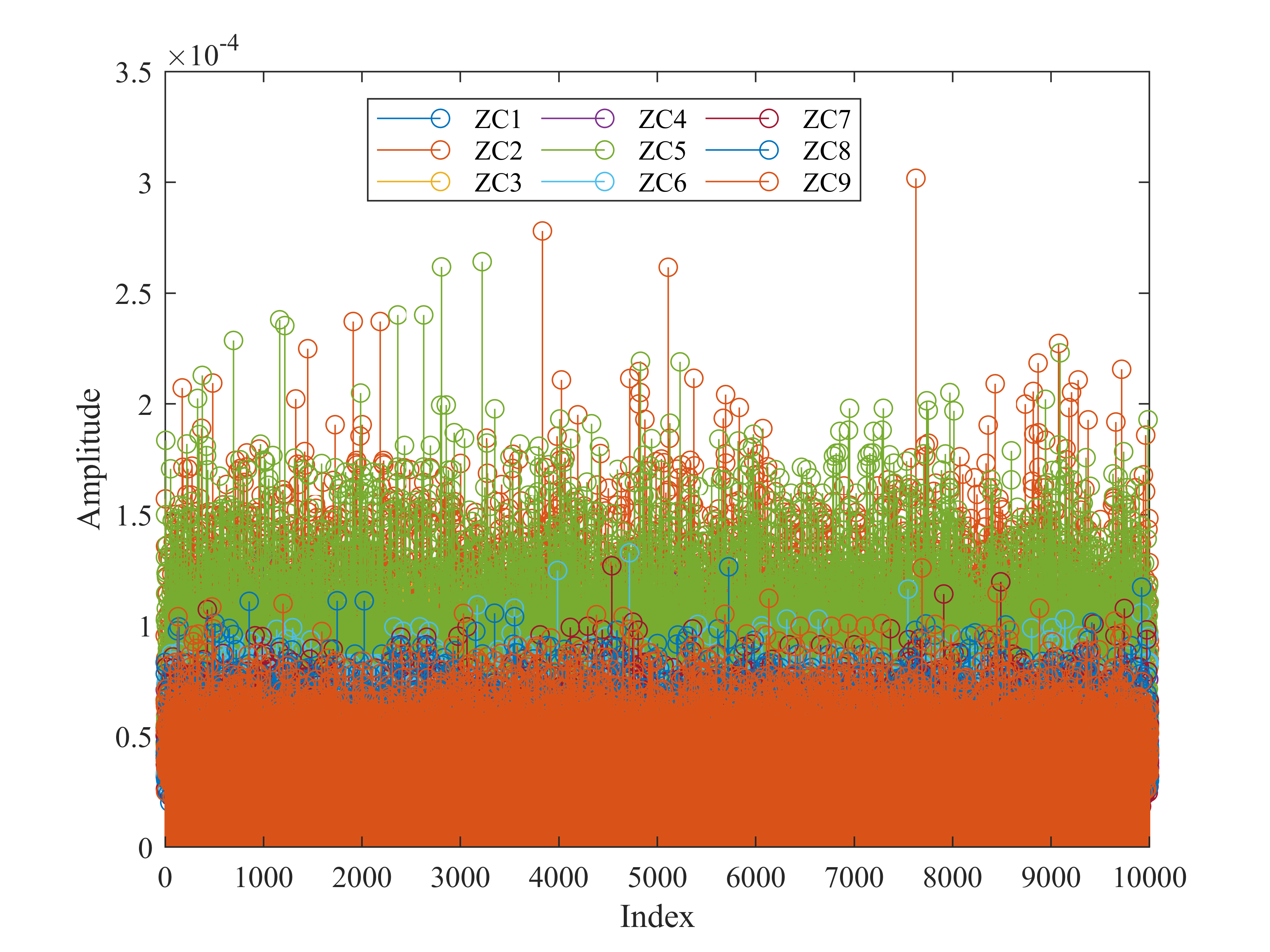}
			\label{zc_1}}
		\hfil
		\subfloat[]{\includegraphics[width=1.7in]{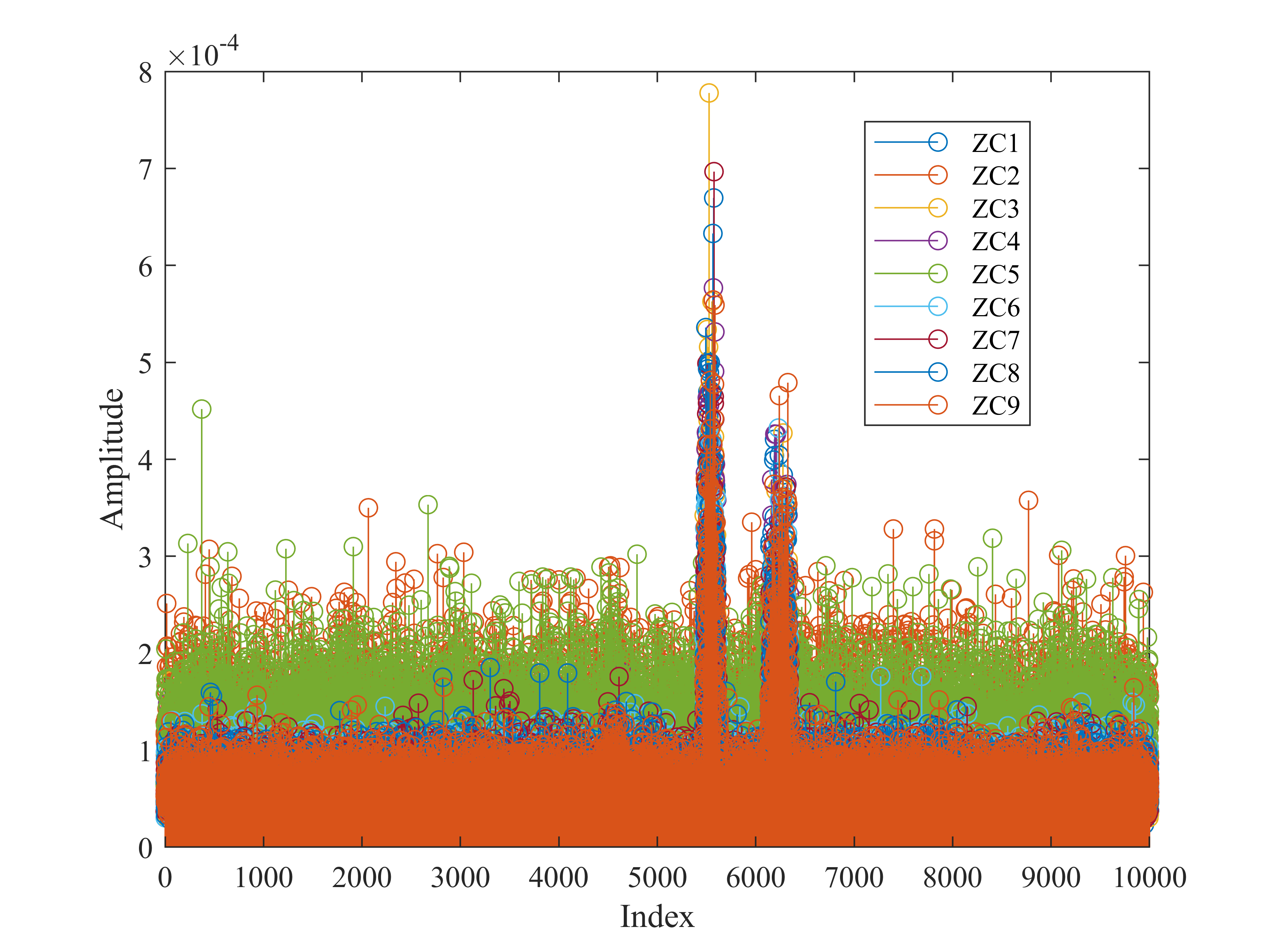}
			\label{zc_2}}
		\hfil
		\subfloat[]{\includegraphics[width=1.7in]{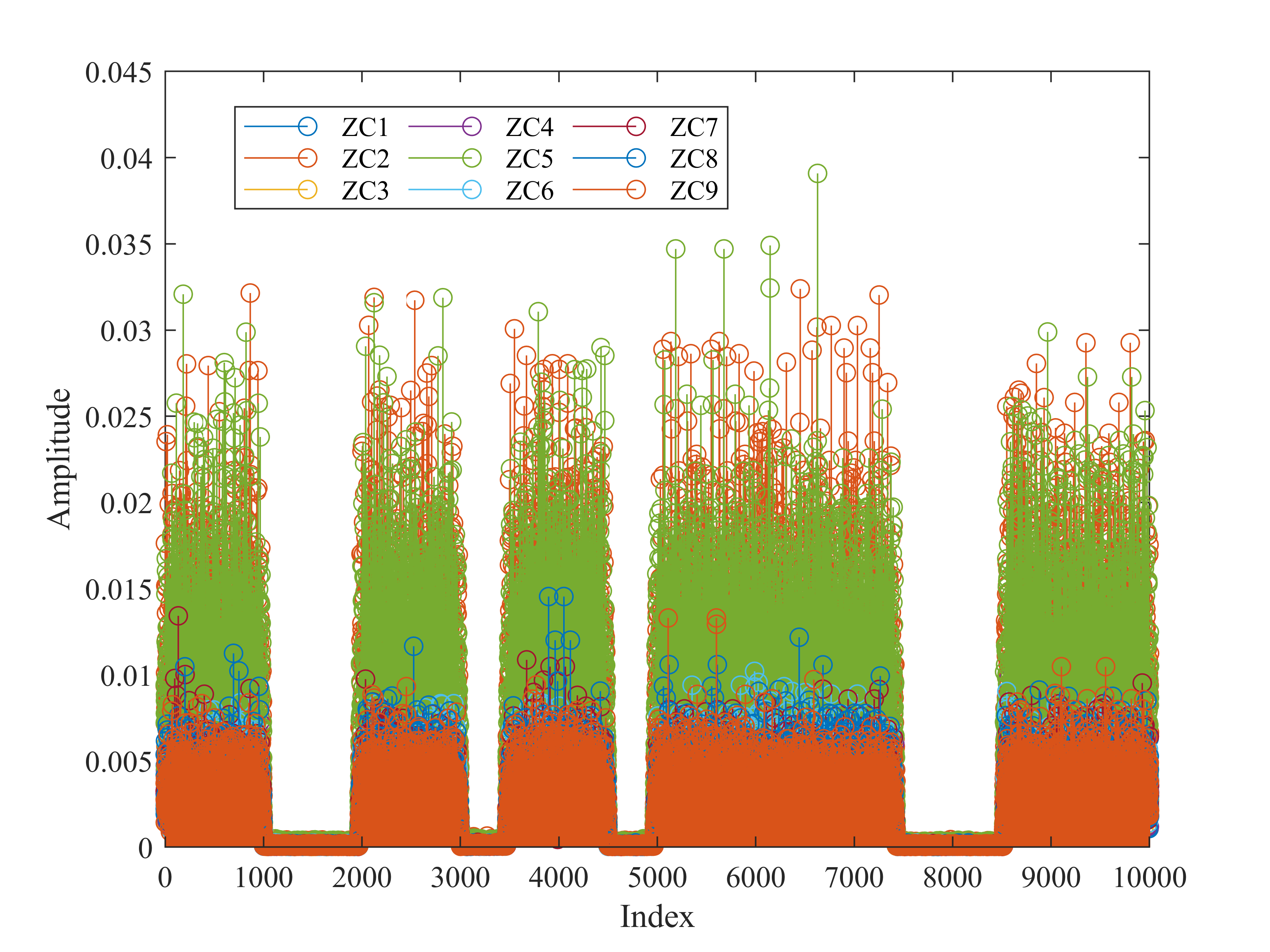}
			\label{zc_3}}
		\hfil
		\subfloat[]{\includegraphics[width=1.7in]{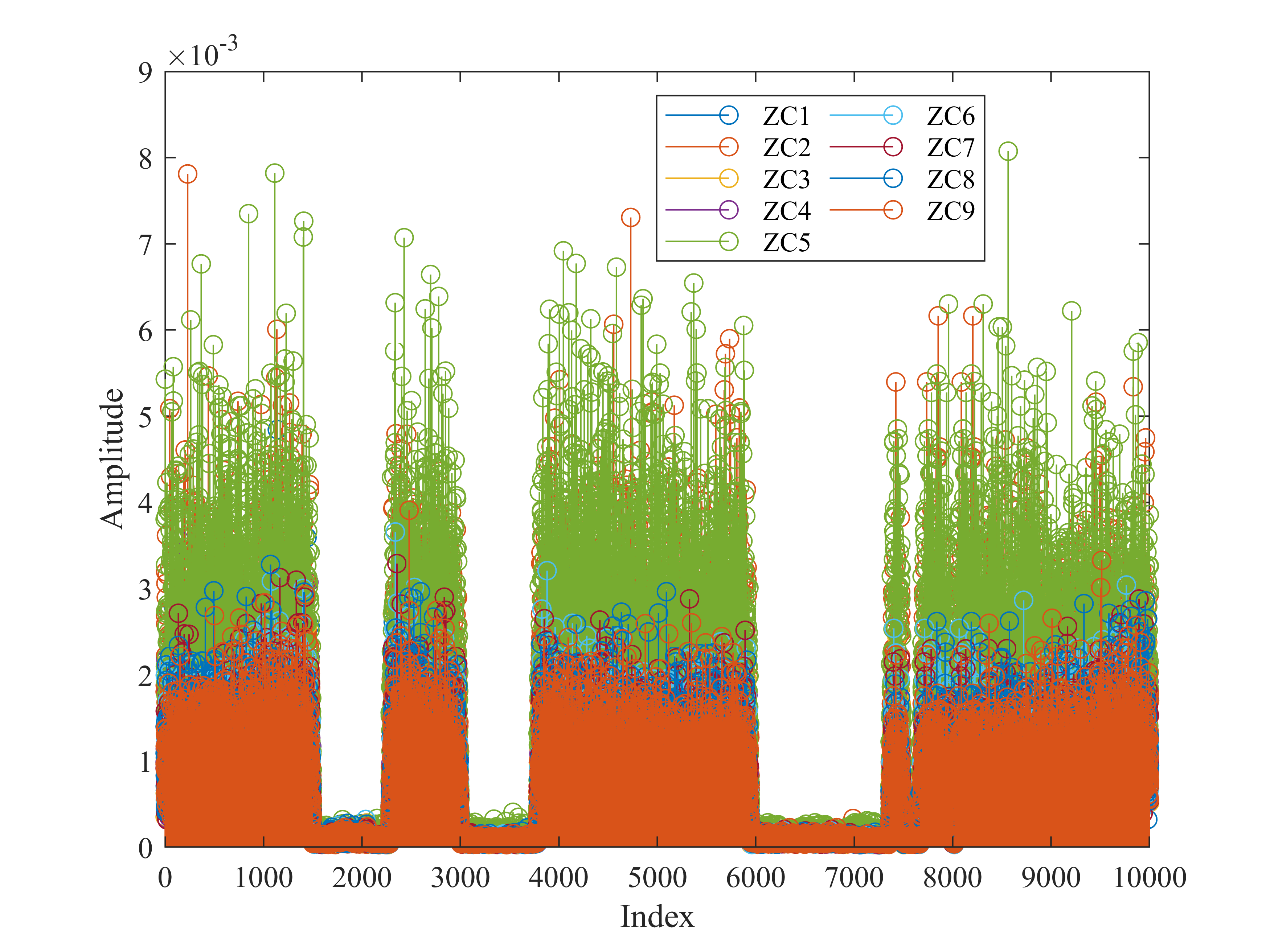}
			\label{zc_4}}
		\hfil
		\subfloat[]{\includegraphics[width=1.7in]{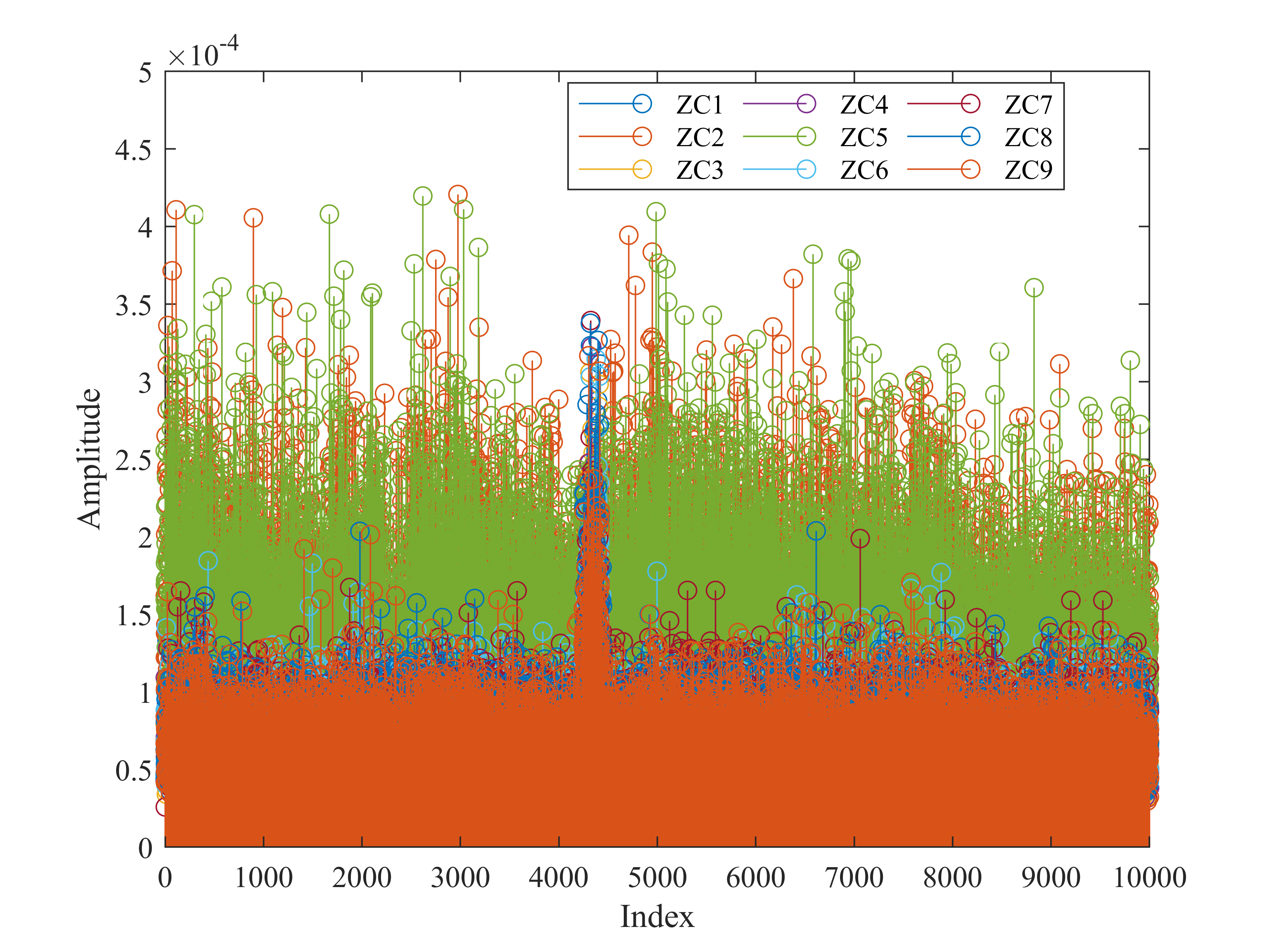}
			\label{zc_5}}
		\hfil
		\subfloat[]{\includegraphics[width=1.7in]{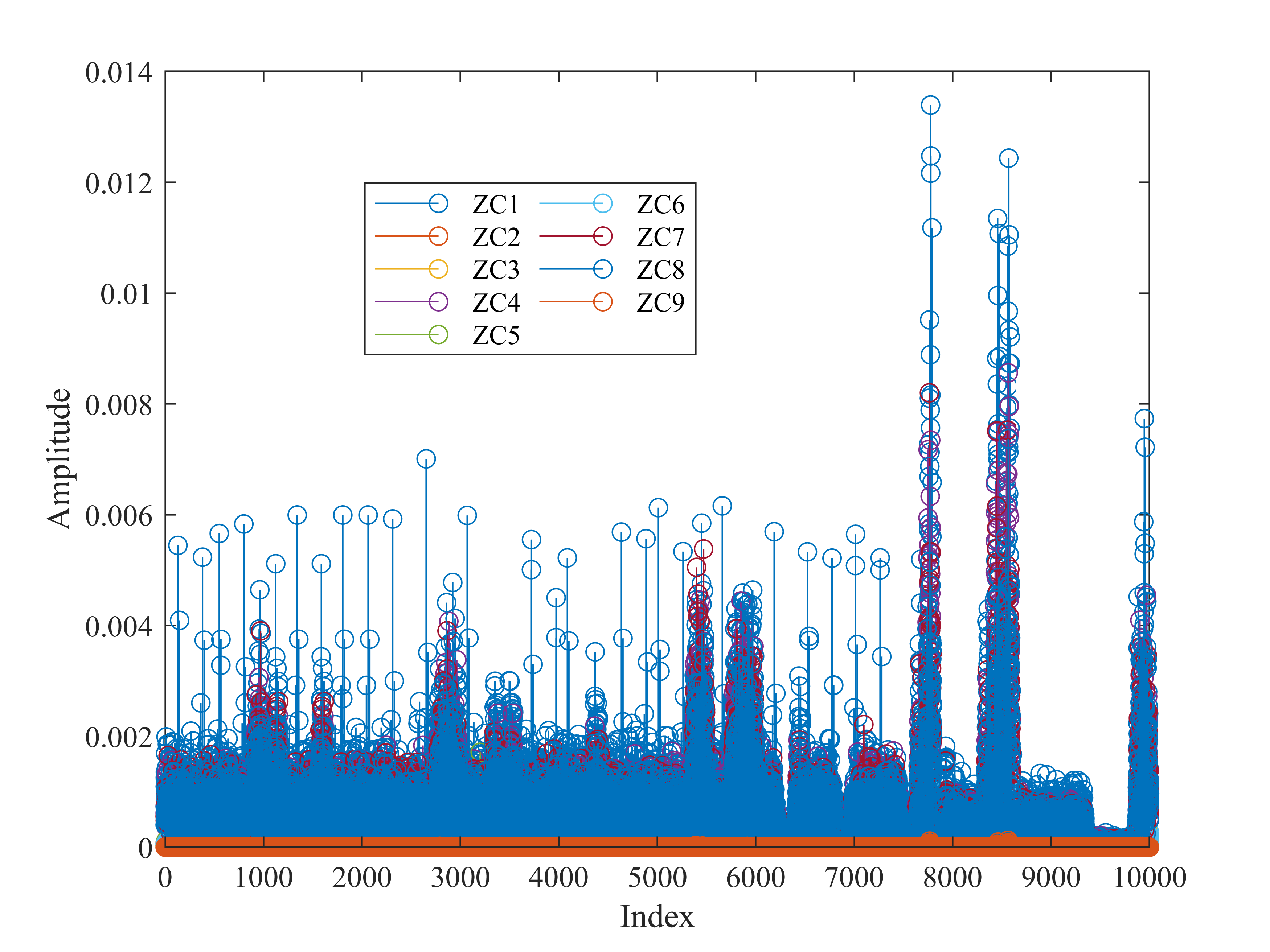}
			\label{zc_6}}
		\hfil
		\subfloat[]{\includegraphics[width=1.7in]{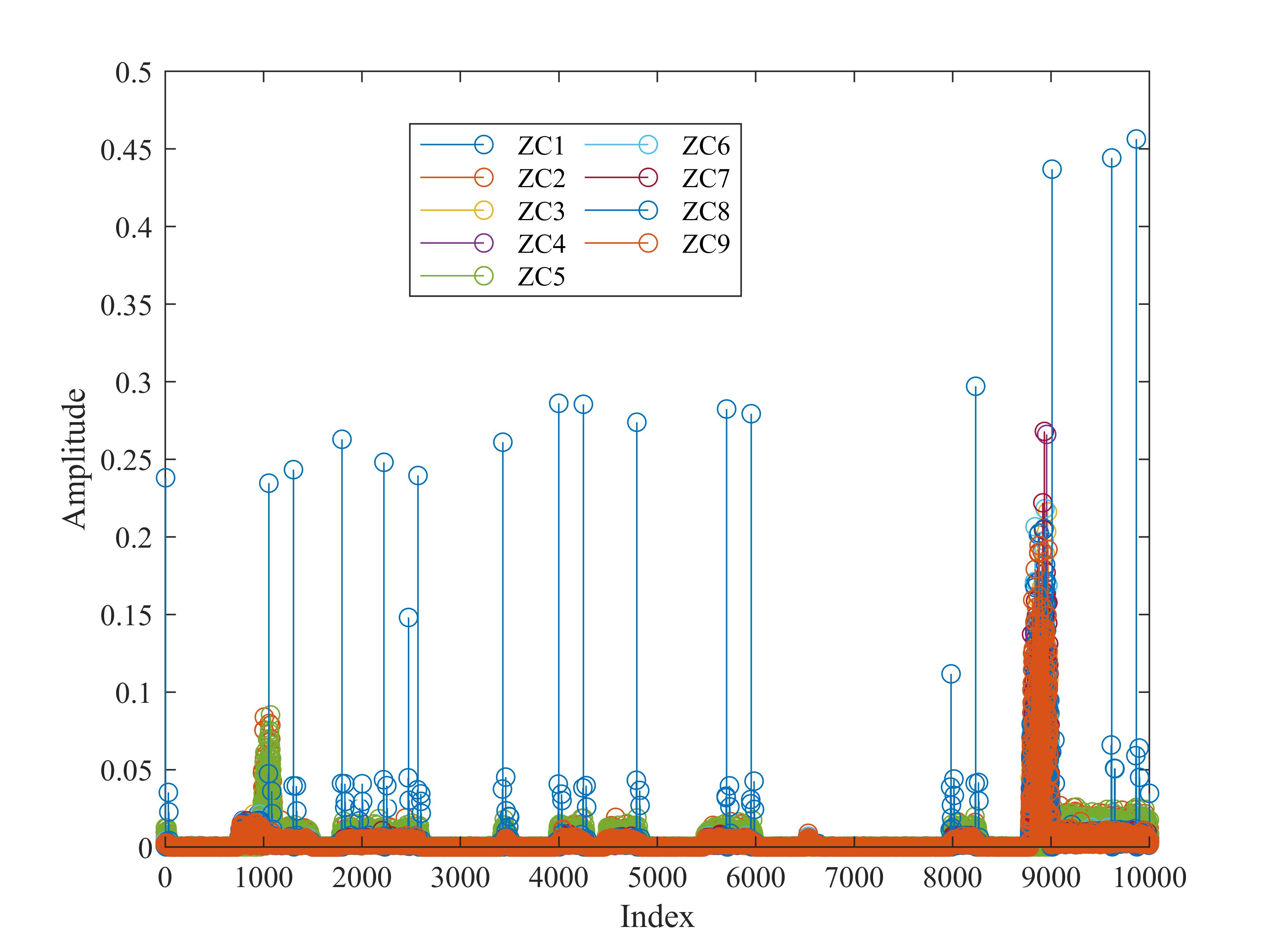}
			\label{zc_7}}
		\hfil
		\subfloat[]{\includegraphics[width=1.7in]{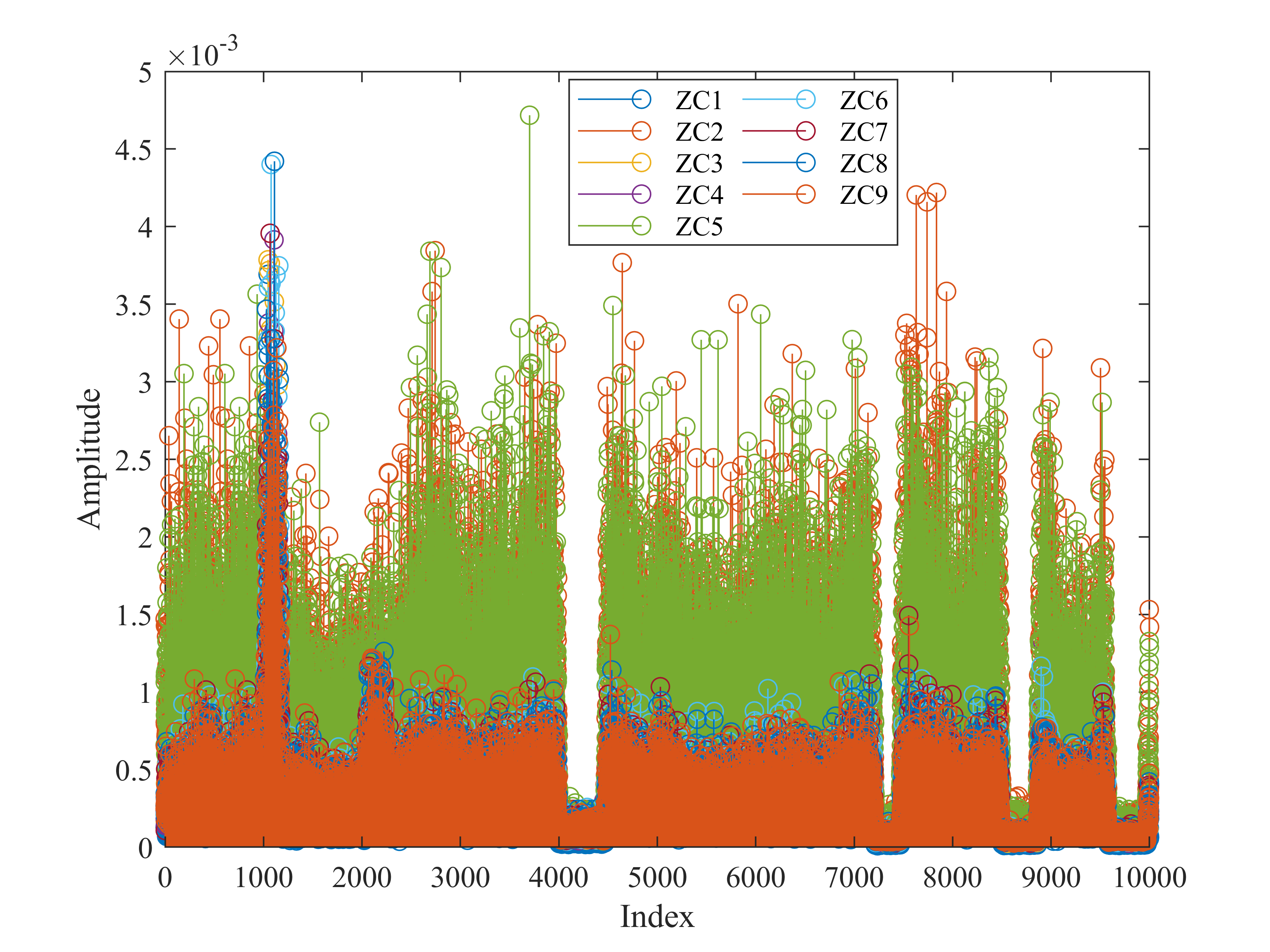}
			\label{zc_8}}
		\hfil
		\subfloat[]{\includegraphics[width=1.7in]{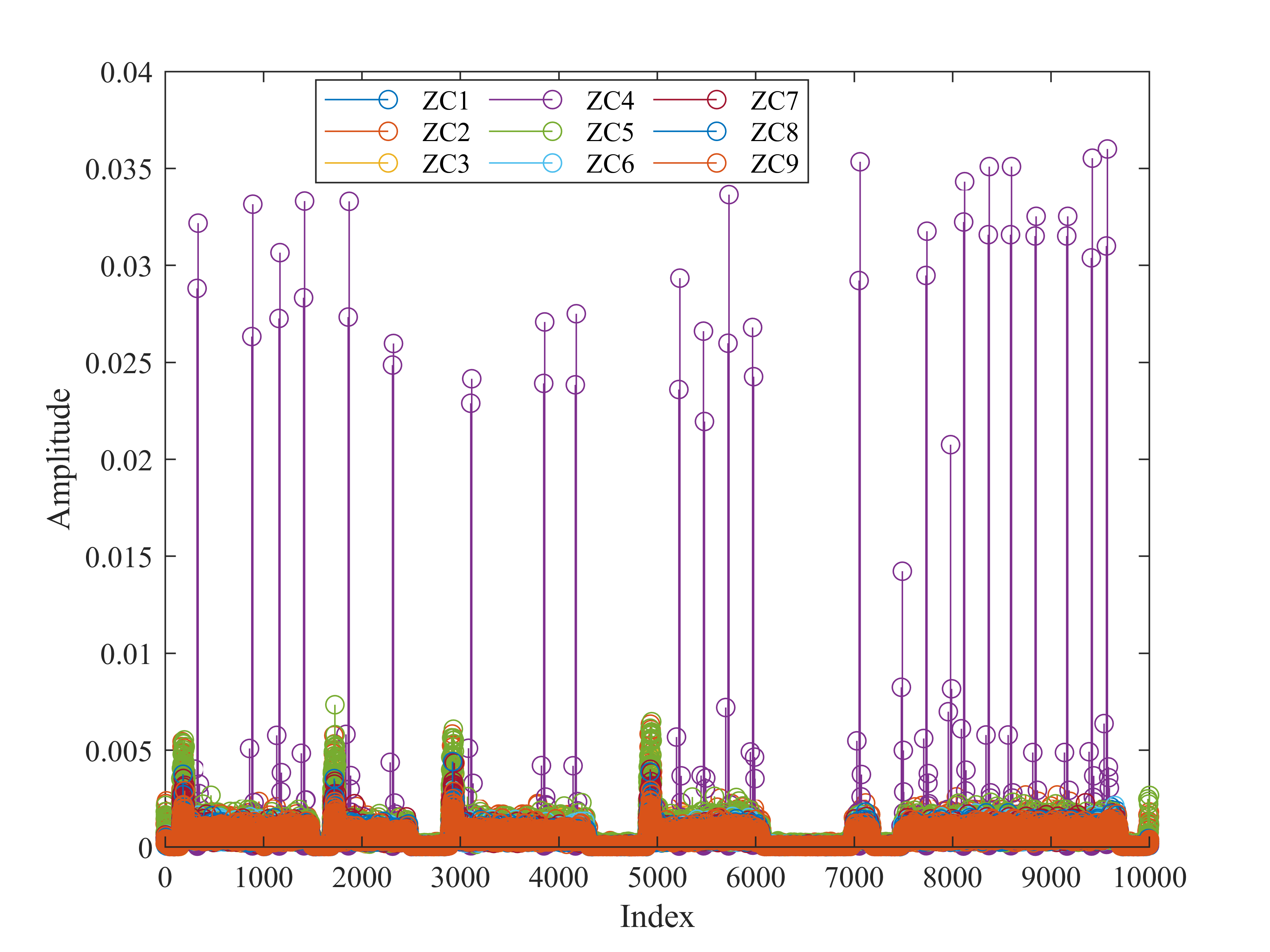}
			\label{zc_9}}
		\hfil
		\subfloat[]{\includegraphics[width=1.7in]{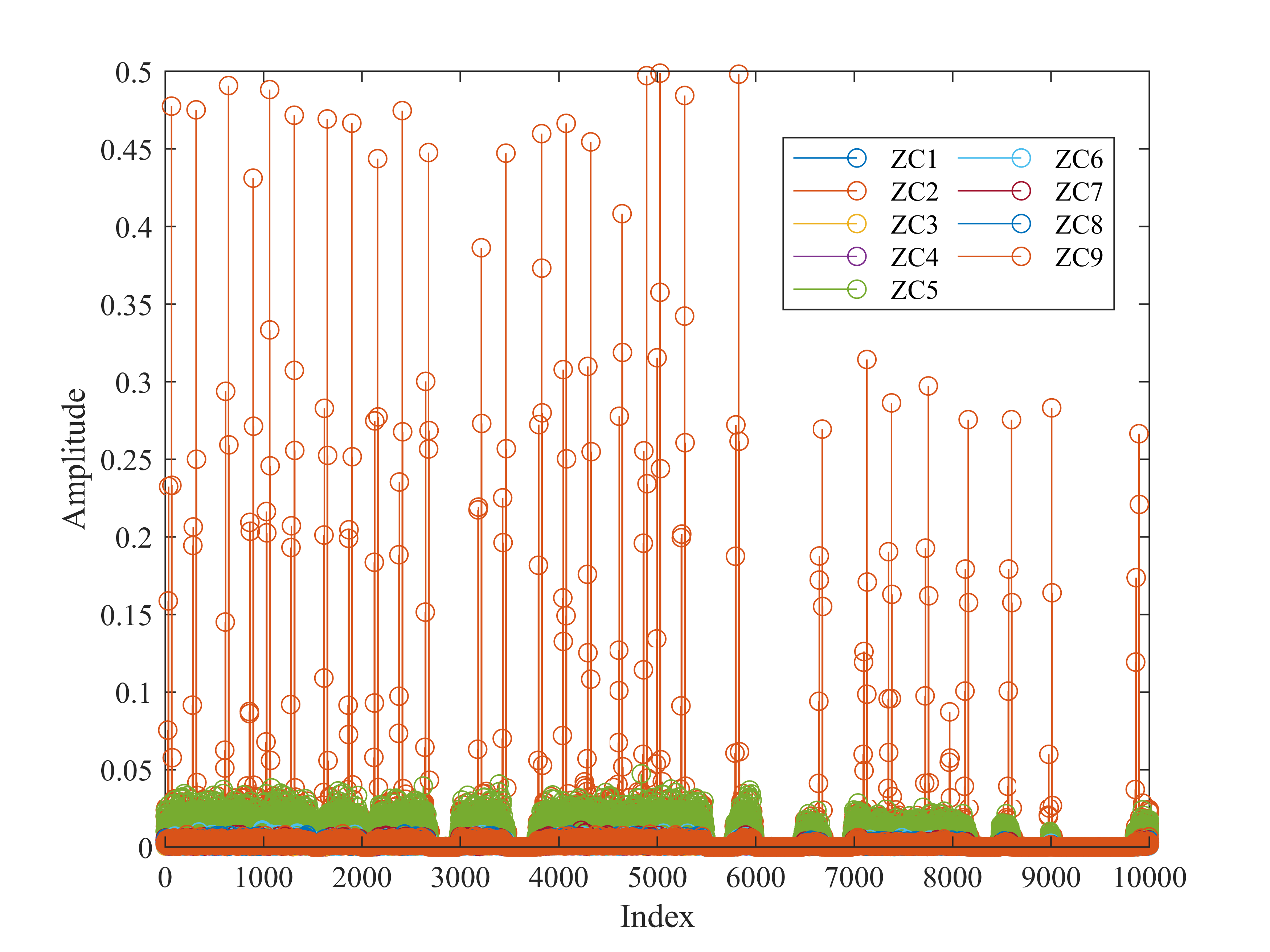}
			\label{zc_10}}
		\hfil
		\subfloat[]{\includegraphics[width=1.7in]{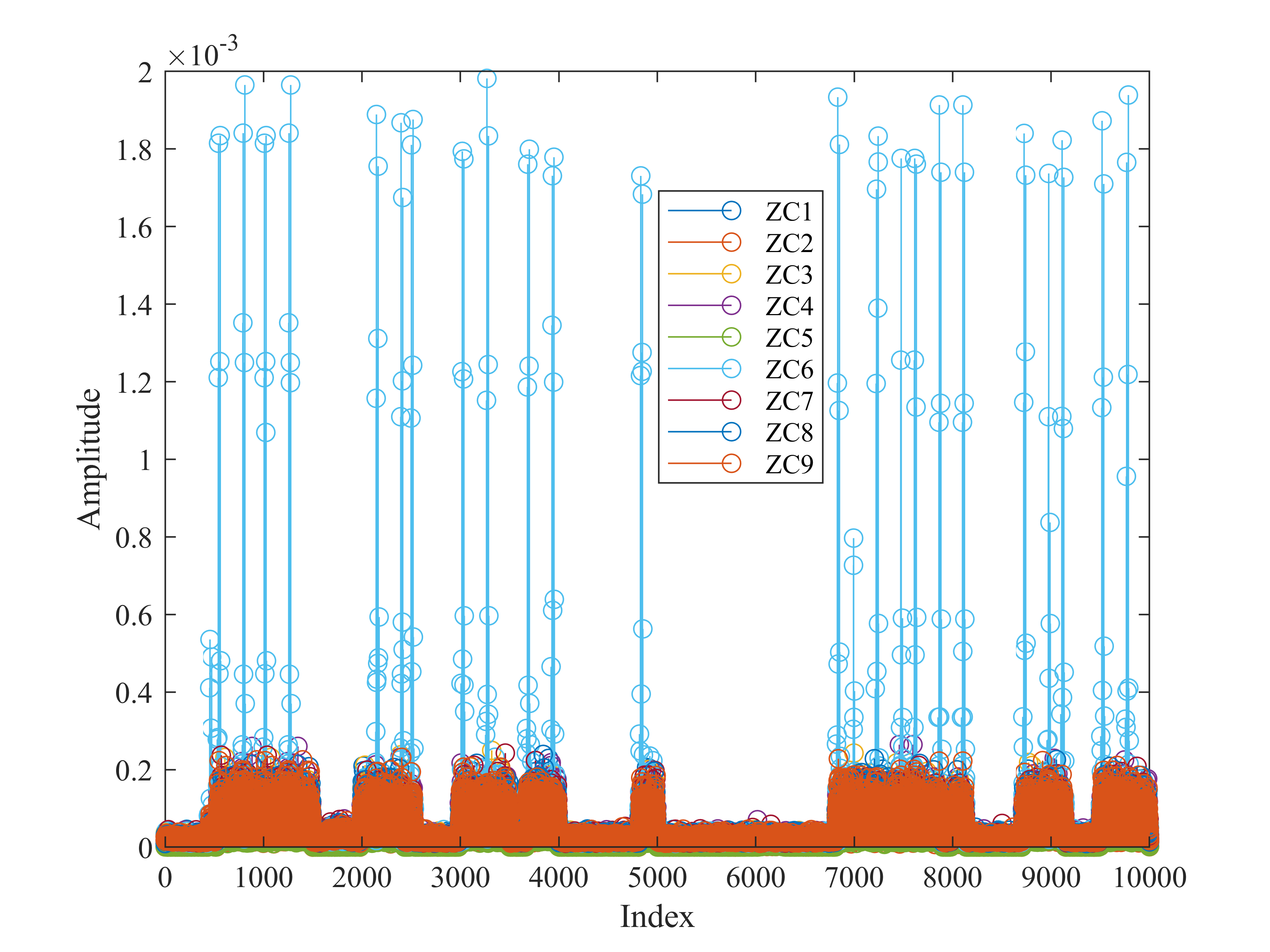}
			\label{zc_11}}
		\hfil
		\subfloat[]{\includegraphics[width=1.7in]{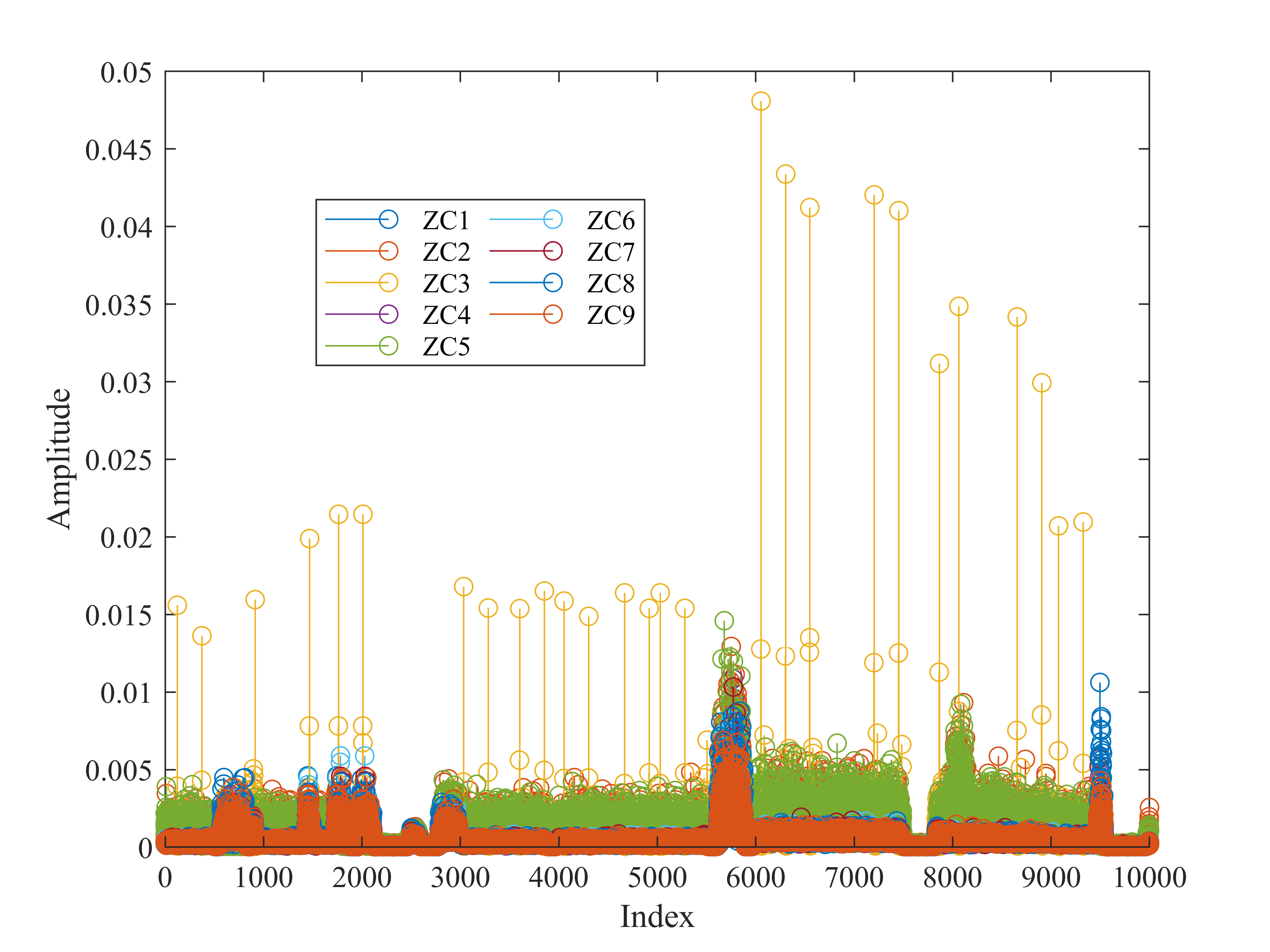}
			\label{zc_12}}
		\hfil
		\subfloat[]{\includegraphics[width=1.7in]{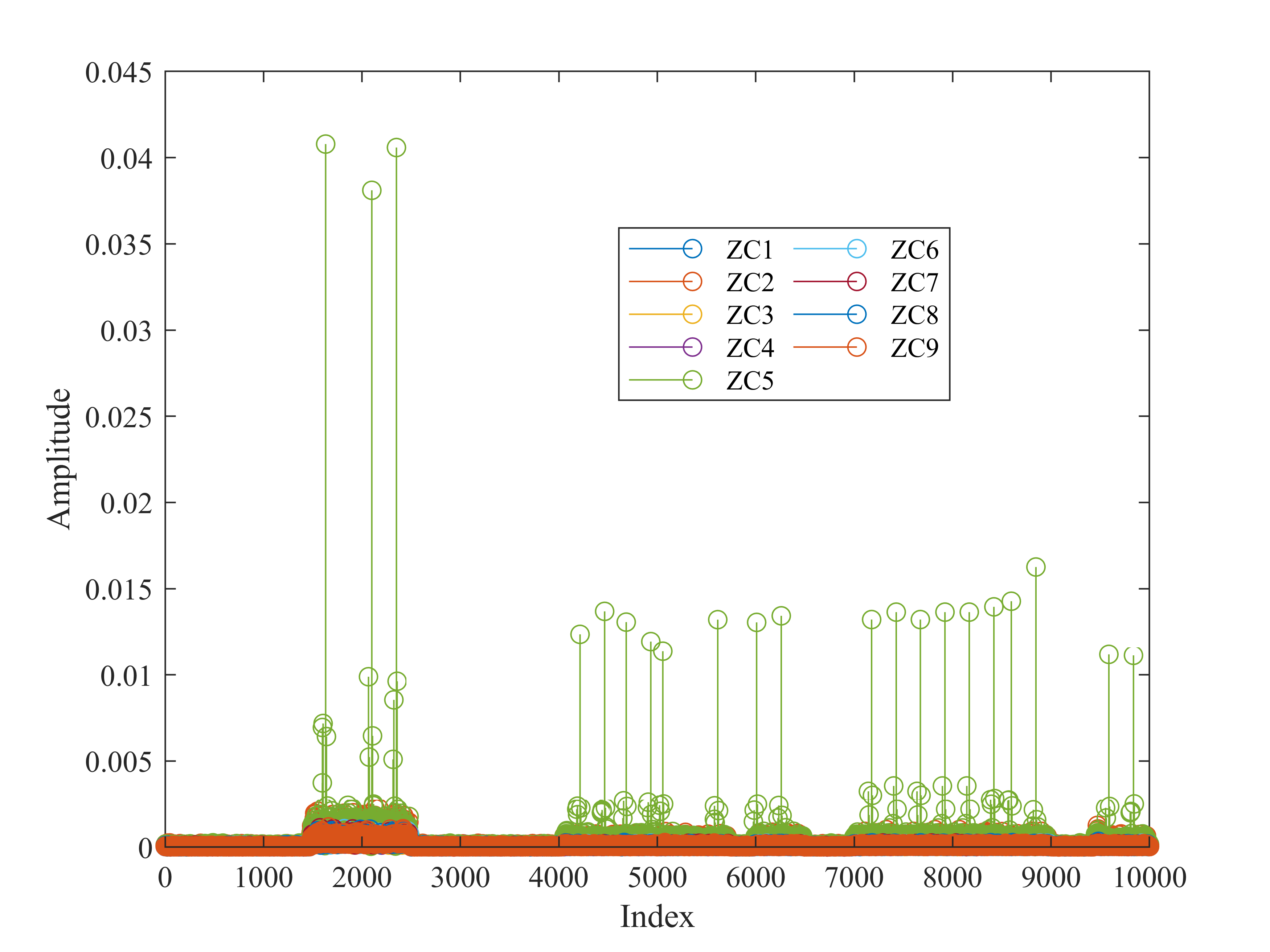}
			\label{zc_13}}
		\hfil
		\subfloat[]{\includegraphics[width=1.7in]{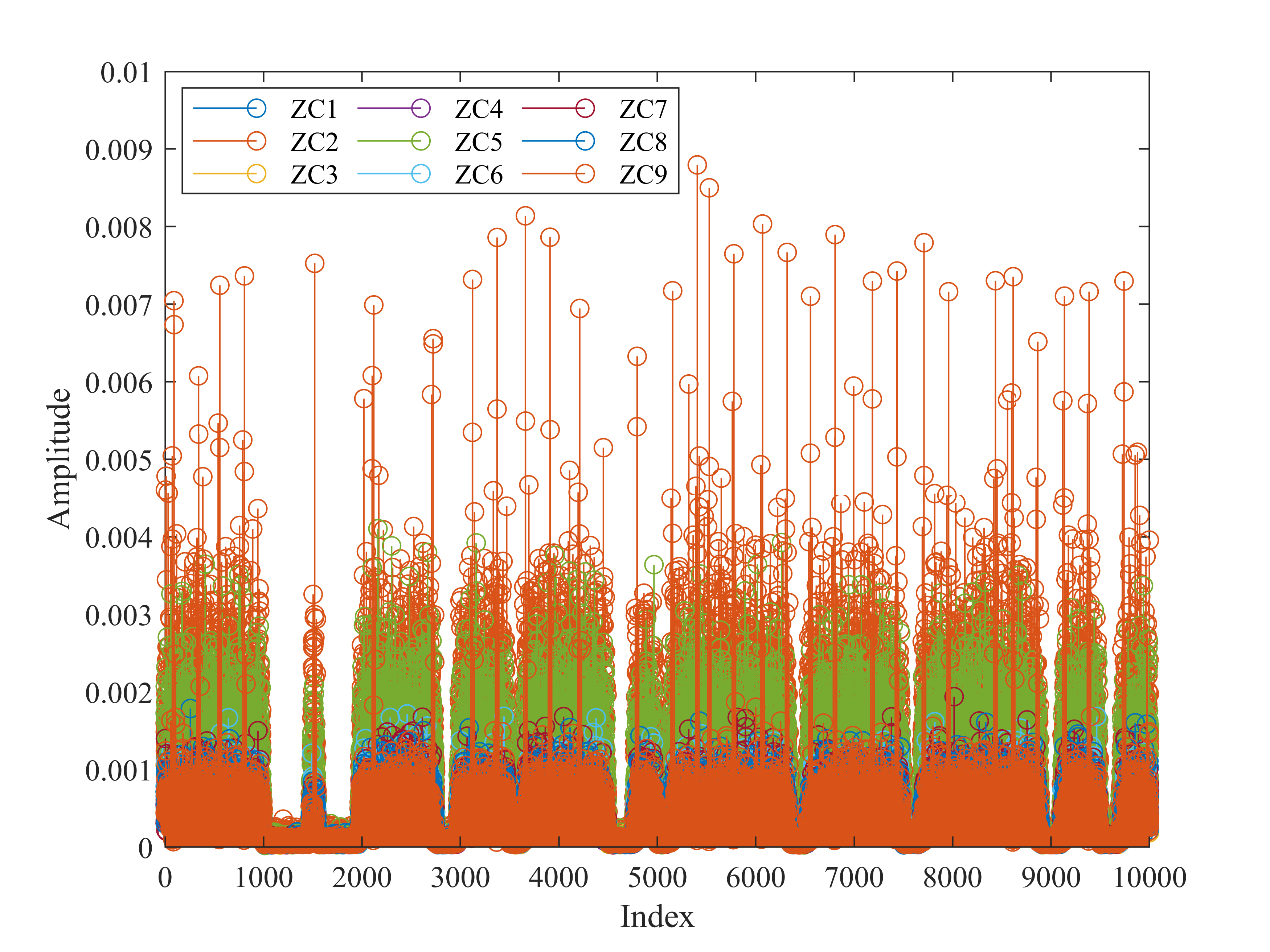}
			\label{zc_14}}
		\hfil
		\subfloat[]{\includegraphics[width=1.7in]{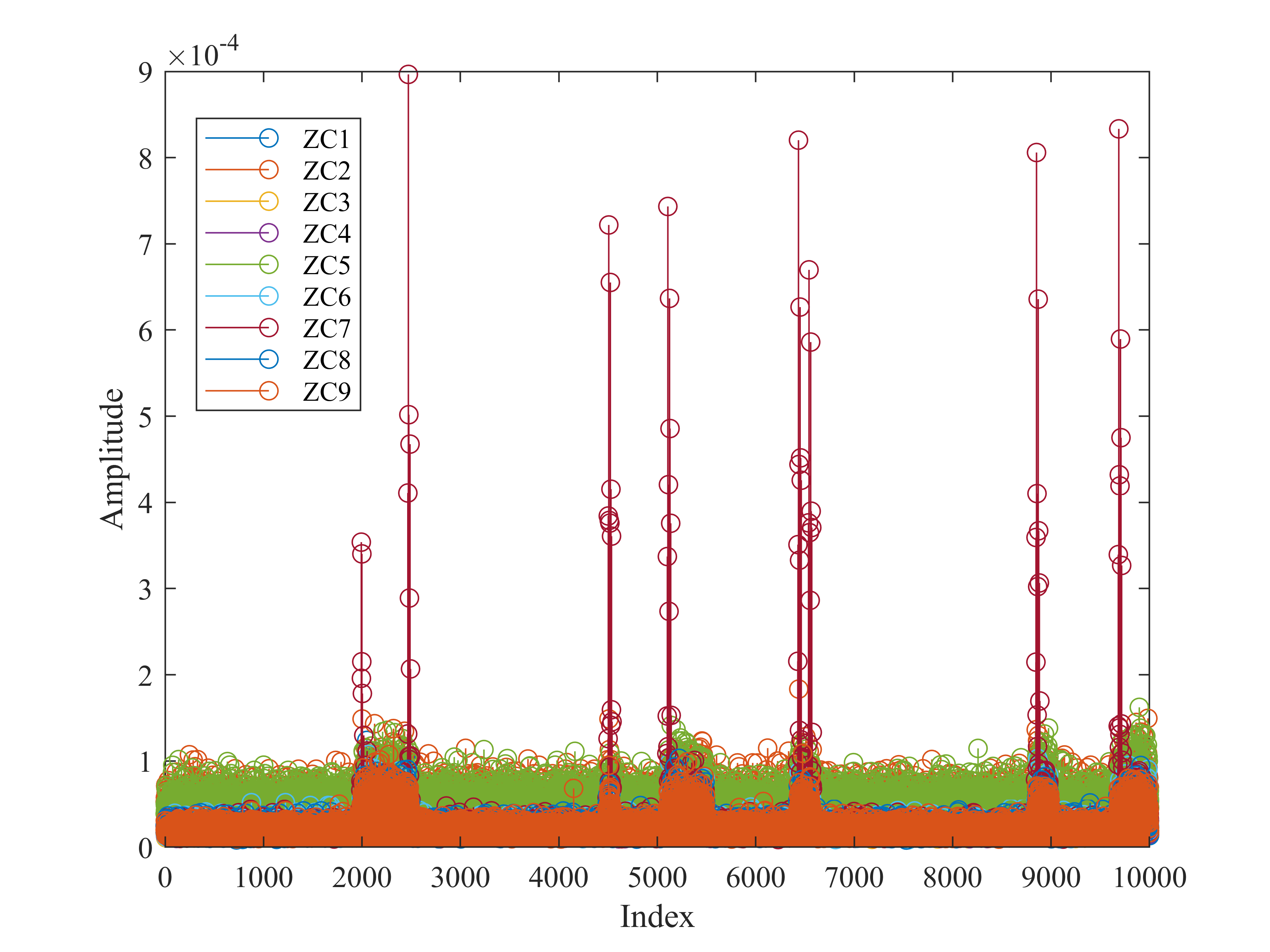}
			\label{zc_15}}
		\hfil
		\subfloat[]{\includegraphics[width=1.7in]{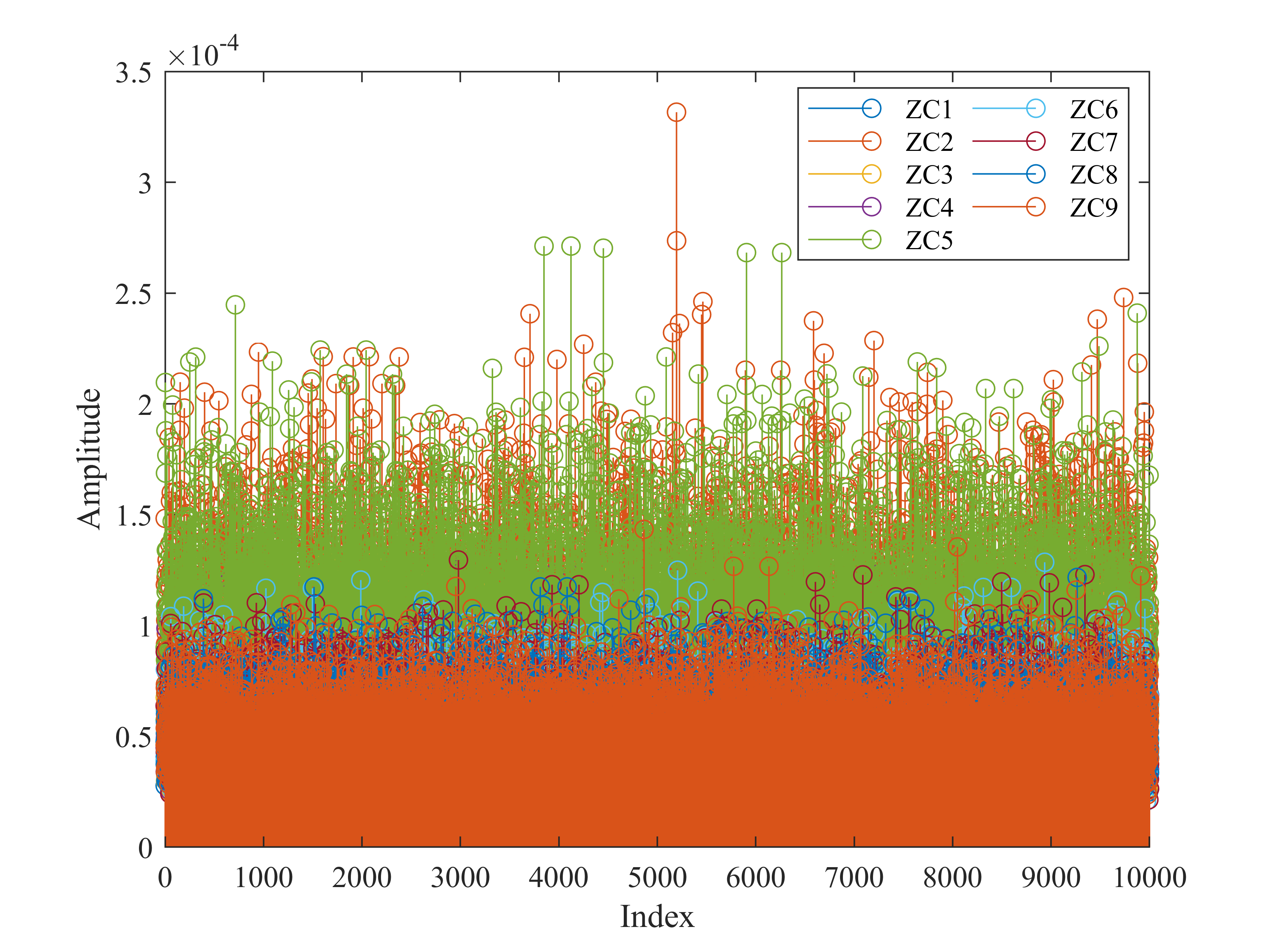}
			\label{zc_16}}
		\caption{\(\tiny{\mathbf{R}_i}\) of RF signals at SNR = 15 dB. (a) T0000D00S00. (b) T0001D00S00. (c) T0010D00S00. (d) T0011D00S00. (e) T0100D00S00. (f) T0101D00S00. (g) T0110D00S00. (h) T0111D00S00. (i) T1000D00S00. (j) T1001D00S00. (k) T1010D00S00. (l) T1011D00S00. (m) T1100D00S00. (n) T1101D00S00. (o) T1110D00S00. (p) T1111D00S00.}
		\label{Fig_zc}
	\end{figure*}

\end{enumerate}

\section{Drone OODD Algorithm}
\label{sec3}

\begin{figure}[!t]
	\centering
	\includegraphics[width=1.5in]{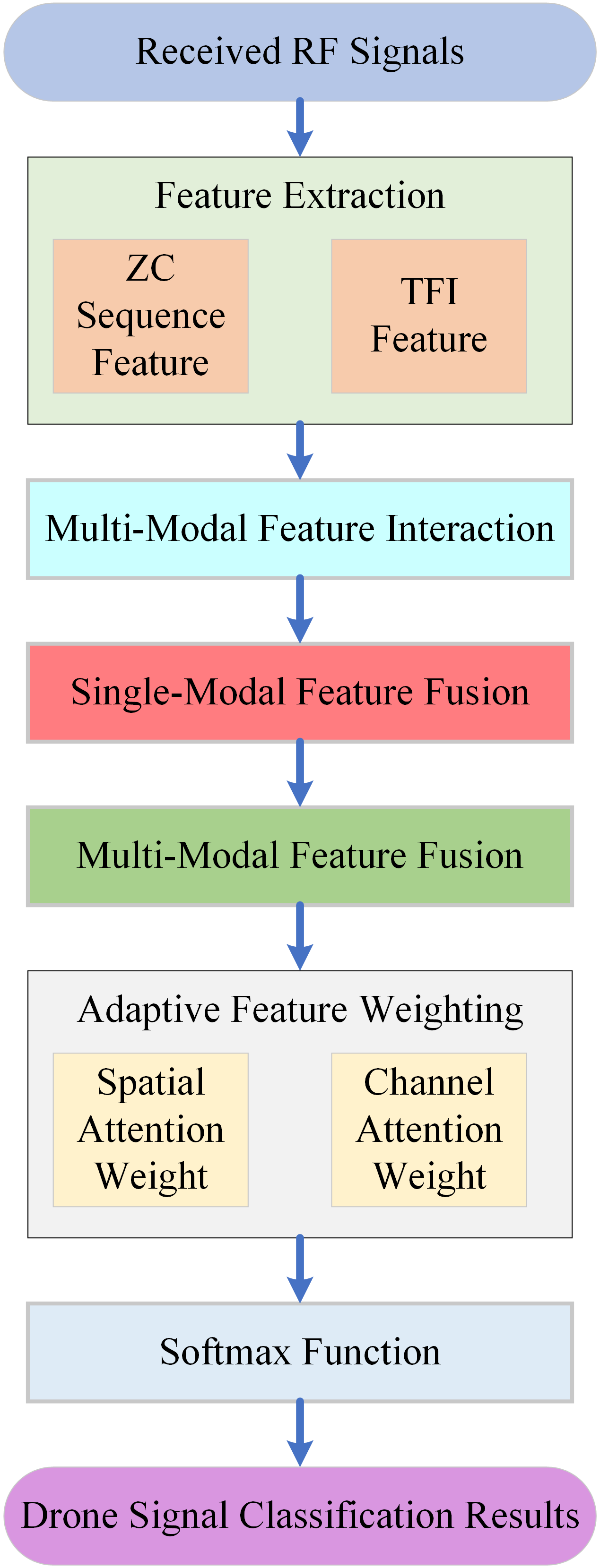}
	\caption{A sketch of the proposed drone RID algorithm.}
	\label{Fig-net}
\end{figure}

To achieve the OODD of drone signals, we propose an algorithm for cognitive fusion of ZC sequence feature and TFI feature as shown in Fig. \ref{Fig-net}. The proposed algorithm comprises four main modules, which are feature extraction, multi-modal feature interaction (MMFI), single-modal feature fusion (SMFF), multi-modal feature fusion (MMFF), and adaptive feature weighting (AFW). Each of these components will be elaborated in detail in the following sections.

\subsection{Feature Extraction Module}

\begin{figure}[!t]
	\centering
	\includegraphics[width=2in]{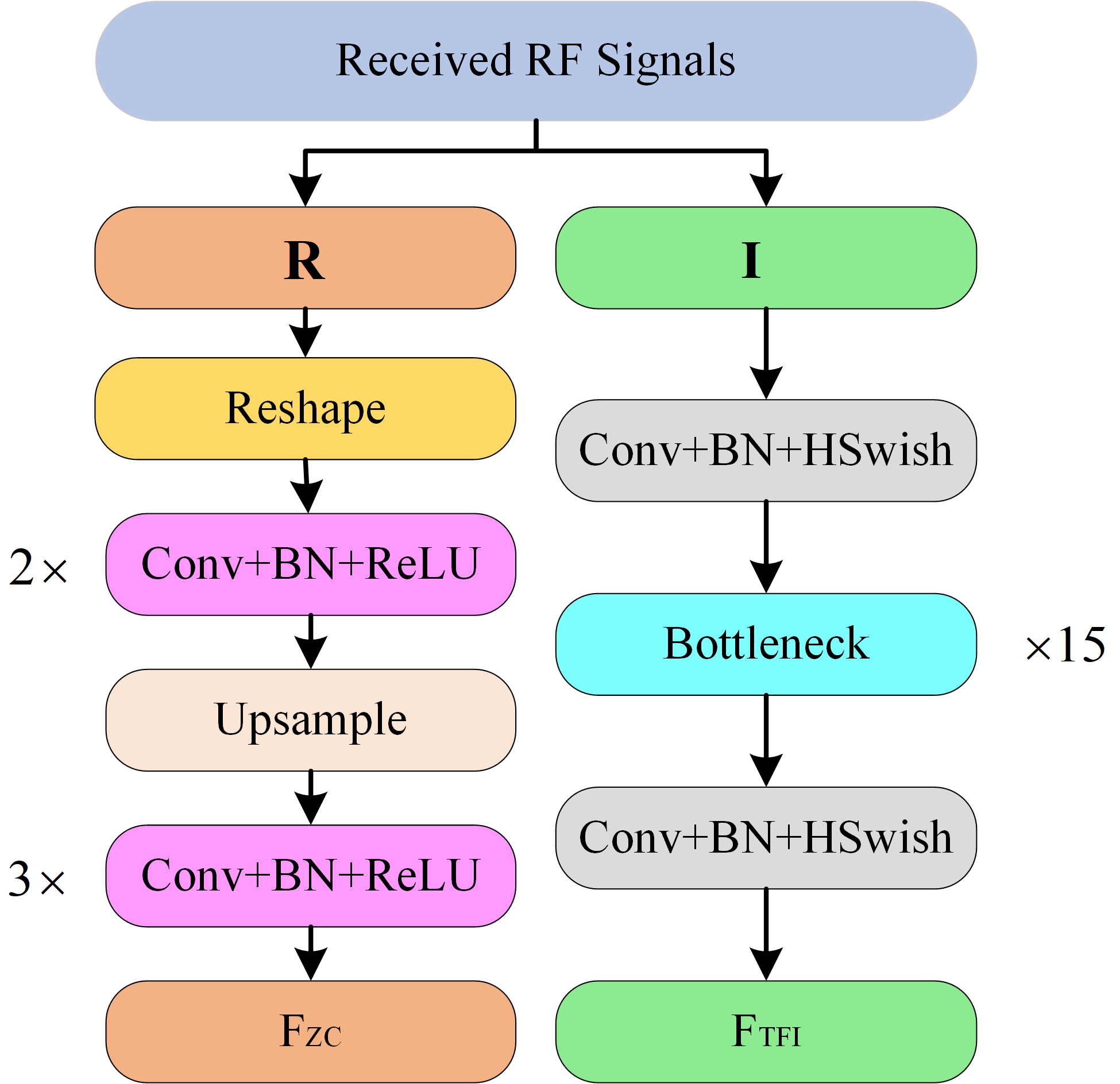}
	\caption{Detailed network architecture for feature extraction module.}
	\label{Fig-extraction}
\end{figure}

The feature extraction module derives \(\mathbf{I}\) and \(\mathbf{R}\) from \(x(l)\), which are subsequently transformed into \(F_{\text{TFI}}\) and \(F_{\text{ZC}}\) with identical dimensions via parallel network branches as shown in Fig. \ref{Fig-extraction}, thereby facilitating the subsequent cognitive fusion process.

Specifically, for the sequence processing branch of the DroneRFa dataset, \(\mathbf{R}\in {{\mathbb{R}}^{9\times 10^6}}\) is reshaped into \({{\mathbb{R}}^{3\times 3\times 10^6}}\) to enhance spatial structure. Two consecutive 2D convolutional (Conv) layers followed by batch normalization (BN) and ReLU activation function are applied to compress the channel dimension while capturing local spatial features. The \(3\times 3\) spatial dimensions are then upsampled to ensure alignment with the output features of the image processing branch, facilitating subsequent fusion. This is followed by three additional Conv-BN-ReLU layers, which extract deeper hierarchical features and enhance representational capacity, and the final output feature is \(F_{\text{ZC}}\in {{\mathbb{R}}^{7\times 7 \times 960}}\).

For the image processing branch, \(\mathbf{I}\in {{\mathbb{R}}^{3\times 3\times 224}}\) is first processed by Conv-BN-HSwish layer to extract initial features. Unlike the ReLU function commonly used in conventional convolutional networks, the smoother HSwish function is suitable for lightweight architectures. Subsequently, feature map passes through 15 consecutively stacked Bottleneck blocks from the MobileNetV4 network\textcolor{blue}{\cite{refv4}}, each comprising \(3\times 3\) convolution, depthwise separable convolution, channel attention mechanism, and residual connection. This structure enables the extraction of rich image features while maintaining a lightweight network design. Finally, another Conv-BN-HSwish layer is applied to obtain the output feature \(F_{\text{TFI}}\in {{\mathbb{R}}^{7\times 7 \times 960}}\). 

\subsection{MMFI Module}
The MMFI module explores the complementary features across modalities from both channel and spatial dimensions, while effectively suppressing redundant information and inter-modal interference. Given that the network architectures for the channel and spatial dimensions share the same structure, which differs only in certain operations such as Global Average Pooling (GAP), Global Max Pooling (GMP), and convolutional kernel parameters, only the structure and operation of the channel dimension are detailed herein.

\begin{figure}[!t]
	\centering
	\includegraphics[width=3in]{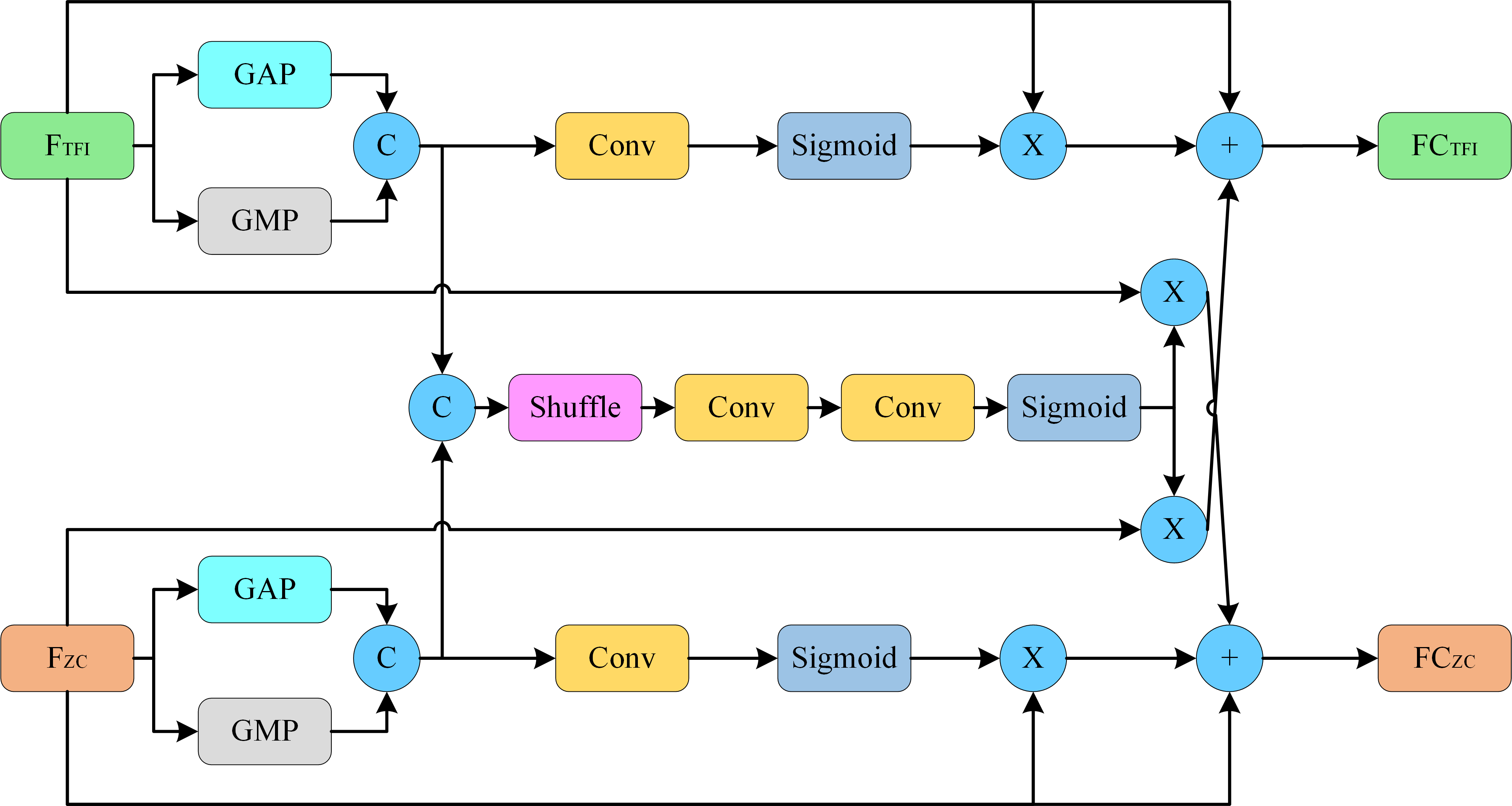}
	\caption{Detailed network architecture in channel dimension for MMFI module.}
	\label{Fig-interaction}
\end{figure}

The network structure is shown in Fig. \ref{Fig-interaction}, both \(F_{\text{TFI}}\in {{\mathbb{R}}^{H\times W \times C}}\) and \(F_{\text{ZC}}\in {{\mathbb{R}}^{H\times W \times C}}\) are independently processed through GAP and GMP to generate four corresponding feature vectors \(\in {{\mathbb{R}}^{1\times 1 \times C}}\). These vectors are concatenated along the channel dimension to form new composite vectors, which are subsequently passed through a convolutional layer followed by a sigmoid activation function to compress the feature size. Notably, an additional channel shuffle operation is applied to enhance the mixing of multi-modal features \(\in {{\mathbb{R}}^{1\times 1 \times 4C}}\). Let \(F_1\), \(F_2\), \(F_3\), and \(F_4\) denote \(\text{GAP}( {F_{\text{TFI}}} )\), \(\text{GMP}( {F_{\text{TFI}}} )\), \(\text{GAP}( {F_{\text{ZC}}} )\), and \(\text{GMP}( {F_{\text{ZC}}} )\), respectively, the operations can be expressed as
\begin{equation}
	\label{Eq5}
	{F_5}=\text{Sigmoid}\{ \text{Conv}( \text{Conc}[ F_1;F_2 ] ) \},
\end{equation}
\begin{equation}
	\label{Eq6}
	{F_6}=\text{Sigmoid}\{ \text{Conv}( \text{Conc}[ F_3;F_4 ] ) \},
\end{equation}
\begin{equation}
	\label{Eq7}
	{F_7}=\text{Sigmoid}\{ \text{Conv}^2[\text{Shuffle}( \text{Conc}[ F_1;F_2;F_3;F_4 ] )] \}.
\end{equation}

Then the weighted vectors \(F_5\), \(F_6\), and \(F_7\) are multiplied with \(F_{\text{TFI}}\) and \(F_{\text{ZC}}\) to generate refined feature maps, where redundant information is suppressed and complementary features are preserved. The features obtained after multi-modal interaction along the channel dimension can be computed by
\begin{equation}
	\label{Eq8}
	\text{FC}_{\text{TFI}}={F_{\text{TFI}}}+{F_5}{F_{\text{TFI}}}+{F_7}{F_{\text{ZC}}},
\end{equation}
\begin{equation}
	\label{Eq9}
	\text{FC}_{\text{ZC}}={F_{\text{ZC}}}+{F_6}{F_{\text{ZC}}}+{F_7}{F_{\text{TFI}}}.
\end{equation}

Similar to the network architecture along the channel dimension, the spatial dimension modifies only a subset of operation parameters. Specifically, vectors \(\in {{\mathbb{R}}^{H\times W \times 1}}\) are obtained by applying GAP and GMP to \(F_{\text{TFI}}\) and \(F_{\text{ZC}}\), which are concatenated along the channel dimension. Then the Conv and Sigmoid operation are utilized to generate the weighted vectors \(F_8\), \(F_9\), and \(F_{10}\in {{\mathbb{R}}^{H\times W \times 1}}\). The features obtained after multi-modal interaction along the spatial dimension are given by
\begin{equation}
	\label{Eq10}
	\text{FS}_{\text{TFI}}={F_{\text{TFI}}}+{F_8}{F_{\text{TFI}}}+{F_{10}}{F_{\text{ZC}}},
\end{equation}
\begin{equation}
	\label{Eq11}
	\text{FS}_{\text{ZC}}={F_{\text{ZC}}}+{F_9}{F_{\text{ZC}}}+{F_{10}}{F_{\text{TFI}}}.
\end{equation}

\subsection{SMFF Module}

\begin{figure}[!t]
	\centering
	\includegraphics[width=3.5in]{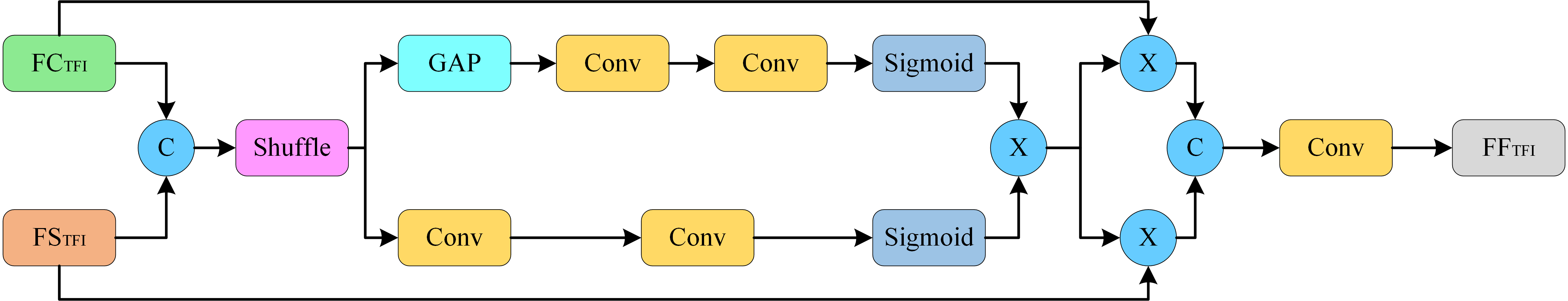}
	\caption{Detailed network architecture for SMFF module with TFI feature.}
	\label{Fig-fusion}
\end{figure}

Building upon the fully interacted and selectively refined multi-modal features, \(\text{FC}_{\text{TFI}}\), \(\text{FC}_{\text{ZC}}\), \(\text{FS}_{\text{TFI}}\), and \(\text{FS}_{\text{ZC}}\)
are obtained, which require further integrated processing to accurately represent the information from different modalities. As illustrated in Fig. \ref{Fig-fusion}, \(\text{FC}_{\text{TFI}}\) and \(\text{FS}_{\text{TFI}}\) are fused through a series of operations to produce \(\text{FF}_{\text{TFI}}\), thereby achieving comprehensive integration and selection of the information contained in the original features across both channel and spatial dimensions.

Specifically, \(\text{FC}_{\text{TFI}}\) and \(\text{FS}_{\text{TFI}}\) are concatenated and shuffled along the channel dimension to ensure sufficient information mixing. On one hand, the mixed feature undergo GAP to obtain feature \(\in {{\mathbb{R}}^{1\times 1 \times 2C}}\), which compresses the spatial dimensional information, followed by two successive Conv layers and a sigmoid activation function to produce the channel-weighted vector \(F_{11}\in {{\mathbb{R}}^{1\times 1 \times C}}\). On the other hand, the mixed feature is processed through two successive Conv layers and a sigmoid activation function to generate the spatial-weighted vector \(F_{12}\in {{\mathbb{R}}^{H\times W \times C}}\), which is then element-wise multiplied with \(F_{11}\) to obtain \(F_{13}\). Then the fusion of TFI feature can be expressed as
\begin{equation}
	\label{Eq12}
	\text{FF}_{\text{TFI}}=\text{Conv}\{\text{Conc}[F_{13}\text{FC}_{\text{TFI}} ; F_{13}\text{FS}_{\text{TFI}}]\}.
\end{equation}

Similarly, the fusion of ZC sequence feature can be given by
\begin{equation}
	\label{Eq13}
	\text{FF}_{\text{ZC}}=\text{Conv}\{\text{Conc}[F_{14}\text{FC}_{\text{ZC}} ; F_{14}\text{FS}_{\text{ZC}}]\},
\end{equation}
where \(F_{14}\) is the Weighted vector fully integrated and selected through channel and spatial dimensions for \(\text{FC}_{\text{ZC}}\) and \(\text{FS}_{\text{ZC}}\).

\subsection{MMFF Module}

\begin{figure*}[!t]
	\centering
	\includegraphics[width=6in]{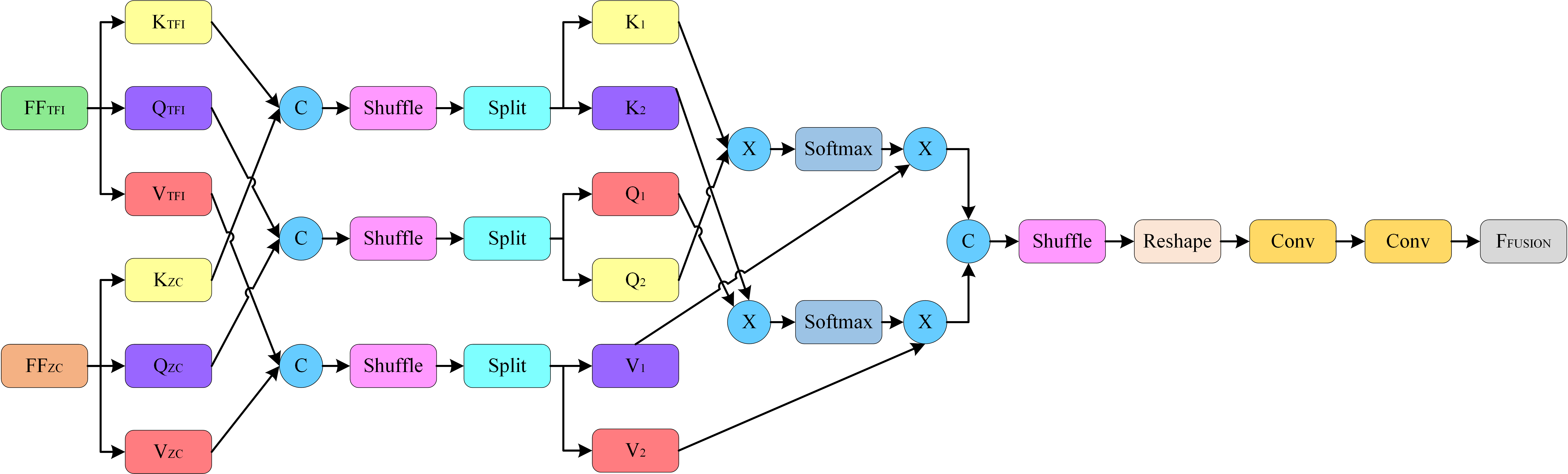}
	\caption{Detailed network architecture for MMFF module.}
	\label{Fig-crossattention}
\end{figure*}

Although preliminary interaction and fusion have been performed between the TFI feature and the ZC sequence feature, applying attention mechanisms within a single modality may result in insufficient exploitation of modal information. Therefore, it is necessary to impose cross-attention on the fused features to enhance inter-modal information utilization, and the network architecture is provided in Fig. \ref{Fig-crossattention}.

Specifically, for \(\text{FF}_{\text{TFI}}\) and \(\text{FF}_{\text{ZC}}\), the network is trained to generate \(K_{\text{TFI}}\), \(Q_{\text{TFI}}\), \(V_{\text{TFI}}\), \(K_{\text{ZC}}\), \(Q_{\text{ZC}}\), and \(V_{\text{ZC}}\in {{\mathbb{R}}^{HW \times D}}\), while \(D=\frac{C}{4}\) is often employed to reduce computational complexity. These features are then concatenated in pairs along the channel dimension, shuffled, and subsequently separated into \(K_1\), \(Q_1\), \(V_1\), \(K_2\), \(Q_2\), and \(V_2\in {{\mathbb{R}}^{HW \times D}}\), thereby ensuring thorough intermixing of information from different modalities. The normalized attention weights for TFI feature and ZC sequence feature can be expressed as
\begin{equation}
	\label{Eq14}
	\text{AW}_\text{TFI}=\text{Softmax}(\frac{Q_1K_1^T}{D}),
\end{equation}
\begin{equation}
	\label{Eq15}
	\text{AW}_\text{ZC}=\text{Softmax}(\frac{Q_2K_2^T}{D}).
\end{equation}

Then the weighted result of the mutually attended salient information between the TFI feature and ZC sequence feature is given by
\begin{equation}
	\label{Eq16}
	F_{15}=\text{AW}_\text{TFI}V_1,
\end{equation}
\begin{equation}
	\label{Eq17}
	F_{16}=\text{AW}_\text{ZC}V_2.
\end{equation}

The shuffle operation is applied after concatenating \(F_{15}\) and \(F_{16}\) along the channel dimension, followed by Reshape function to obtain feature \(\in{{\mathbb{R}}^{H\times W \times D}}\). Subsequently, two Conv layers are added to compress the channel information and obtain the final multi-modal fusion feature \(F_\text{Fusion}\), which can be expressed as
\begin{equation}
	\label{Eq18}
	F_\text{Fusion}= \text{Conv}^2\{\text{Reshape}[\text{Shuffle}(\text{Conc}[F_{15};F_{16}])]\}.
\end{equation}

\subsection{AFW Module}

For \(F_\text{Fusion}\) obtained through cognitive fusion of TFI feature and ZC sequence feature, which inherently contains the time–frequency characteristics of the communication protocol, it is necessary to analyze and extract discriminative information from both the spatial and channel dimensions to enable OODD.

\begin{figure}[!t]
	\centering
	\includegraphics[width=2.5in]{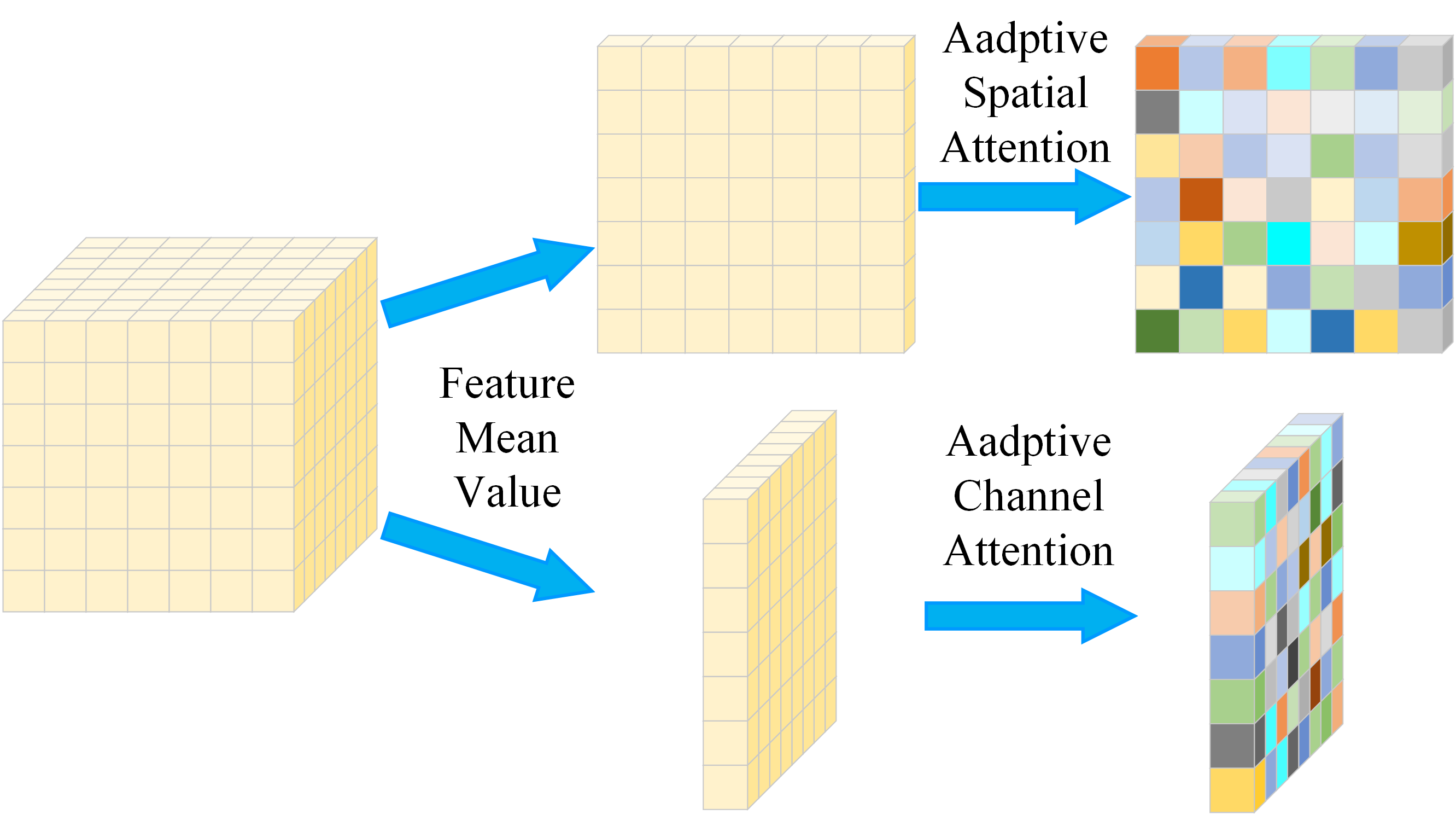}
	\caption{Detailed network architecture for AFW module.}
	\label{Fig-afw}
\end{figure}

As illustrated in Fig. \ref{Fig-afw}, \(F_\text{Fusion}\) is processed through parallel spatial and channel branches to compute inter-class similarity and variance, thereby obtaining the discriminative score for adaptive feature weighting.

Let the training set contain \(P\) distinct classes, with the \(p-\)th class comprising \(Q_p\) samples, and each sample associated with the fused feature \(F_\text{Fusion}^{p,q}\in{{\mathbb{R}}^{H\times W \times D}}\). The mean value of the \(p-\)th class's features along the spatial dimension can be computed by
\begin{equation}
	\label{Eq19}
	M_\text{S}^{p,h,w}=\frac{1}{{{Q}_{p}D}}\sum\limits_{q=1}^{{{Q}_{p}}}\sum\limits_{d=1}^{D}{F_{\text{Fusion}}^{p,q,h,w,d}},
\end{equation}

Let \(\delta(\cdot,\cdot)\) denote the operation of computing the cosine similarity between elements, and the inter-class similarity along the spatial dimension can be given by
\begin{equation}
	\label{Eq20}
	S_S^{h,w}=\frac{1}{P(P-1)}\sum\limits_{p_1=1}^{P}\sum\limits_{\substack{p_2=1 \\ p_2\neq p_1}}^{P}{ \delta(M_\text{S}^{p_1,h,w},M_\text{S}^{p_2,h,w}) },
\end{equation}
while the inter-class variance along the spatial dimension can be expressed as
\begin{equation}
	\label{Eq21}
	V_\text{S}^{h,w}=\frac{1}{P}\sum\limits_{p=1}^{P}{ (M_S^{p,h,w}-\frac{1}{P}\sum\limits_{p=1}^{P}{M_\text{S}^{p,h,w}})^2 }.
\end{equation}

Then the discriminative score used for adaptive spatial attention weighting during the training process can be computed by
\begin{equation}
	\label{Eq22}
	W_S^{h,w}=\alpha S_S^{h,w} - (1-\alpha) V_S^{h,w},
\end{equation}
where \(\alpha\) denotes a trainable parameter of the network, and \(W_S^{h,w}\) will be mapped by the network into the adaptive spatial attention weight \(\in[0,1]\)\textcolor{blue}{\cite{ref192}}. Spatial grids exhibiting smaller \(S_S^{h,w}\) and larger \(V_\text{S}^{h,w}\) are assigned higher attention to facilitate effective OODD. Similarly, the adaptive channel attention weight \(W_C^{d}\) can also be computed, and channels with smaller \(S_C^{d}\) and larger \(V_\text{C}^{d}\) will receive more attention.

\section{Numerical Results}
\label{sec4}
\subsection{Simulation Setup}

\(x(l)\) acquired from various types of drone, diverse flight distances, different LoS conditions, and \(f_s\) is added with AWGN to generate datasets with SNRs ranging from -15 dB to 15 dB in 2 dB intervals. These samples are subsequently partitioned into training, validation, and testing sets in an 8:1:1 ratio. The hardware environment consists of an NVIDIA GeForce RTX 4060 Ti GPU and an 11th Gen Intel(R) Core(TM) i7-14700K @ 3.40GHz, with the remaining simulation parameters are provided in Table \ref{tab:table2}. It is noteworthy that, due to the small size of the sequence features, a larger batchsize can be used to load and train the network. However, the high redundancy information necessitates more epochs to extract abstract features and achieve performance convergence. In contrast, image features are larger in size with lower redundancy, allowing for smaller batchsizes and fewer epochs. Nevertheless, all samples in the training set will be loaded for network training during each epoch, such that the batchsize per load does not affect network performance. Similarly, all algorithms terminate training based on the convergence of evaluation metrics on the validation set to fully reveal algorithm performance, which results in differing numbers of epochs across algorithms.

\begin{table}[!t]
	\caption{Simulation Parameters \label{tab:table2}}
	\centering
	\begin{tabular}{c c}
		\hline
		Parameter & Value \\
		\hline
		Batchsize for Sequence-based Algorithm & 256 \\
		Batchsize for TFI-based Algorithm & 128 \\
		Batchsize for TFI-Sequence-based Algorithm & 128 \\
		Epoch for Sequence-based Algorithm & 100 \\
		Epoch for TFI-based Algorithm & 60 \\
		Epoch for TFI-Sequence-based Algorithm & 60 \\
		Learning Rate & 0.0001 \\
		\hline
	\end{tabular}
\end{table}

The baseline and the proposed algorithms are given as follows.
\begin{enumerate}[leftmargin=0pt, itemindent=2pc, listparindent=\parindent]
	\item{\textit{I/Q Sequences-based CNN (IQ-CNN)}}: \(x(l)\) is directly employed as the input feature to the CNN, whose backbone is same as the feature extraction module of the proposed algorithm. The output of the Softmax function is subsequently utilized to perform OODD.
	
	\item{\textit{ZC Sequence Features-based CNN (ZC-CNN)}}: ZC sequence feature is generated and utilized to train the CNN, and OODD is achieved by the output of the Softmax function.
	
	\item{\textit{TFI Features-based MobileNetV4 (TFI-MobileNet)}}: TFI feature is generated and utilized to train the MobileNetV3, and OODD is achieved by the output of the Softmax function.
	
	\item{\textit{ZC Sequence Feature is concatenated with TFI Feature (ZC-TFI-Conc)}}: The ZC sequence feature and TFI feature are directly concatenated, rather than processed through the proposed MMFI and MMFF module, and is subsequently utilized to perform OODD based on the Softmax function.

	\item{\textit{Reconstruction Error based on the Cognitive Fusion of ZC Sequence Feature and TFI Feature (Fusion-Recon)}}: ZC Sequence Feature and TFI Feature are fused based on the proposed MMFI and MMFF module, while OODD is achieved by leveraging Mahalanobis distance as reconstruction error from an encoder–decoder architecture.
	
	\item{\textit{Confidence Score based on the Cognitive Fusion Feature (Fusion-Conf)}}: The confidence score branch is added to perform OODD based on the cognitive fusion feature.
	
	\item{\textit{Channel Discrimination based on the Cognitive Fusion Feature (Fusion-Channel)}}: The adaptive channel weight is applied based on channel discrimination of the cognitive fusion feature, and OODD is achieved by the output of the Softmax function.
	
	\item{\textit{Spatial Discrimination based on the Cognitive Fusion Feature (Fusion-Spatial)}}: The adaptive Spatial weight is applied based on spatial discrimination of the cognitive fusion feature, and OODD is achieved by the output of the Softmax function.
	
	\item{\textit{Channel and Spatial Discrimination based on the Cognitive Fusion Feature (Fusion-Proposed)}}: The adaptive channel and spatial weight is applied based on the proposed AFW module, and OODD is achieved by the output of the Softmax function.

	The drone broadcast frames detection will be evaluated by the metrics as accuracy, precision, and recall, which can be computed by
	\begin{equation}
		\label{Eq20}
		\text{Accuracy}=\frac{T\!P+T\!N}{T\!P+T\!N+F\!P+F\!N},
	\end{equation}
	\begin{equation}
		\label{Eq21}
		\text{Precision}=\frac{T\!P}{T\!P+F\!P}, 
	\end{equation}
	\begin{equation}
		\label{Eq22}
		\text{Recall}=\frac{T\!P}{T\!P+F\!N},
	\end{equation}
	where \(T\!P\), \(T\!N\), \(F\!P\), and \(F\!N\) denote the true positive, true negative, false positive, and false negative, respectively.
\end{enumerate}

\subsection{Simulation Results}

\begin{figure*}[!t]
	\centering
	\subfloat[]{\includegraphics[width=2.5in]{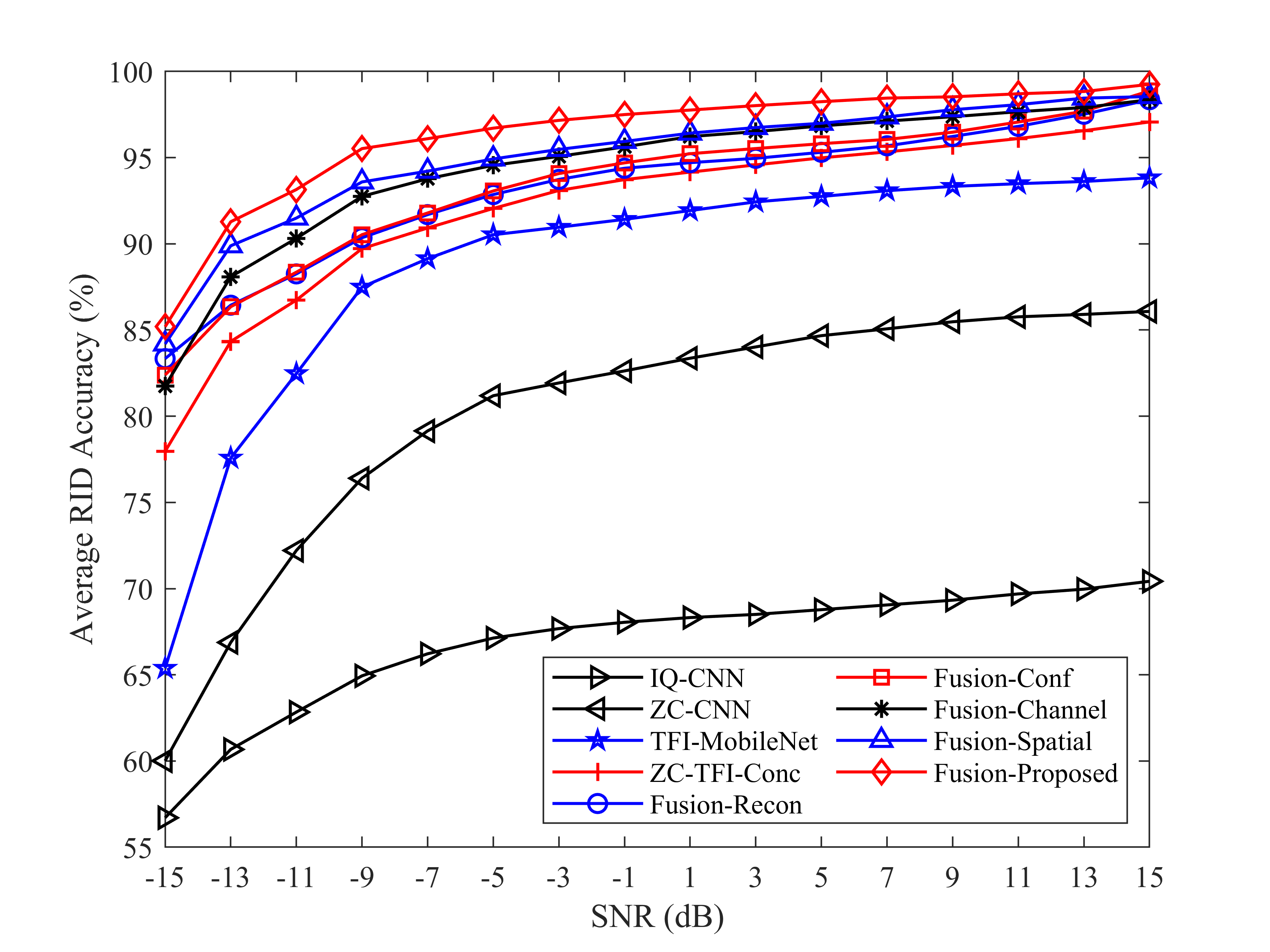}
		\label{1_first}}
	\hfil
	\subfloat[]{\includegraphics[width=2.5in]{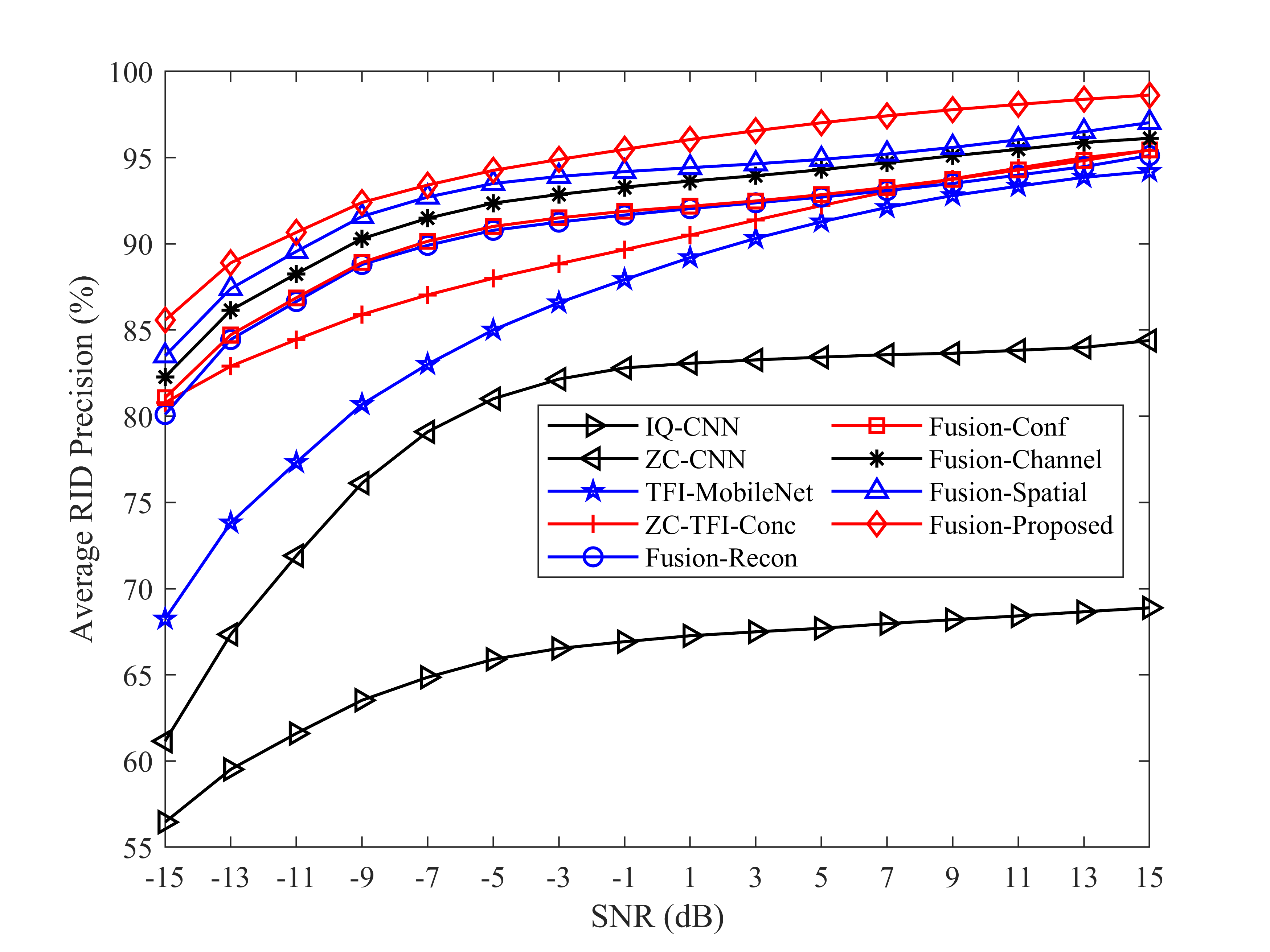}
		\label{2_second}}
	\hfil
	\subfloat[]{\includegraphics[width=2.5in]{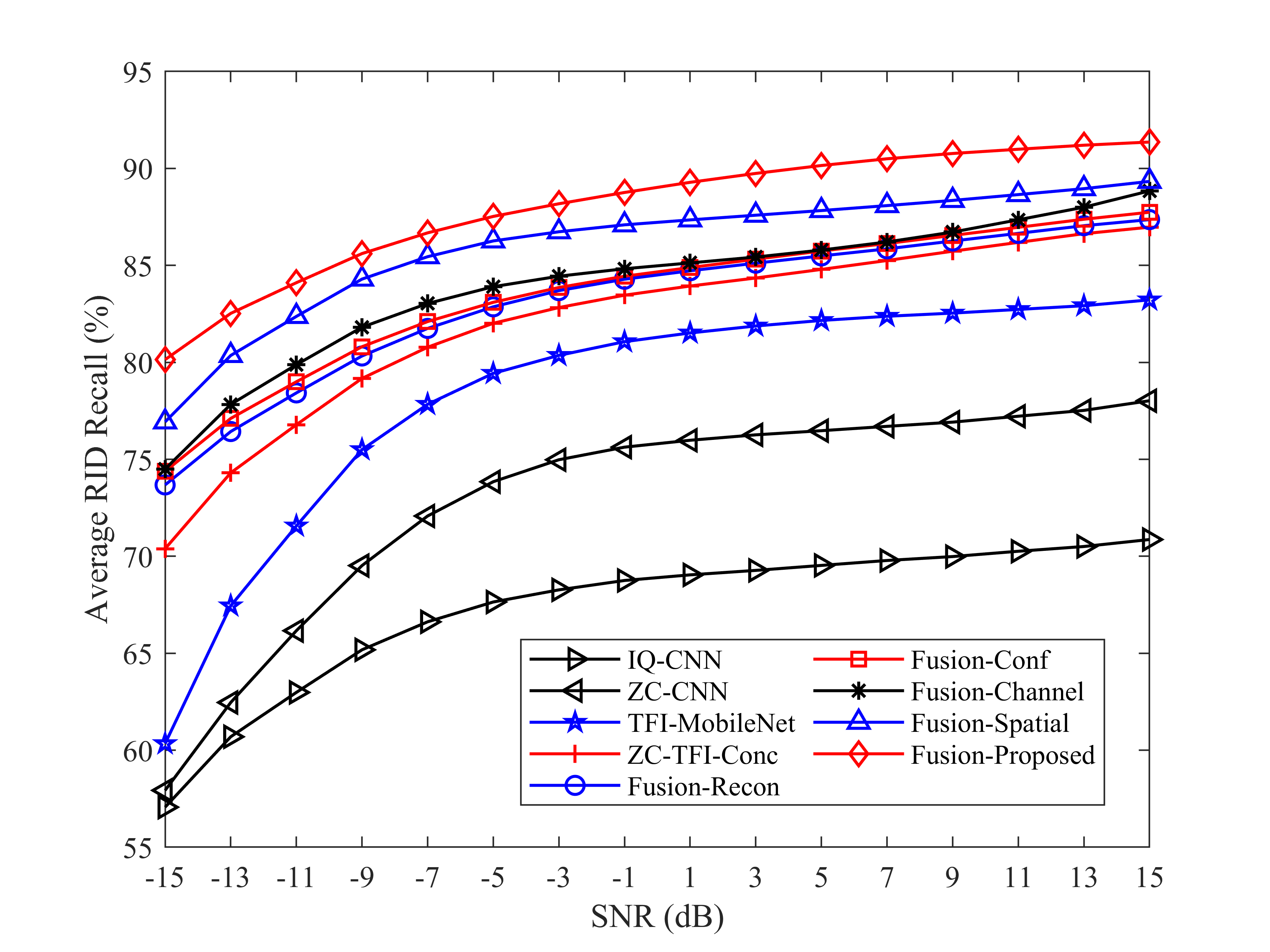}
		\label{3_third}}
	\hfil
	\subfloat[]{\includegraphics[width=2.5in]{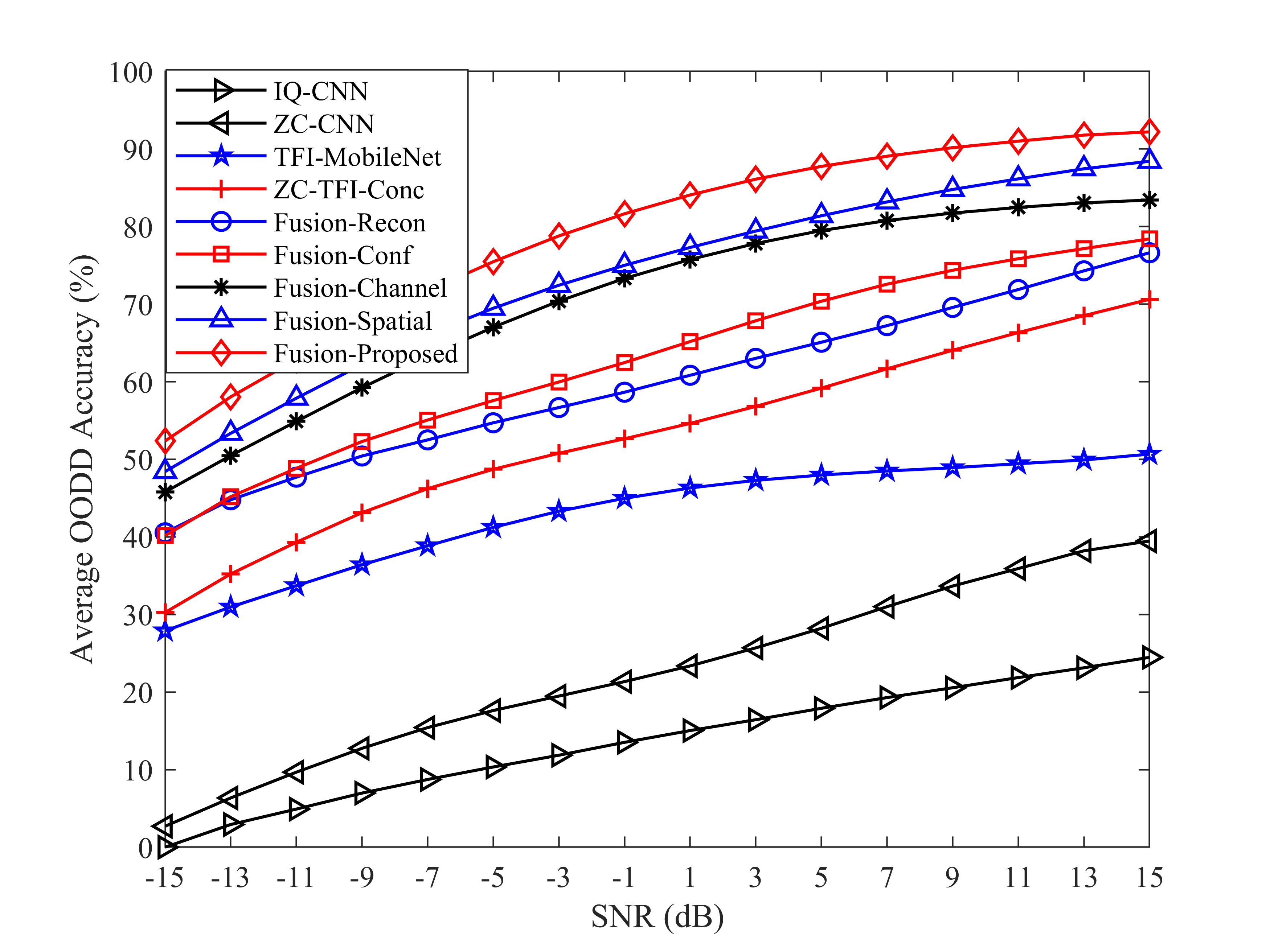}
		\label{4_forth}}
	\caption{Evaluation metrics of different algorithms versus SNR on DroneRFa. (a) Average RID accuracy. (b) Average RID precision. (c) Average RID recall. (d) Average OODD accuracy.}
	\label{Fig_metric}
\end{figure*}

Since OODD constitutes a component of the RID task, the RID metrics of different algorithms are evaluated as shown in Fig. \ref{1_first}, Fig. \ref{2_second}, and Fig. \ref{3_third}, where the performance of all algorithms improves by increasing SNR. However, sequence-based algorithms exhibit inferior performance as they rely solely on time-domain features, which are highly susceptible to interference and exhibit low discriminative capability. It can be observed that utilizing TFI feature alone, or in combination with ZC sequence feature, can effectively enhance RID performance, as TFI can reflect the time–frequency characteristic differences specified by communication protocols. However, simple concatenation of the two different modality features yields inferior results compared to cognitively guided fusion, since the proposed MMFI, SMFF, and MMFF module can effectively integrate features while eliminating redundant information. Compared with reconstruction-based and confidence-based algorithms commonly used for object image classification, the proposed algorithm, which exploits time–frequency characteristic differences of fused features from both channel and spatial dimensions, demonstrates superior performance.

Meanwhile, the detection performance for OOD samples T01111 in the DroneRFa dataset is illustrated in Fig. \ref{4_forth}, with the mean metrics summarized in Table \ref{tab:table3}. Compared with existing algorithms, the proposed algorithm can effectively perform OODD in the drone RID task, achieving at least a 1.7\% improvement in RID accuracy and a minimum 7.5\% gain in OODD accuracy. Although sequence-based algorithms exhibit higher computational complexity in terms of floating point operations per second (FLOPs), their simplicity in network architecture results in the fewest parameters. In contrast, the proposed method, owing to the deployment of specially designed feature interaction, fusion, and time–frequency difference-based weighting modules, entails higher computational complexity and a larger number of parameters.

\begin{table*}[!t]
	\caption{Mean Evaluation Metrics of Different Algorithms \label{tab:table3}}
	\centering
	\begin{tabular}{c c c c c c c}
		\hline
		Algorithm & RID Accuracy & RID Precision & RID Recall & OODD Accuracy & FLOPs & Parameters\\
		\hline
		IQ-CNN & 66.77\% & 65.62\% & 67.28\% & 13.62\% & 1.94 G & 0.46 M \\
		ZC-CNN & 80.05\% & 79.42\% & 72.98\% & 22.55\% & 2.18 G & 0.61 M \\
		TFI-MobileNet & 88.71\% & 86.23\% & 78.30\% & 42.90\% & 0.81 G & 4.22 M \\
		ZC-TFI-Conc & 92.07\% & 89.58\% & 82.10\% & 53.01\% & 1.22 G & 5.79 M \\
		Fusion-Recon & 93.17\% & 90.68\% & 83.12\% & 59.66\% & 1.59 G & 6.57 M \\
		Fusion-Conf & 93.37\% & 90.95\% & 83.47\% & 62.71\% & 1.29 G & 6.84 M \\
		Fusion-Channel & 94.37\% & 92.26\% & 83.85\% & 70.56\% & 1.80 G & 15.08 M \\
		Fusion-Spatial & 95.01\% & 93.17\% & 85.98\% & 73.31\% & 1.80 G & 15.08 M \\
		Fusion-Proposed & \textbf{96.18\%} & \textbf{94.72\%} & \textbf{87.79\%} & \textbf{78.80\%} & \textbf{1.93 G} & \textbf{16.81 M} \\
		\hline
	\end{tabular}
\end{table*}

\begin{figure}[!t]
	\centering
	\includegraphics[width=2.5in]{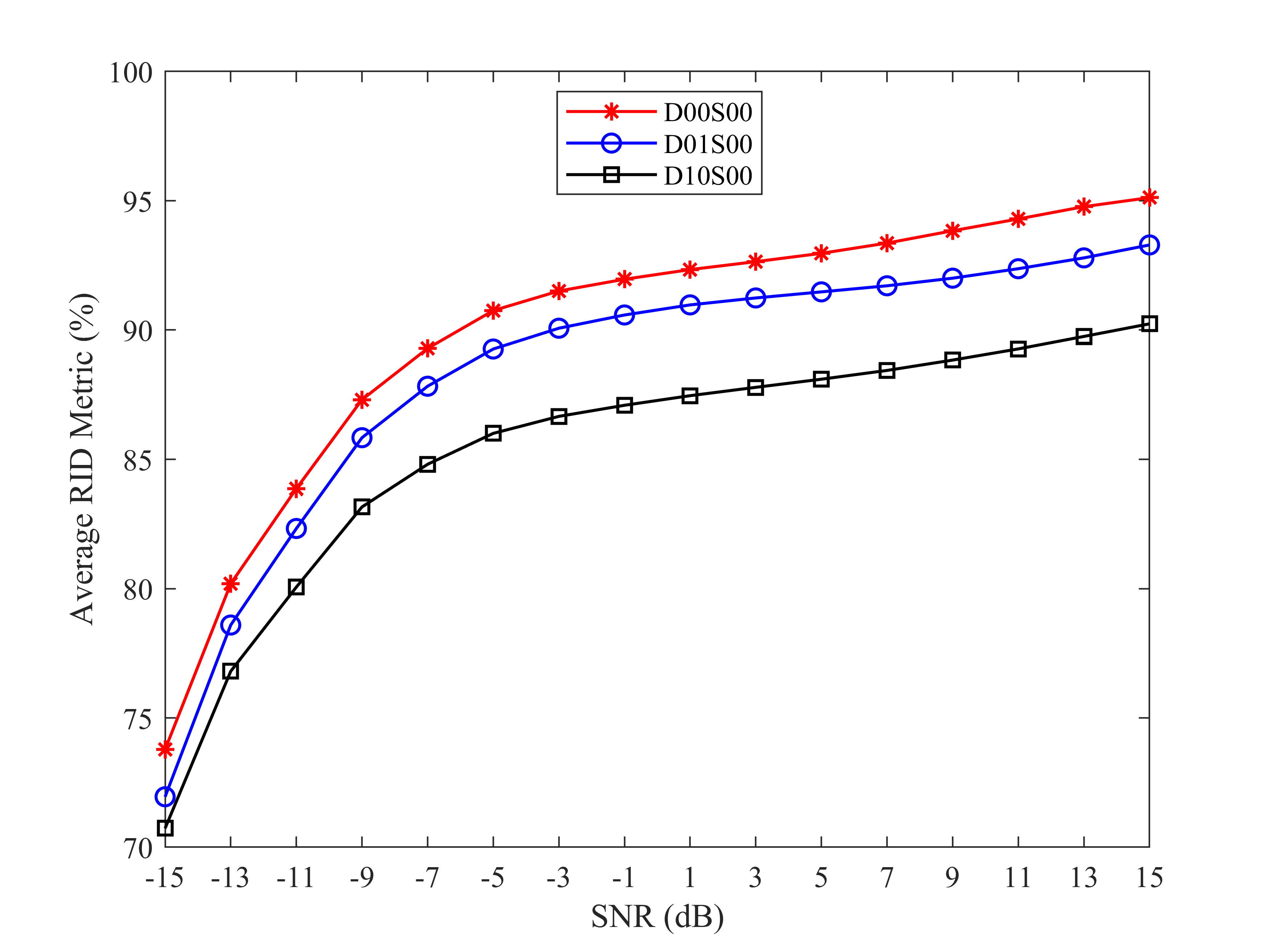}
	\caption{Evaluation metric versus drone flight distances on DroneRFa.}
	\label{Fig_distance}
\end{figure}

Due to the high mobility of drones, the received RF signals may originate from varying distances, thereby resulting in different signal and feature strengths. Consequently, the simulation results of the proposed algorithm under different flight distances are provided in Fig. \ref{Fig_distance}, while the average RID metric is obtained by averaging accuracy, precision, and recall. It can be observed that the average RID metric increases with the rise of SNR regardless of the flight distance. Under the regulatory flight ceiling of 120 m, the RID metric for drones at a distance of 150 m can reach up to 90\%, which is tolerable given the presence of OOD samples. As the flight distance decreases, the features used for identification become more robust, with D00 and D01 achieving identification accuracies that are 5.3\% and 3.5\% higher than that of D10, respectively, thereby validating the robustness of the proposed algorithm to variations in flight distance.

\begin{figure}[!t]
	\centering
	\includegraphics[width=2.5in]{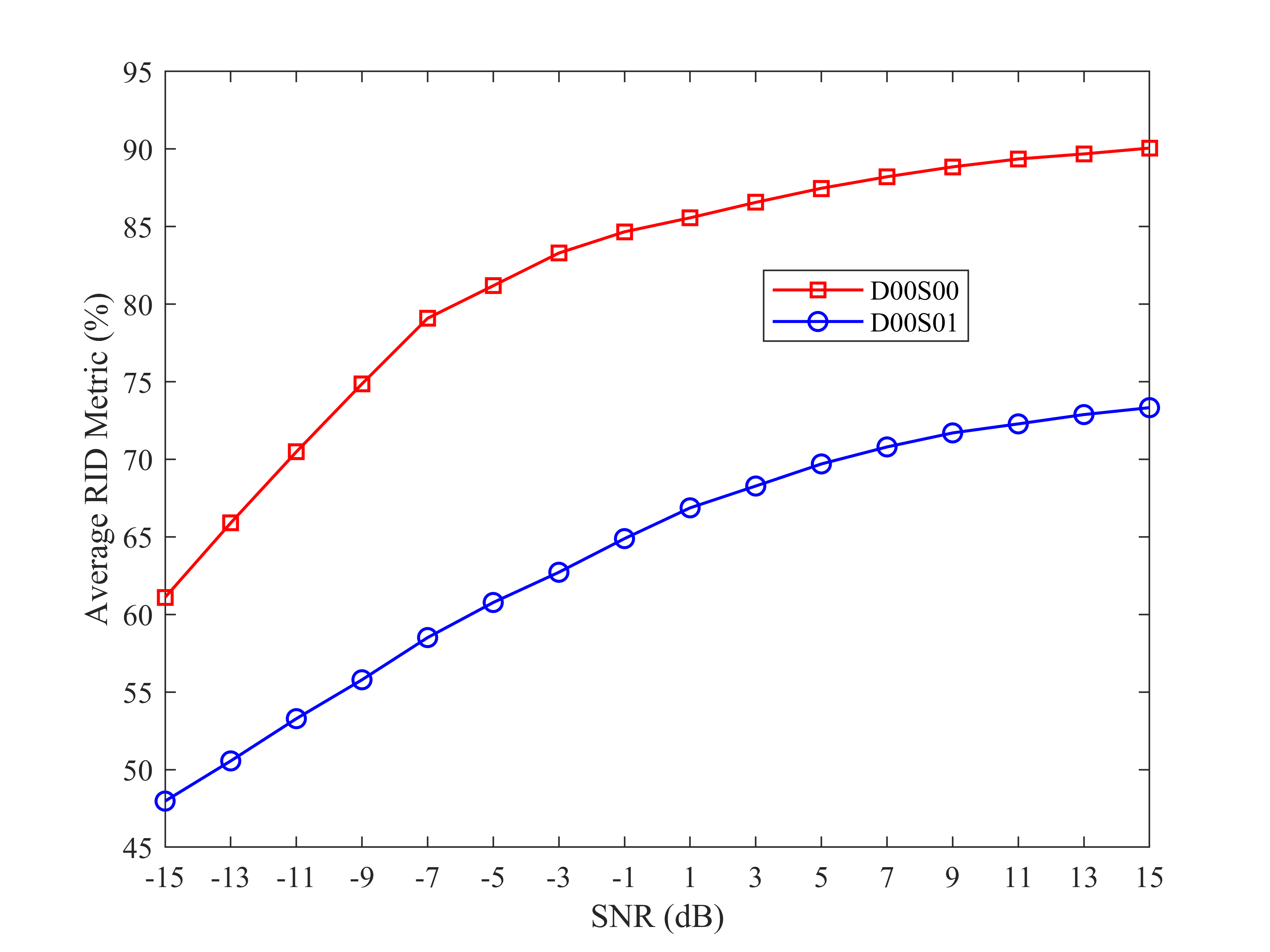}
	\caption{Evaluation metric versus flight conditions on DroneRFb-DIR.}
	\label{Fig_visual}
\end{figure}

Since the received RF signals may originate from NLoS paths in densely built environments, evaluating the robustness of the proposed algorithm under different flight environments is necessary. As shown in Fig. \ref{Fig_visual}, the average RID metric in LoS flight conditions is 28\% higher than that in NLoS, primarily because the average drone signal strength differs by 20 dB between the two conditions. Considering that drones or detectors are often mobile or can be cross-deployed in practical scenarios, the 73.34\% RID accuracy achieved in transient NLoS flight environment is considered acceptable.

\begin{figure}[!t]
	\centering
	\includegraphics[width=2.5in]{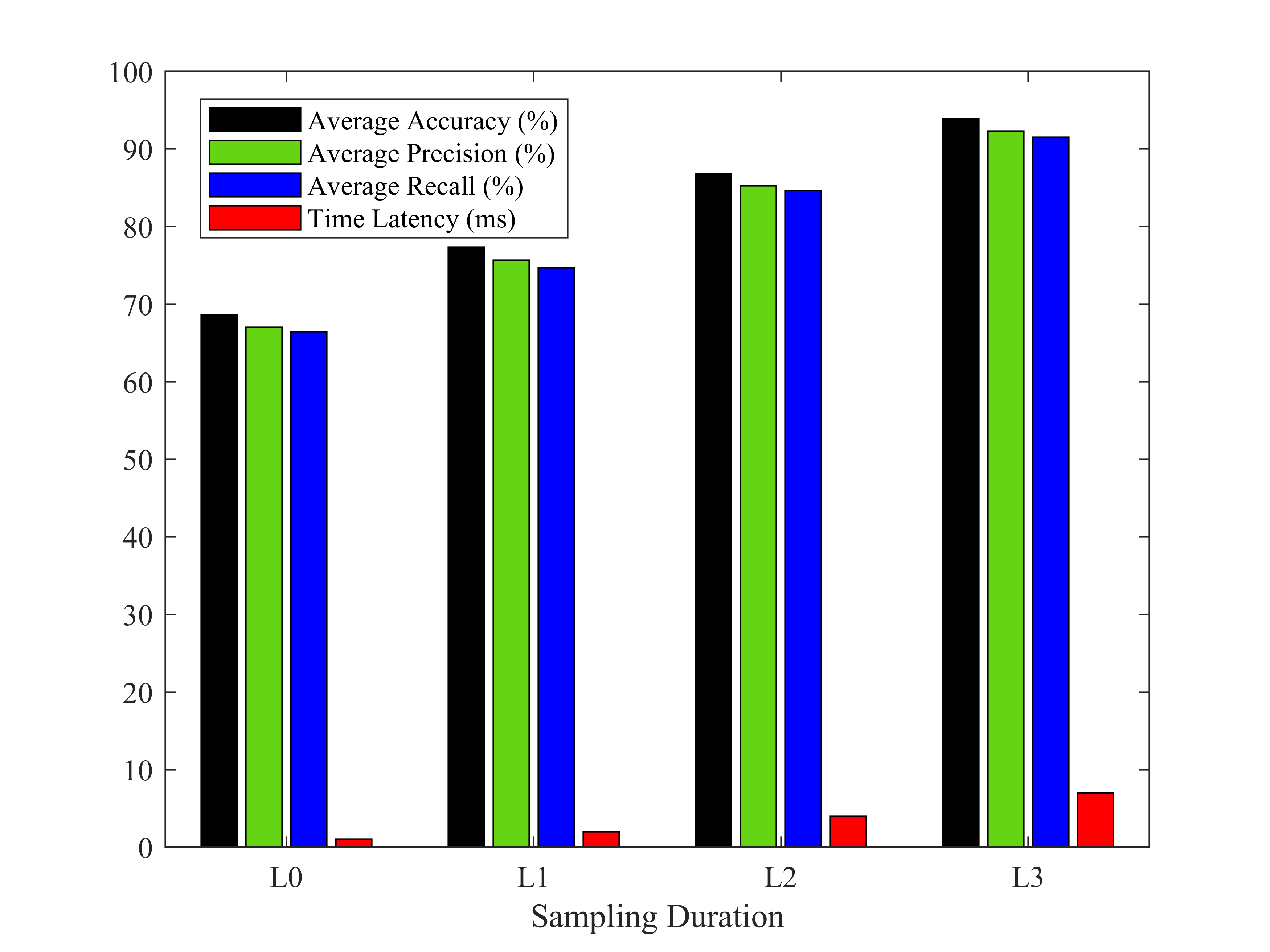}
	\caption{Evaluation metric versus sampling durations.}
	\label{Fig_duration}
\end{figure}

As previously mentioned, different sampling durations not only affect the representation of time-frequency characteristics within TFIs of the same size, but also influence the complexity of feature generation. Therefore, the performance of the proposed algorithm under varying sampling durations is provided in Fig. \ref{Fig_duration}. As the sampling duration increases, richer and more comprehensive inter-frame information in the TFI featur is provided to enhance OODD and RID performance. However, although the complexity of generating ZC sequence features is unaffected by \(L\) since they are all derived from \(10^6\) random samples, the complexity of TFI feature generation scales linearly with \(L\). When the sampling duration increases from \(\text{L}2\) to \(\text{L}3\), the feature generation latency rises from 4 ms to 6 ms, while the RID accuracy improves by 8\%. Since TFI generated from signals with 0.1 s sampling duration already contains sufficiently rich information for effective OODD in RID task, \(\text{L}3\) is selected as the optimal value to achieve the best trade-off between RID accuracy and time latency. 

\begin{figure}[!t]
	\centering
	\includegraphics[width=2.5in]{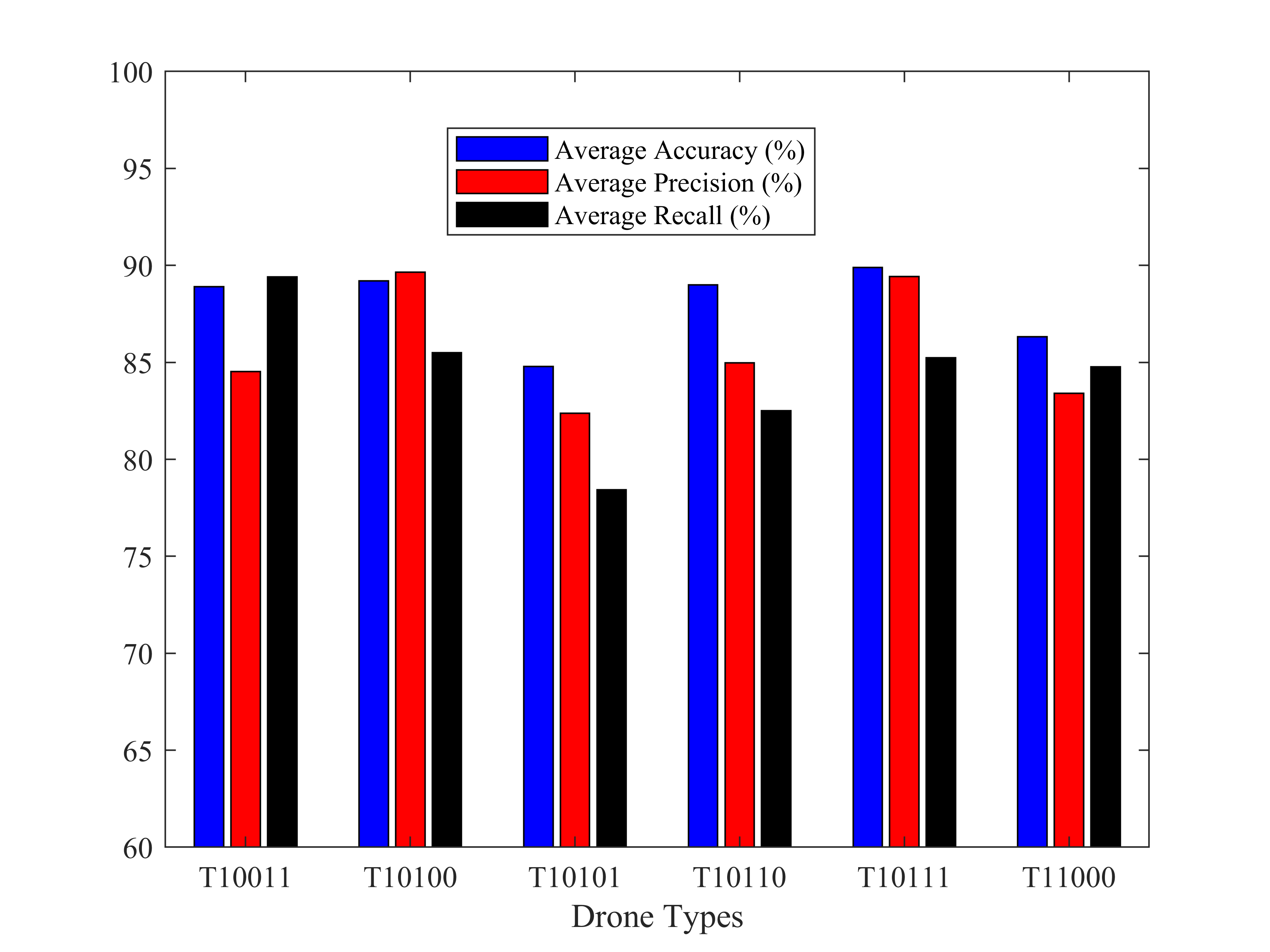}
	\caption{Evaluation metric versus different types of drone on RFUAV.}
	\label{Fig_Type}
\end{figure}

Since ZC sequences are primarily employed by DJI drones, while OODD must account for unknown and non-standard drone communication protocols, the OODD metrics for six types of drones are evaluated as shown in Fig. \ref{Fig_Type}. It can be seen that drones not using ZC sequences can also be effectively detected with a minimum average accuracy of 81.86\%, validating the robustness and effectiveness of the proposed algorithm in achieving OODD within the RID task by leveraging the time-frequency characteristic differences along the spatial and channel dimensions of the cognitive fused features.

\begin{figure}[!t]
	\centering
	\includegraphics[width=2.5in]{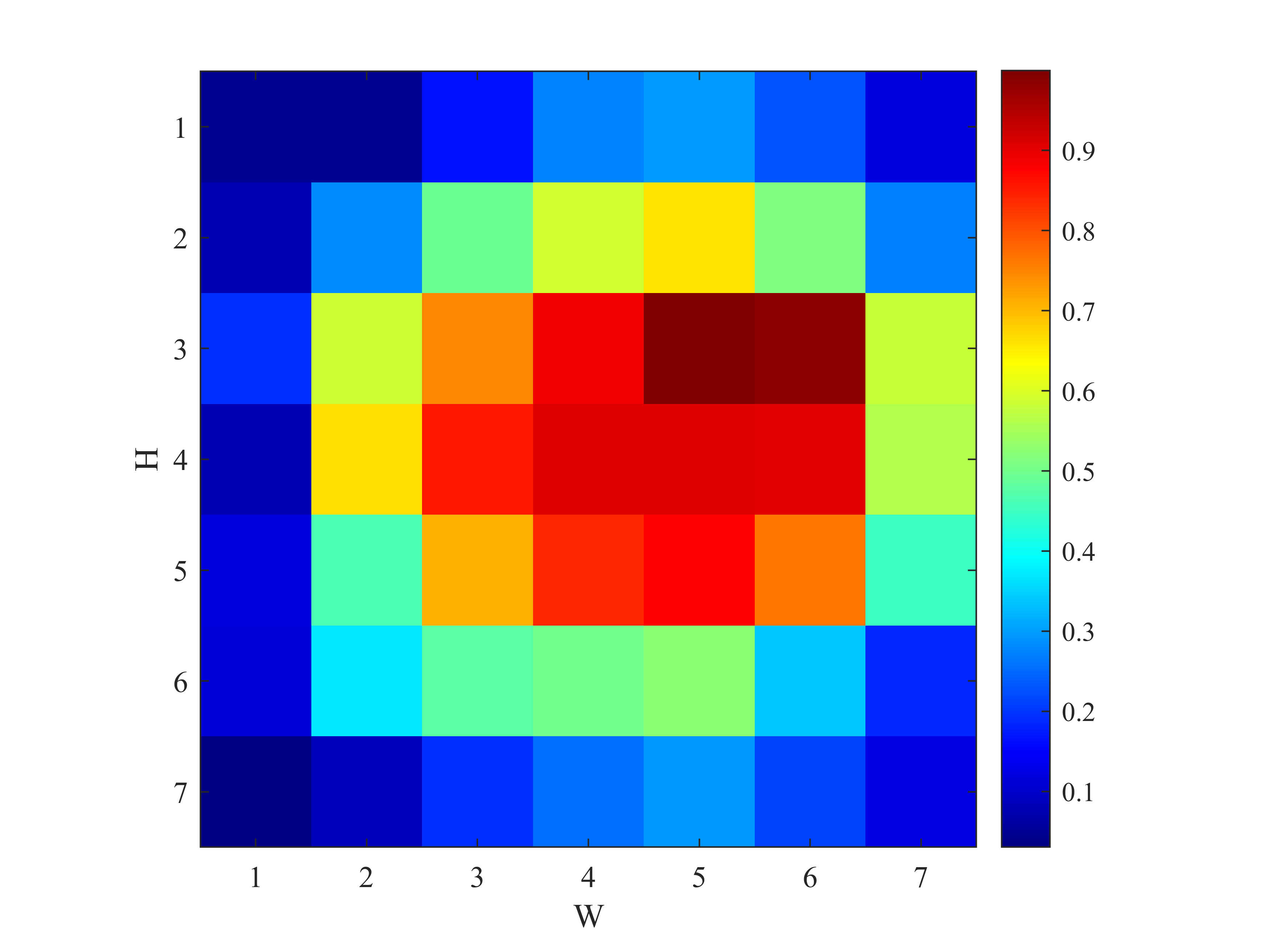}
	\caption{Adaptive spatial weight visualization.}
	\label{Fig_Spatial}
\end{figure}

To more intuitively illustrate the effect of the AFW module, the adaptive spatial weights are visualized as shown in Fig. \ref{Fig_Spatial}, while the channel-wise weights of size \(1\times 1\times 240\) are too large for display. Since ZC sequence features either exhibit obvious and periodic peaks at the corresponding ZC sequence positions or no peaks at all, their impact on spatial weights is minimal. As shown in Fig. \ref{Fig_tfi}, most types of drone signal are concentrated in the central frequency region, while the beginning and end of the time dimension may contain incomplete signal information due to truncation. Therefore, the spatial weights are adaptively assigned to the central positions of \(H\times W\), with less attention paid to the time and frequency edges.

\section{Conclusion}
\label{sec5}
This paper proposed a drone signal OODD algorithm based on the cognitive fusion of ZC sequences and TFI under the RID task. The ZC sequences were identified by analyzing the communication protocols of DJI drones, while the TFI were generated to capture the time–frequency characteristics of drone signals with unknown or non-standard communication protocols. A feature extraction framework was designed to enhance and align the features from different modalities, followed by multi-modal feature interaction, single-modal feature fusion, and multi-modal feature fusion modules to achieve complementary feature integration. Discrimination scores along spatial and channel dimensions were computed and transformed into adaptive attention weights to improve the discriminative capability of the fused features. Simulation results demonstrated that the proposed algorithm improved the RID and OODD metrics by 1.7\% and 7.5\%, respectively, compared with the existing algorithms. Furthermore, the proposed algorithm exhibited strong robustness across various flight conditions and drone types.

\end{document}